\documentclass[]{wechat}
\usepackage[toc,page,header]{appendix}

\usepackage{minitoc}
\usepackage{amsfonts}
\usepackage{amssymb}
\usepackage{tabularx}
\usepackage{listings}
\usepackage{xcolor}
\usepackage{cancel}

\usepackage{tabulary,multirow,xspace}
\usepackage{fixmath,mathtools,nicefrac,mmstyle}
\usepackage{subcaption}
\usepackage{caption}
\usepackage{wrapfig} % for wrapfigure
\usepackage[misc]{ifsym} % to use \Letter
\usepackage{colortbl}

\usepackage{wrapfig}
\usepackage{multicol}
\usepackage[most]{tcolorbox}
\usepackage{pifont}

\definecolor{codegreen}{rgb}{0,0.6,0}
\definecolor{codegray}{rgb}{0.5,0.5,0.5}
\definecolor{codepurple}{rgb}{0.58,0,0.82}
\definecolor{backcolour}{rgb}{0.95,0.95,0.92}
\definecolor{boxgreen}{RGB}{61,155,122}
\definecolor{boxgreenbg}{RGB}{212,240,227} 

\lstdefinestyle{mystyle}{
    backgroundcolor=\color{backcolour},   
    commentstyle=\color{codegreen},
    keywordstyle=\color{magenta},
    numberstyle=\tiny\color{codegray},
    stringstyle=\color{codepurple},
    basicstyle=\ttfamily\footnotesize,
    breakatwhitespace=false,         
    breaklines=true,                 
    captionpos=b,                    
    keepspaces=true,                 
    numbers=none,                    
    numbersep=5pt,                  
    showspaces=false,                
    showstringspaces=false,
    showtabs=false,                  
    tabsize=2
}
\definecolor{mygray1}{gray}{.95}
\definecolor{mygray2}{gray}{.9}
\definecolor{mygray3}{gray}{.95}
\usepackage{pifont}% http://ctan.org/pkg/pifont

\newlength\savewidth
\newcolumntype{x}[1]{>{\centering\arraybackslash}p{#1pt}}

\newcommand{\app}{\raise.17ex\hbox{$\scriptstyle\sim$}}

\makeatletter
\DeclareRobustCommand\onedot{\futurelet\@let@token\@onedot}
\def\@onedot{\ifx\@let@token.\else.\null\fi\xspace}

\makeatother

\makeatletter

\newcommand{\Rmnum}[1]{\expandafter\@slowromancap\romannumeral #1@}
\makeatother

\usepackage{xcolor}
\usepackage{graphicx}
\usepackage{amssymb}
\usepackage{pifont}
\usepackage{floatrow}
\usepackage{amsmath} 
\usepackage{float}
\usepackage{wrapfig}
\usepackage{multirow}
\usepackage{tcolorbox}
\tcbuselibrary{breakable, skins, raster}
\usepackage{listings}
\usepackage{listings}

\definecolor{commentgreen}{rgb}{0.1, 0.4, 0.1}
\definecolor{keywordblue}{rgb}{0.1, 0.1, 0.7}
\definecolor{stringred}{rgb}{0.7, 0.1, 0.1}

\lstdefinestyle{mystyle}{
    commentstyle=\color{commentgreen},
    keywordstyle=\color{keywordblue},   
    stringstyle=\color{stringred},
    basicstyle=\ttfamily\scriptsize, 
    breaklines=true,
    keepspaces=true,
    showstringspaces=false,
    frame=none,                     
    language=Python, 
}

\definecolor{gtable}{rgb}{0.0, 0.5, 0.0}

\usepackage[utf8]{inputenc}
\usepackage{amssymb}
\usepackage{bbding}
\usepackage{pifont}
\usepackage{wasysym}
\usepackage{utfsym}
\usepackage{fontawesome}
\usepackage{graphicx}
\definecolor{gray}{gray}{0.8} 

\definecolor{top}{HTML}{E7F0DC}         
\definecolor{baseline}{HTML}{EEEEEE}         
\definecolor{close_source}{HTML}{CBF1F5}     
\definecolor{open_source}{HTML}{FDE7BB}         
\definecolor{baseline}{HTML}{EEEEEE}  
\definecolor{title_green}{HTML}{3D9B7A}

\definecolor{gain}{HTML}{34a853}  %

\definecolor{lost}{HTML}{ea4335}  %

\definecolor{my_red}{HTML}{FF0000}         % 淡灰色
\definecolor{my_purple}{HTML}{AA96DA} 
\definecolor{my_orange}{HTML}{F07B3F} 

\definecolor{my_box_red}{HTML}{FFB4B4}   
\definecolor{my_box_purple}{HTML}{D3CEDF} 
\definecolor{my_box_orange}{HTML}{FFC3A1}

\newcommand{\harnessname}{\textsc{WeAgent-Harness}}
\newcommand{\benchname}{\textsc{VisTarget-Bench}}
\newcommand{\result}[2]{\ensuremath{#1\,{\scriptstyle\pm #2}}}
\newcommand{\bestresult}[2]{\ensuremath{\mathbf{#1}\,{\scriptstyle\pm #2}}}
\newcommand{\pointresult}[1]{\ensuremath{#1}}
\newcommand{\bestpointresult}[1]{\ensuremath{\mathbf{#1}}}
\definecolor{resultrow}{HTML}{DDF4DD}

\title{
  \textcolor{title_green}{WeAgent-MMSearch:} Native Text-Vision Interaction for Multimodal Search Agents
}

\author{
\centerline{
    Zongkai Liu $^{1,2}$\quad
    Hui Zhang $^{1}$\quad
    Liqiang Niu $^{1,\ddagger}$ \quad
    Zhen Cao $^{1}$ \quad
    Han Li $^{1}$ \quad
    Juntao Liu $^{1}$ \quad
}
\centering{
    Wenchao Chen $^{1}$ \quad
    Chengduo Zhao $^{1}$ \quad
    Chao Yu $^{2}$ \quad
    Fandong Meng $^{1,\dagger}$ \quad
}
}
\affiliation[1]{Weixin AI, Tencent}
\affiliation[2]{Sun Yat-sen University}
\contribution[\ddagger]{Project Lead}
\contribution[\dagger]{Corresponding author}

\abstract{
Multimodal search agents extend parametric knowledge with newly emerging and
long-tail evidence from the open web. Yet many existing agentic search environments often expose
retrieved evidence only as text and omit tool-returned images from subsequent
context, reducing visually grounded trajectories to text-only reasoning.
Long-horizon interaction also compounds tool-call, response-length, timeout,
and budget failures, which can discard salvageable trajectories, waste rollout
computation, and disturb policy updates.
To address these issues, we introduce \harnessname, a multimodal agentic harness that supports native text–vision interaction and runtime recovery. Retrieved images receive persistent disk references, allowing the model to inspect, process, and cite them throughout the trajectory. 
Based on this harness, we develop
\textbf{WeAgent-MMSearch}, an integrated system spanning data construction, agentic
post-training, and multimodal rollout. 
For data construction, a strong MLLM uses \harnessname{} to discover, synthesize,
and verify MMSearch-style tasks and collect expert trajectories. 
During post-training, our Failure-Aware
GSPO (FA-GSPO) recovers salvageable abnormal rollouts and filters invalid ones
to improve bounded multimodal planning and search.
We also introduce \benchname, a 150-task human-verified benchmark that pairs
each question with a held-out target image, distinguishing image-retrieval
failures from visual-perception failures. 
Evaluation on \benchname{} and seven public benchmarks shows that agentic
post-training improves the average score by 19.22 points, enabling our model
to outperform similarly sized open-source models and rival models with roughly
ten times its parameter count.
\par\vspace{1mm}
{\small\textbf{Project page:}\enspace
\href{https://kkkaiaiai.github.io/WeAgent-MMSearch/}{\texttt{https://kkkaiaiai.github.io/WeAgent-MMSearch/}}}
}

\begin{document}
\maketitle

\begin{figure}[H]
\centering
\vspace{-2mm}
\includegraphics[width=0.91\textwidth]{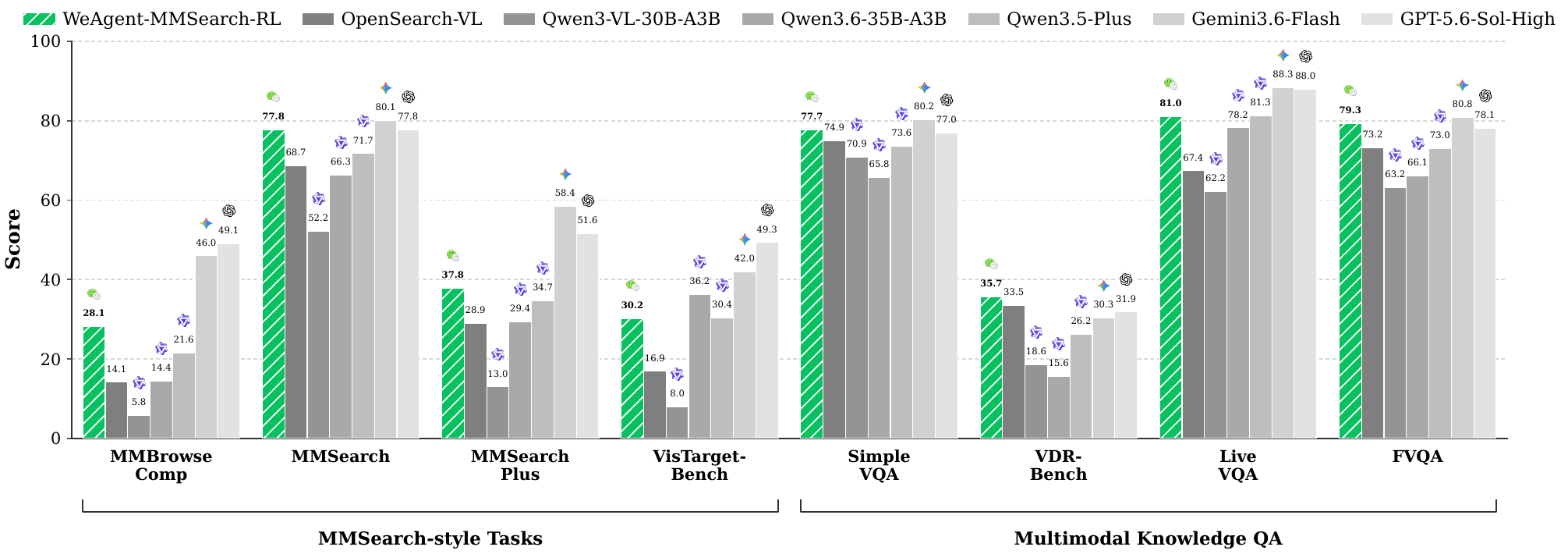}
\caption{Benchmark performance of WeAgent-MMSearch.}
\label{fig:main-results}
\end{figure}

\begin{figure*}[t]
    \centering
    \IfFileExists{images/agent_flow_en.pdf}
    {\includegraphics[width=\textwidth]{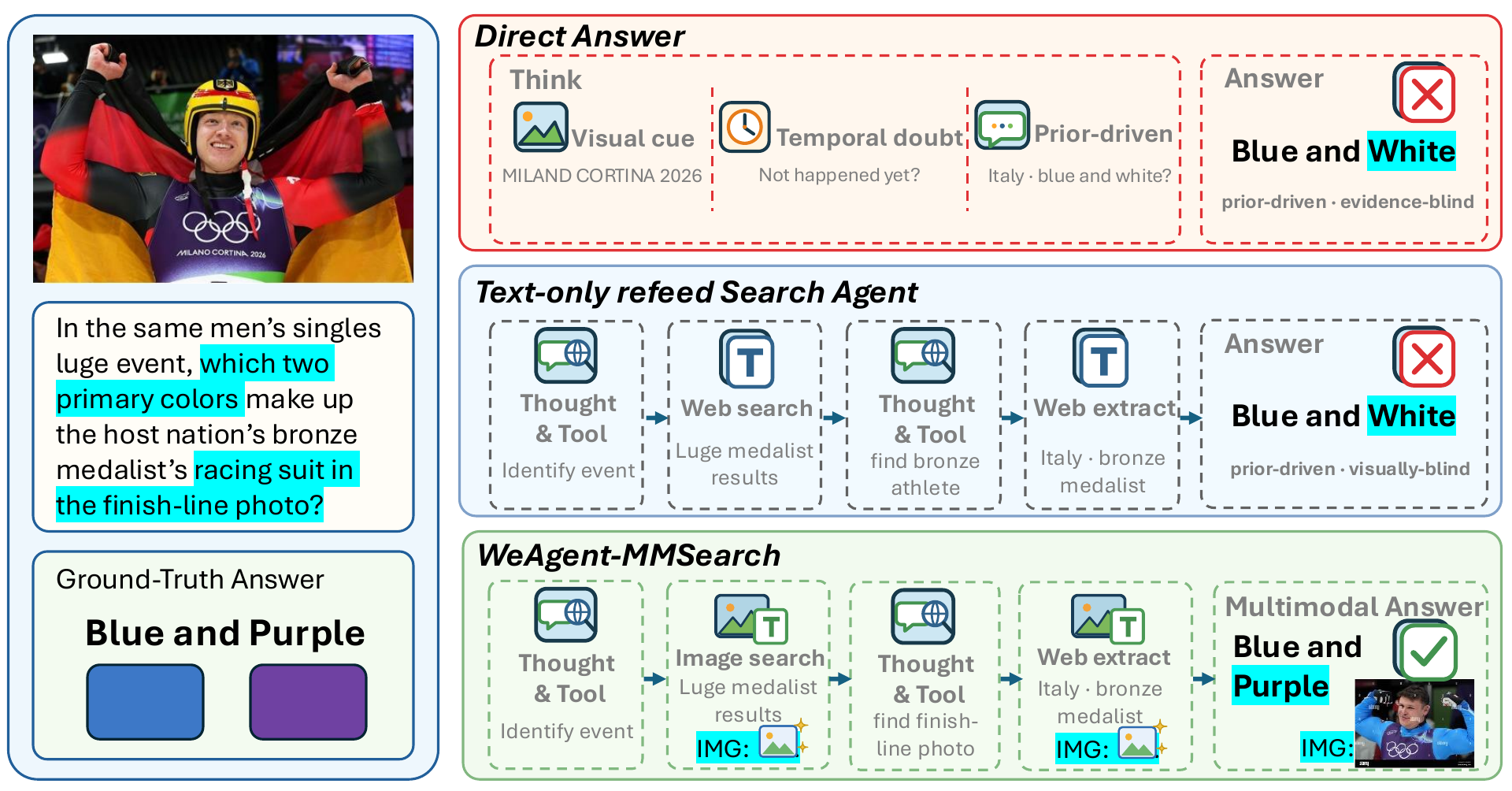}}
    {\fbox{\parbox{0.95\textwidth}{Figure asset \texttt{images/agent\_flow\_en.pdf} is unavailable.}}}
    \caption{A time-sensitive multi-hop search example. An MLLM fails when
    answering directly (\emph{top}). The same MLLM (e.g., Kimi K2.6) also fails
    with a text-only harness that omits retrieved images (\emph{middle}).
    WeAgent-MMSearch preserves and re-feeds retrieved images across turns and
    answers correctly (\emph{bottom}).}
    \label{fig:agent-flow}
    \end{figure*}

    \section{Introduction}
    
    Multimodal large language models (MLLMs) acquire broad visual and textual knowledge from static corpora\citep{openai2024gpt4ocard,team2023gemini,bai2025qwen3vl}, but finite coverage
    and parametric memory leave that knowledge incomplete and increasingly stale.
    Long-tail entities, recent
    events, and source-specific facts may therefore elicit unsupported or outdated
    answers. External tools extend this fixed knowledge boundary: search retrieves
    current evidence, webpages supply source context, and image tools expose visual
    facts unavailable from text alone. Multimodal search (MMSearch) consequently
    requires models to identify information gaps, invoke appropriate tools, and
    integrate external multimodal evidence
    \citep{jin2025searchr1,li2025websailor,hu2023avis,zhang2024vsa,jiang2024mmsearch}.
    
    A common deficiency in many existing agent environments is incomplete visual feedback.
    Many search environments primarily present retrieved evidence as text and omit
    tool-returned images from subsequent policy context. Visual facts that cannot
    be faithfully verbalized may therefore disappear from the model-visible state,
    causing nominally multimodal trajectories to collapse into text-only reasoning.
    Figure~\ref{fig:agent-flow} compares direct answering, an MLLM (e.g., Kimi
    K2.6) using a text-only harness, and WeAgent-MMSearch with persistent visual
    re-feed. Missing critical information from the image modality prevents the
    model from reaching the correct answer.
    
    Additionally, long-horizon interaction amplifies tool-name or parameter errors,
    oversized model outputs, timeouts, and exhausted budgets. These abnormalities
    not only interrupt reasoning and prevent valid answers at inference time, but
    also waste the rollout computation already spent during training. Timeouts and
    overlong trajectories can further create a long tail of disproportionately
    expensive episodes, reducing rollout throughput and overall training
    efficiency. If retained as ordinary samples, these abnormalities also
    contaminate the optimization signal with failures unrelated to the quality of
    the underlying search policy.
    
    A separate major challenge is the scarcity of high-quality supervision. Unlike
    standard visual question answering (VQA) or text-only multi-hop retrieval,
    MMSearch-style tasks must
    retain visual dependence across multiple retrieval steps and remain executable
    under the target tool contract. Their expert trajectories additionally require
    live search, webpage access, image acquisition, and code execution. Task
    synthesis, validity checking, and rollout collection therefore jointly limit
    the supply of open MMSearch-style supervision
    \citep{chu2026redsearcher,chen2026opensearchvl}.
    
    To address these issues, we introduce \harnessname, a multimodal agentic
    harness that supports native text--vision interaction and runtime recovery.
    It integrates search, webpage, image, reverse-image, and code tools. Retrieved
    images are stored on disk under persistent references with provenance,
    allowing the model to inspect, process, and cite them throughout the
    trajectory. This gives the model direct access to information unavailable in
    the text modality. The harness also caches repeated requests, repairs only unambiguous call errors,
    returns structured error observations for self-correction, and intervenes before
    local failures or budget exhaustion invalidate the entire rollout. Detecting
    and handling these abnormalities improves answer completion, reduces inference
    computation wasted on failed interactions, salvages more valid training
    samples, and prevents irrecoverable execution artifacts from contaminating
    policy updates.
    
    Based on this harness, we develop \textbf{WeAgent-MMSearch}, an integrated
    system spanning data construction, agentic post-training, and multimodal
    rollout. During data construction, a strong MLLM uses \harnessname{} as an
    expert MMSearch agent. Operating under the same
    tool and state contract used for training and inference, it discovers seeds
    from Wikipedia \citep{wikipedia} and time-sensitive web events, expands them backward into
    multi-hop reasoning chains, verifies chain validity, composes complete MMSearch-style
    tasks, and collects expert trajectories. This four-stage pipeline constructs
    3.3K SFT trajectories and 4.9K RL prompts in-house. Separately, we collect,
    filter, and normalize 56.2K open-source SFT trajectories and 8.3K open-source
    RL prompts, yielding final mixtures of 59.5K and 13.3K, respectively.
    
    We perform agentic post-training on these data mixtures. SFT teaches a basic
    ReAct-style search policy \citep{yao2022react} from successful demonstrations,
    with tool observations
    and harness interventions masked from the next-token prediction loss. For RL,
    we develop Failure-Aware GSPO (FA-GSPO), an agentic extension of Group Sequence
    Policy Optimization (GSPO) \citep{zheng2025gspo}. It couples group-sequence
    updates with runtime recovery: salvageable abnormal rollouts are retained,
    while remaining invalid samples are filtered from the training signal. These
    mechanisms improve bounded multimodal planning and search.
    
    We also introduce \benchname, a benchmark of 150 human-verified tasks that
    pairs each question with a held-out target image, distinguishing
    image-retrieval failures from visual-perception failures. Evaluation on
    \benchname{} and seven public benchmarks shows that \harnessname{} raises
    the average score of Qwen3-VL-30B-A3B from \(19.13\%\) to \(36.75\%\) and that of
    Kimi K2.6 from \(21.92\%\) to \(58.42\%\), without parameter updates.
    Starting from WeAgent-MMSearch-Base, which combines Qwen3-VL-30B-A3B with
    \harnessname{}, agentic post-training yields WeAgent-MMSearch-SFT
    and WeAgent-MMSearch-RL with average scores of \(51.04\%\) and \(55.97\%\),
    respectively. The final RL model outperforms similarly sized open-source models
    and rivals models with roughly ten times its parameter count.
    
    Our contributions are threefold:
    \begin{itemize}
        \item We introduce \harnessname, a general-purpose multimodal agentic
        harness that supports native text--vision interaction and runtime
        recovery across search, webpage, image, reverse-image, and code tools.
        It stores retrieved images on disk under persistent references with
        provenance, allowing the model to inspect, process, and cite them
        throughout the trajectory. Cache-backed execution, unambiguous error
        recovery, and budget control preserve salvageable rollouts.
        \item Based on this harness, we develop WeAgent-MMSearch with a four-stage
        data construction pipeline. A strong MLLM uses \harnessname{} to discover
        and synthesize MMSearch-style tasks, verify their validity, and collect
        expert trajectories under the same tool and state contract used at
        inference time.
        \item We develop an agentic post-training recipe that combines agentic
        SFT with our FA-GSPO. Agentic SFT learns basic search behavior, while
        FA-GSPO recovers salvageable abnormal rollouts and filters invalid ones
        to improve bounded multimodal planning and search. We also introduce
        \benchname, a benchmark of 150
        human-verified tasks that pairs each question with a held-out target
        image, distinguishing image-retrieval failures from visual-perception
        failures. Evaluation on \benchname{} and seven public benchmarks shows
        that agentic post-training improves the average score by 19.22 points,
        enabling WeAgent-MMSearch-RL to outperform similarly sized open-source
        models and rival models with roughly ten times its parameter count.
    \end{itemize}

\section{Related work}

\paragraph{Agentic search.}
Early visual information-seeking systems follow modular, complementary designs:
AVIS coordinates a planner and reasoner with web, image-search, and visual
perception tools, whereas Vision Search Assistant couples a vision-language
model with a web agent \citep{hu2023avis,zhang2024vsa}. More recent MMSearch
agents adapt the end-to-end reinforcement-learning paradigm of text-search
agents \citep{jin2025searchr1,li2025websailor} to train a single vision-language
model to interleave reasoning and search
\citep{wu2025mmsearchr1,deepmmsearch_r1,huang2026vision,webwatcher}.
OpenSearch-VL and Skywork-R1V4 further expand the tool set with active image
manipulation, including OCR, cropping, and image enhancement
\citep{chen2026opensearchvl,zhang2025skyworkr1v4}. Across these systems, images
embedded in retrieved webpages are often absent from model-visible observations,
and support for retaining tool-returned images for use in later turns remains
uneven. \harnessname{} addresses this gap by persistently storing tool-returned
images under stable references and making them available to subsequent reasoning
and tool calls.

\paragraph{Data and optimization for long-horizon search.}
Long-horizon search learning is constrained by both task scarcity and the cost
of collecting tool-interleaved trajectories. REDSearcher controls synthetic
task difficulty through reasoning-graph topology and evidence dispersion, then
co-designs mid-training and post-training to reduce interaction cost
\citep{chu2026redsearcher}. OpenSearch-VL samples Wikipedia paths, applies fuzzy
entity rewriting and source-anchored visual grounding, and combines the
resulting data with a diverse tool environment and failure-aware agentic RL
\citep{chen2026opensearchvl}. These methods primarily synthesize multi-hop
tasks from existing entity sources, leaving systematic discovery and collection
of new seed entities---especially from time-sensitive events---outside their
main scope. In contrast, our harness-native pipeline expands beyond fixed
entity sources by discovering both Wikipedia entities and time-sensitive web
events, then constructing visually dependent backward multi-hop tasks under
the same execution contract used for training and inference.

\paragraph{Multimodal search benchmarks.}
MMSearch evaluates multimodal search through component-level and end-to-end
tasks \citep{jiang2024mmsearch}; MMSearch-Plus emphasizes provenance-aware
search and evidence attribution \citep{tao2025mmsearchplus}; and MM-BrowseComp
uses expert-crafted, multi-hop browsing questions grounded in multimodal web
content \citep{li2025mmbrowsecomp}. These benchmarks broaden the scope and
difficulty of multimodal information seeking, but they are not designed around
the target-image diagnostic studied here. In \benchname, every question is
paired with a held-out target image, and the agent returns both its answer and
an index to the supporting image. Comparing this selection with the held-out
target separates final-answer accuracy from target-image retrieval and from
answer correctness after a match, helping diagnose whether the bottleneck is
finding the intended image or using it.

\section{\harnessname: Multimodal Agentic Harness}

\begin{figure*}[t]
\centering
\includegraphics[width=\textwidth]{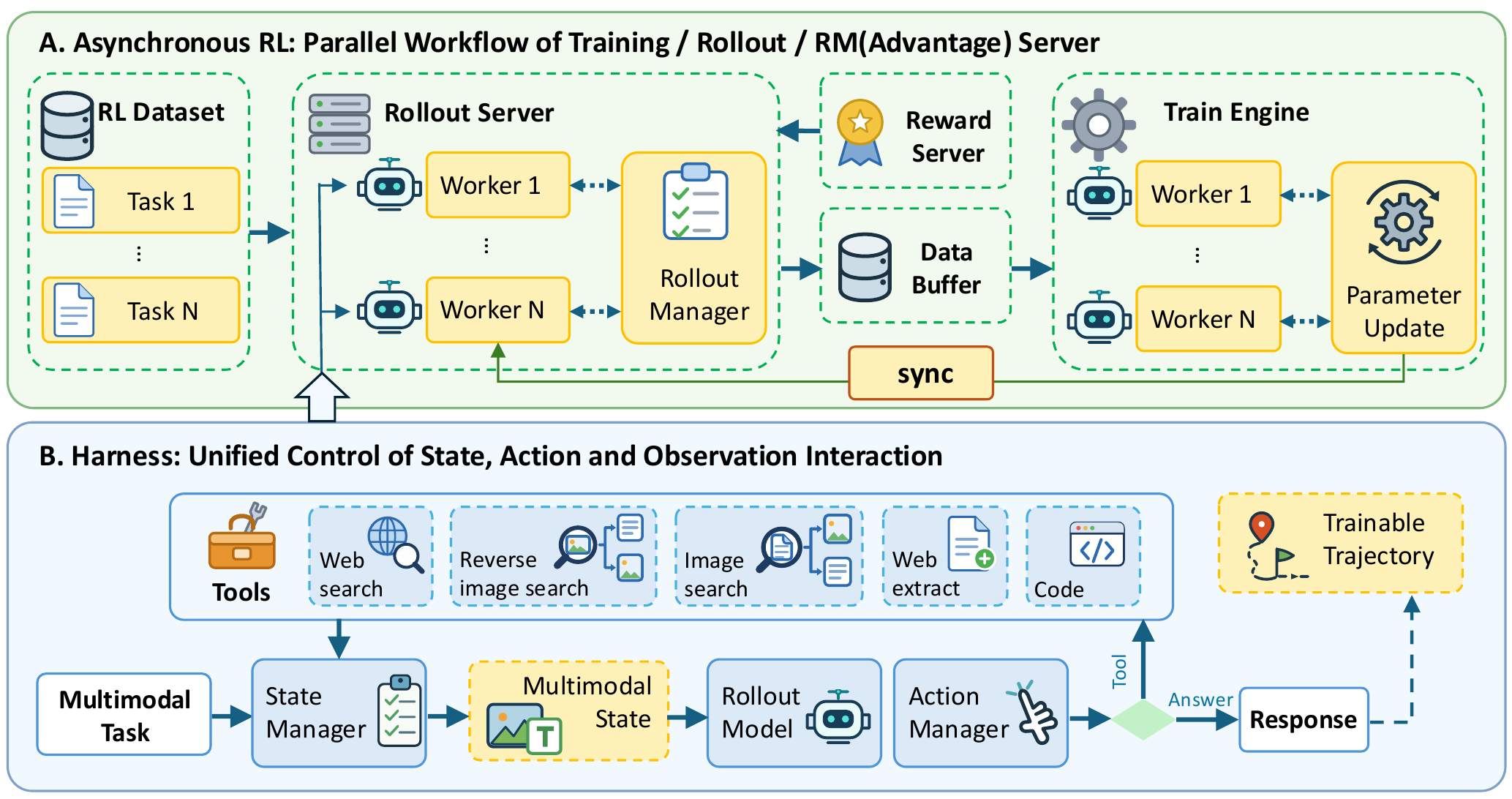}
\caption{Overview of the asynchronous RL infrastructure and \harnessname.
\textbf{Top:} the SGLang rollout server, SGLang/API reward server, and Megatron
training engine \citep{zheng2024sglang,shoeybi2019megatron} advance concurrently.
\textbf{Bottom:} \harnessname{} owns the state, action, and text--image
observation cycle between the rollout model and tool server. It manages
evidence across turns, routes tool calls, constructs aligned multimodal
observations, and serializes the result as a trainable trajectory.}
\label{fig:harness-overview}
\end{figure*}

\harnessname{} is the shared interaction substrate for task construction,
trajectory collection, post-training, and inference. Its design addresses two
requirements established in the Introduction. First, the policy must retain
newly acquired visual evidence as usable state throughout a long interaction.
Second, open-web execution must remain reliable and bounded enough to produce
trainable trajectories. To this end, \harnessname{} exposes
\texttt{web\_search}, \texttt{web\_extract},
\texttt{web\_search\_for\_image},
\texttt{web\_search\_based\_on\_image}, and \texttt{execute\_code} through one
multimodal state and evidence contract. Figure~\ref{fig:harness-overview}
situates this contract within the asynchronous RL infrastructure.

\subsection{Persistent Multimodal State}

\harnessname{} maintains a persistent evidence workspace across turns. Textual
history and registered images are jointly serialized into each subsequent
policy context, allowing newly retrieved pixels to directly condition later
reasoning and tool calls.

\paragraph{Addressable evidence objects.}
% Table~\ref{tab:harness-tools} summarizes the native tool interface.
The outputs of \texttt{web\_extract}, \texttt{web\_search\_for\_image},
\texttt{web\_search\_based\_on\_image}, and \texttt{execute\_code} are
normalized into multimodal observations containing text and any returned
images. Each image is bound to a model-visible reference and stable identity,
such as \texttt{<user\_image\_1>} or \texttt{<search\_image\_3>}, while retaining
its source metadata. Later turns can therefore revisit, reverse-search, or
compare the same visual evidence without ambiguity.

\subsection{Reliable and Bounded Execution}
\label{sec:harness-execution}

\paragraph{Cache-backed search balances freshness and latency.}
Open-web tool latency is heavy-tailed and can make a rollout group wait for a
small number of slow searches. We therefore maintain a shared, exact-match
cache over normalized search requests. Successful responses observed during
data collection and training are consolidated into the cache pool together
with their returned content and provenance. Distributed RL workers consume a
stable snapshot: a cache hit returns immediately through the same
tool schema, whereas a miss invokes the live search backend. 

This two-path policy trades a small amount of freshness for predictable
latency only when an identical request has already been observed; unseen or
time-sensitive requests still reach the live web. Periodic refreshes prevent
the pool from becoming a fixed offline corpus.

\paragraph{Recovery and budgets shape trainable rollouts.}
\harnessname{} intervenes when a runtime action is locally repairable. It
corrects unambiguous mismatches in tool names, parameter names, and call
serialization; when intent is unclear, it instead returns a structured error
observation so that the policy can self-correct. Repeated invalid calls are
detected and suppressed, breaking otherwise unproductive tool-use loops. When
a single model response exceeds \texttt{max-response-len}, the
overlong output is rejected and the policy receives one regeneration attempt,
preventing one anomalous turn from discarding an otherwise usable trajectory.
Explicit turn, tool, context, and wall-clock budgets bound every episode. When
the regular turn budget is exhausted, further tool use is disabled and the
policy receives one final evidence-synthesis turn. Trajectories that exceed the
overall context-length or wall-clock limit are instead terminated and excluded
from training. This design preserves recoverable trajectories for optimization
while preventing stalled, repetitive, or indefinitely long episodes from
dominating RL rollouts.

Together, these mechanisms define more than runtime behavior: persistent
evidence determines what the policy can observe, while cache lookup, recovery,
and budgets determine which interactions become finite, replayable training
episodes. The following sections use this same contract first as a data factory
and then as the environment for SFT, RL, and inference.

\section{Harness-Native MMSearch Data Pipeline}
\label{sec:data-pipeline}

We instantiate Kimi K2.6 on \harnessname{} as the expert MMSearch agent
\citep{moonshotai2026k26}. It uses the same tools, observations, and evidence
registry later available during post-training and inference. Four in-house
stages transform visually grounded seeds into executable trajectories: seed
generation, backward multi-hop expansion, task validation, and trajectory
sampling. A subsequent normalization step maps external corpora into the same
contract before the final SFT and RL mixtures are formed.

\begin{figure*}[t]
\centering
\includegraphics[width=\textwidth]{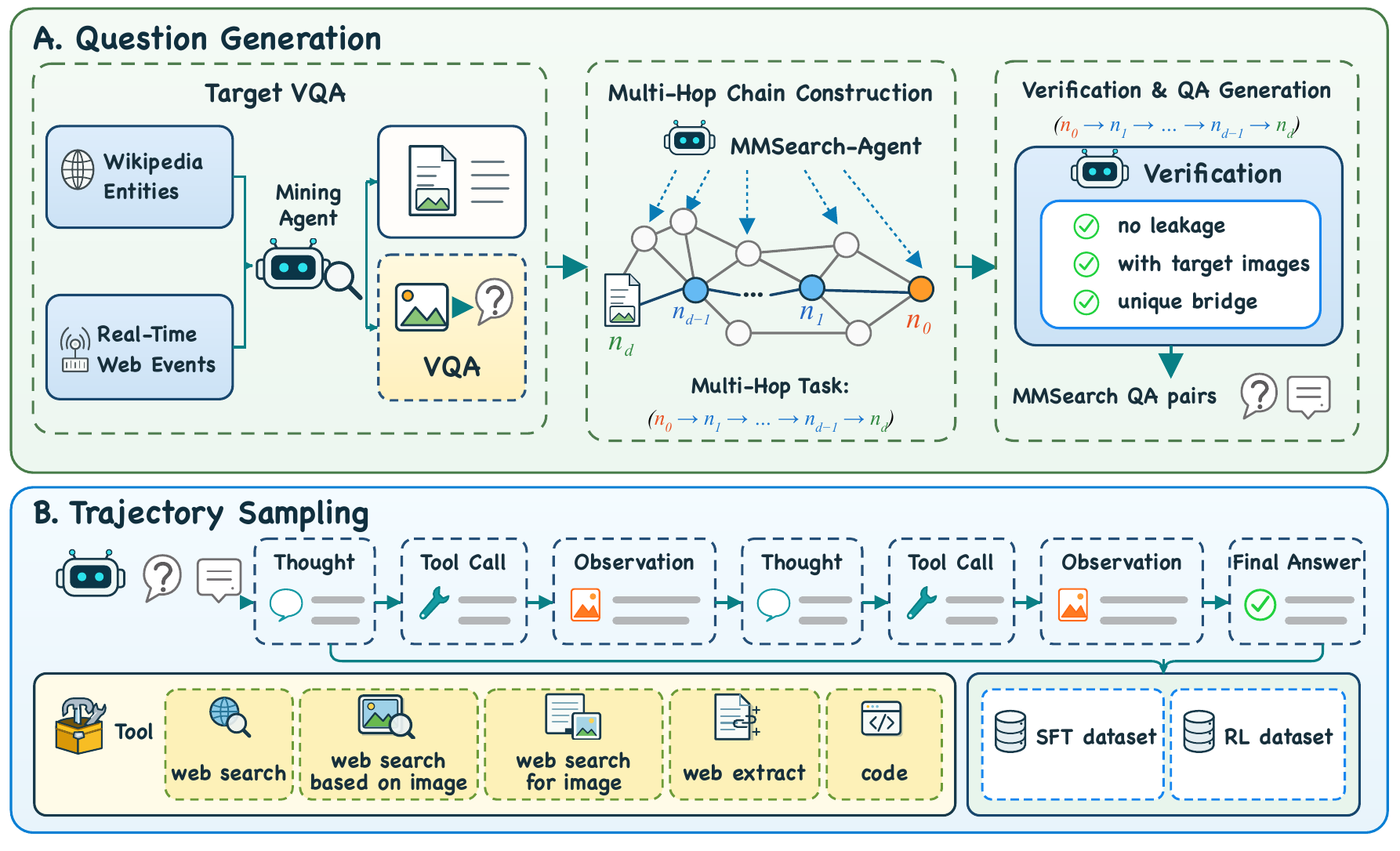}
\caption{Overview of the \harnessname-native data pipeline.
\textbf{A. Task construction:} Wikipedia entities and time-sensitive web events
provide target-image seeds, which are expanded backward into multi-hop chains
and validated for image dependence, bridge uniqueness, and answer leakage.
\textbf{B. Trajectory sampling:} the same harness executes accepted tasks;
successful trajectories form SFT data, while queries with mixed rollout
outcomes are retained for RL.}
\label{fig:data-construction}
\end{figure*}

Figure~\ref{fig:data-construction} shows the four-stage pipeline. The terminal
visual question-answering (VQA) item is derived from a Stage~1 visually grounded
seed and supplies a terminal image and pixel-dependent answer rather than the
final task itself. Earlier hops are generated backward from this endpoint, and
only chains that remain visually dependent and executable through
\harnessname{} are retained. The accepted tasks are then sampled as alternating
reasoning, tool-call, and multimodal-observation turns under the same evidence
contract used for learning.

\subsection{Stage 1: Visually Grounded Seed Generation}

We first construct a seed dataset of visually grounded facts. The expert
MMSearch agent searches time-sensitive topics across sports/games,
science/technology/nature, finance, business/brands/organizations, among others. For each topic,
construction requests advance through date windows at five-day intervals
between January~1 and July~30, 2026 (e.g., January~1--5 and January~5--10),
which limits repeated discovery of the same events. For each retained event,
the collector stores an authoritative source page, event date, ``as-of''
timestamp, localized target image, source provenance, and a concise fact that
must be read from that image.
For the complementary stable-fact branch, we download the publicly available
Wikipedia data \citep{wikipedia}, mine each record's concept name, definition, and category, and
insert only these fields into the construction prompt. The live-event branch
allows the seed pool to reflect information unavailable in a fixed corpus.

A seed is accepted only if its target image is reachable through \harnessname,
its answer depends on visible pixels rather than a title, URL, caption, or page
metadata, and its provenance can be audited. The search results collected in
this stage also populate the shared cache pool described above. Consequently,
the seed dataset contains both the terminal visual evidence needed for later
construction and an executable route by which an agent can recover it.

\subsection{Stage 2: Backward Multi-Hop Expansion}

Starting from each verified seed, the model expands the terminal visual fact
into a complex search problem. The resulting multi-hop
reasoning process follows the latent dependency chain $(n_0 \longrightarrow \cdots \longrightarrow n_{d-1} \longrightarrow n_d)$, 
where \(d\in\{2,3,4\}\) is the number of retrieval hops, \(n_d\) is the
verified VQA endpoint from Stage~1, and \(n_0\) is the visual or textual anchor
shown to the user. The endpoint contains target image \(C\) and a short answer
\(D\) that must be read from its pixels. For every intermediate node \(n_i\),
where \(1\le i<d\), the answer identifies the entity required to reach node
\(n_{i+1}\).

We construct this chain backward. Each agent invocation receives only the
current node \(n_i\): the expert MMSearch agent searches the open web for related
evidence and writes one QA pair whose answer resolves to that entity, producing
the immediate predecessor \(n_{i-1}\). A controller repeats this single-step
procedure \(d\) times to obtain the complete multi-hop retrieval chain. Solving
the composed task therefore requires following the bridge relations, retrieving
target image \(C\), and inspecting its pixels to produce \(D\).

\subsection{Stage 3: Task Validation}

After multi-hop expansion, the expert MMSearch agent performs semantic
verification of each candidate. Structurally, we require complete anchors,
bridge relations, endpoints, and provenance; at the retrieval level, we filter
out records with a missing target image, a non-unique bridge, or an answer
exposed by the URL, metadata, or question text. The agent further checks that
the reasoning chain is internally consistent and that the answer genuinely
depends on target image \(C\), then synthesizes the final user-facing query from
the validated chain. Each accepted record retains query \(q\), answer \(D\), the
latent reasoning chain, target image \(C\), and audit evidence.

\subsection{Stage 4: Trajectory Sampling}

The expert MMSearch agent next solves the accepted tasks under the same
\harnessname{} contract. Since retrieved images are re-fed into subsequent
turns, a sampled execution realizes the visual dependencies imposed during
construction.
We represent a solver trajectory as
\begin{equation}
    \tau =
    \bigl(q,(a_0,o_0),\ldots,(a_T,o_T),y\bigr),
\end{equation}
where $q$ is the constructed query, $T$ indexes the final interaction turn,
$a_t$ and $o_t$ are the action and observation at turn $t$, and $y$ is the
terminal answer. Observation $o_t$ may contain text and newly retrieved pixels.
The corresponding episode bundle preserves tool calls, observations, visual
evidence, and terminal outputs. Thus successful demonstrations are executable
records of the target environment.

The verified endpoint answer determines whether $y$ is correct. Correct and
complete trajectories form the SFT set. Failed or incomplete executions are
excluded from behavior cloning, but the underlying tasks remain useful for RL.
We construct the RL query set by stratifying model-solved and model-unsolved
tasks, balancing demonstrated feasibility against headroom for
policy improvement. Benchmark candidates are held out at the seed and task
levels before either split is used for optimization.

\subsection{Open-Source Normalization and Final Mixtures}
\label{sec:open-data-curation}

The four stages are followed by a separate mixture-construction step.
The resulting trajectories and prompts are complemented by open-source SFT and
RL data from Metis \citep{yan2026metis}, OpenSearch-VL
\citep{chen2026opensearchvl}, and REDSearcher \citep{chu2026redsearcher}.
These sources expose heterogeneous tool names, action schemas, and answer
conventions. Directly merging them would make the same serialized action carry
source-dependent semantics. We therefore convert all three sources into a
common contract before mixing them with the in-house data.

\paragraph{SFT normalization.}
We first remove records containing corrupted or garbled text, trajectories in
which any individual model response exceeds 4,096 output tokens, and
trajectories whose total serialized length exceeds 48K tokens. We then place
every tool in a source-specific namespace. For example, tools imported from
REDSearcher receive the prefix \texttt{RED\_}; analogous prefixes are assigned
to the other sources. This avoids collisions when two datasets use the same
tool name for different interfaces and gives the policy an unambiguous action
vocabulary.
Finally, every assistant turn is rewritten into one of two canonical forms: an
intermediate action contains
\texttt{<think>...</think><tool\_call>...</tool\_call>}, whereas a terminal
turn contains
\texttt{<think>...</think><response>...</response>}. Consequently, source
identity changes the available tool namespace but not the trajectory grammar
learned by the SFT policy.

\paragraph{RL query filtering.}
For every candidate query, we sample an 8-rollout group from
WeAgent-MMSearch-SFT and
discard queries for which all sampled answers are correct or all are incorrect.
Uniformly successful queries provide little headroom, while uniformly failed
queries provide no within-group reward contrast; both are uninformative for
group-relative policy optimization. We additionally use Kimi K2.6
\citep{moonshotai2026k26} to audit the reference labels and remove ambiguous or
non-verifiable targets. This step filters, for example, some open-source records
whose nominal label is a refusal placeholder such as ``I cannot find the final
answer.'' The retained RL prompts therefore combine non-trivial outcome
variance with a concrete, judgeable target.

\paragraph{Dataset scale and composition.}
After normalization and filtering, the SFT mixture contains 59.5K
trajectories: 3.3K in-house trajectories (5.5\%) and 56.2K open-source
trajectories (94.5\%). The RL mixture contains 13.3K prompts: 4.9K in-house
prompts (37.2\%) and 8.3K open-source prompts (62.8\%). 

\section{\benchname: Diagnosing Visual Evidence Use}
\label{sec:evidence-benchmark}

Final-answer accuracy alone cannot reveal whether an agent retrieved the
intended visual evidence or used it correctly after retrieval. \benchname{}
therefore pairs every task with a held-out target image and asks the agent to
index its supporting image. Comparing this selection with the target separates
final-answer accuracy from target-image retrieval and from answer correctness
after a match.
The benchmark is constructed from seeds held out before SFT and RL splitting,
so it tests late-stage visual dependence without overlapping the training
mixtures.

\subsection{Construction and Human Verification}

\begin{figure*}[t]
\centering
\IfFileExists{images/vistarget_composition.pdf}
{\includegraphics[width=0.8\textwidth]{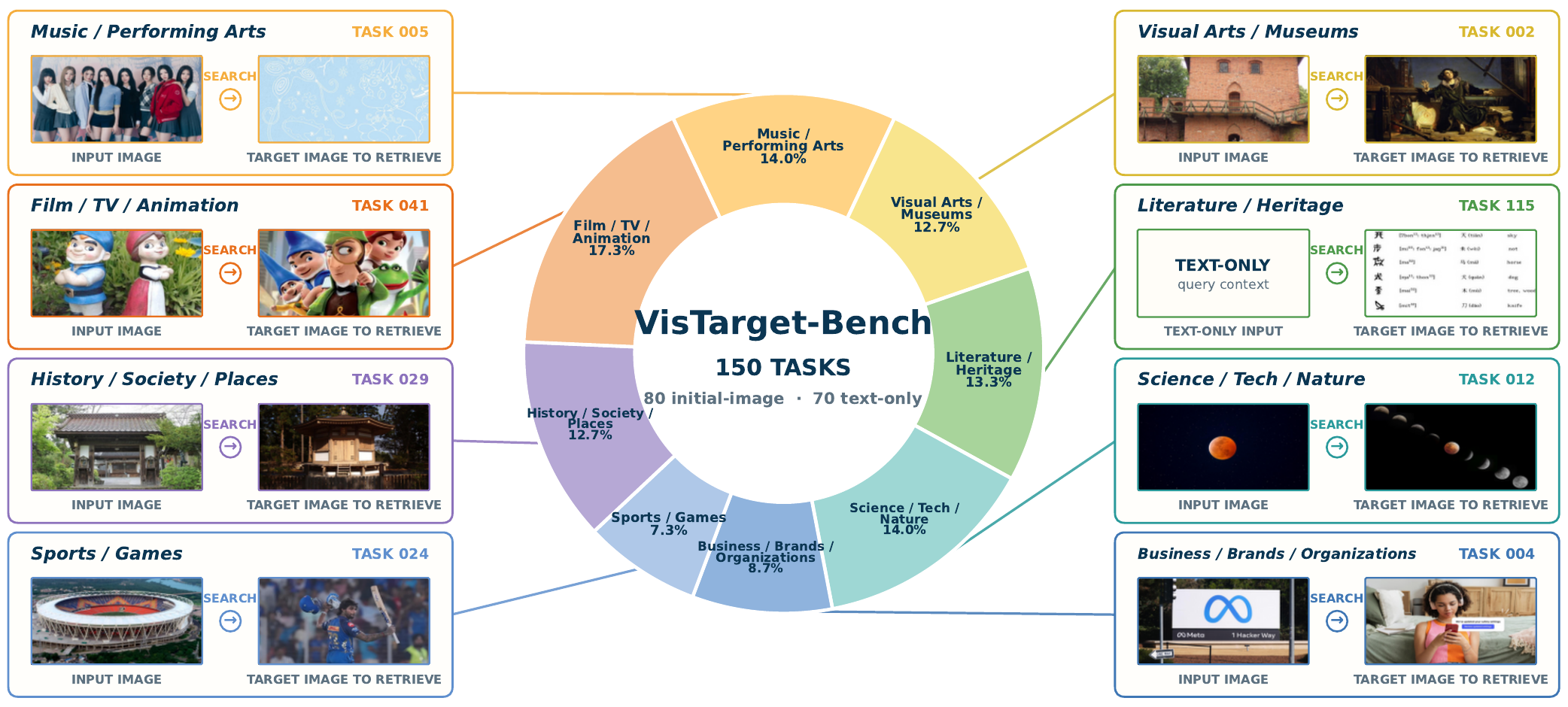}}
{\fbox{\parbox{0.95\textwidth}{Figure asset \texttt{images/vistarget\_composition.pdf} is unavailable.}}}
\caption{Composition and representative tasks of \benchname. The central
donut shows the refined terminal-evidence domain distribution, while each
surrounding pair shows an input image (or text-only input) followed by the
held-out target image that the agent must retrieve through image search.
Colored connectors associate each example with its domain.}
\label{fig:vistarget-composition}
\end{figure*}

\benchname{} follows the same harness-native construction process described in
Section~\ref{sec:data-pipeline}. Starting from held-out seeds, the expert
MMSearch agent performs backward multi-hop expansion, task validation, and
tool-based solvability checks through \harnessname. The only additional stage
is human verification and correction; Figure~\ref{fig:vistarget-composition}
summarizes the resulting benchmark and representative tasks.

Human reviewers correct factual and relational ambiguities, remove answer
shortcuts, and confirm that each task requires retrieving and reading the
target image through a valid tool path. The final benchmark contains 150 tasks:
80 (\(53.3\%\)) include an input image and 70 (\(46.7\%\)) begin with text only.

\subsection{Final-Answer and Visual-Evidence Metrics}

At evaluation time, the model returns a final answer and the index of its
supporting image. Accuracy is computed solely from the final answer.
For task \(i\), the selected image \(\widehat{C}_i\) and held-out target
\(C_i\) are presented to a VLM judge
\(J(\widehat{C}_i,C_i)\in\{0,1\}\). Let
\(I_i\in\{0,1\}\) indicate that the final response cites at least one
tool-retrieved image, \(M_i=J(\widehat{C}_i,C_i)\) denote target-image
correspondence, and \(A_i\in\{0,1\}\) denote final-answer correctness. If no
valid image is cited, \(M_i=0\). We report final-answer accuracy together with
three stage-wise image-grounding rates:
\begin{equation}
    \begin{aligned}
        \operatorname{Cite}
        = \frac{\sum_{i=1}^{N}I_i}{N},\;
        \operatorname{Match}
        = \frac{\sum_{i=1}^{N}M_i}{\sum_{i=1}^{N}I_i},\;
        \operatorname{Match{+}Acc}
        = \frac{\sum_{i=1}^{N}A_iM_i}{\sum_{i=1}^{N}M_i}.
    \end{aligned}
    \label{eq:image-grounding-rates}
\end{equation}
Accuracy measures final-answer correctness. Image Cite measures whether the
response includes visual evidence, Target Match measures whether a cited image
matches the held-out target, and Match+Acc measures answer correctness after
target acquisition. A conditional rate is defined as zero when its conditioning
set is empty.

\section{Agentic Post-Training}

Using the final mixtures from Section~\ref{sec:open-data-curation}, we perform
two complementary post-training stages. SFT learns basic reasoning and
tool-use behavior from successful demonstrations, while end-to-end RL further
encourages the model to discover more effective behavioral strategies.
The complete SFT, RL, and runtime tool configurations are reported in
Appendices~\ref{app:implementation} and~\ref{app:tool-config}.

\subsection{Supervised Fine-tuning}

We initialize the policy with correct, complete demonstration trajectories and
apply the standard next-token prediction objective. Environment observations
returned by tools and intermediate intervention prompts inserted by
\harnessname{} as user-role messages are masked; the loss is computed only on
tokens generated by the model. This stage teaches a basic ReAct-style retrieval
policy \citep{yao2022react}: at the beginning of each turn, the model reasons
over the available
evidence, searches when external information is needed, and revises its plan
after receiving a new visual observation.

\subsection{Agentic Reinforcement Learning}

SFT is limited to demonstrated behavior and does not teach the policy how to
recover when a long-horizon tool interaction goes wrong. We therefore optimize
the policy with Failure-Aware GSPO (FA-GSPO), our agentic extension of Group
Sequence Policy Optimization (GSPO) \citep{zheng2025gspo}. The central design
choice is to separate \emph{runtime recoverability} from
\emph{optimization eligibility}: candidate trajectories are sampled online,
and only trajectories that remain useful after recovery are retained for the
current policy update.

\paragraph{Failure-aware recovery.}
Vanilla GSPO treats every completion in a prompt group as a valid sample for
group-level reward normalization. This assumption may not hold during agentic
rollouts: a tool-call timeout or an overlong response can produce a truncated
trajectory whose signal is unrelated to MMSearch.
Figure~\ref{fig:harness-recovery} shows that such cases are not rare in
long-horizon search. FA-GSPO therefore recovers first and then filters. An
overlong turn is regenerated once, reducing the compute waste of discarding
the entire trajectory; a malformed tool call is returned as a structured
observation for self-correction on the next turn; and when the interaction
budget is exhausted, one final tool-free answer is triggered. Trajectories
that remain overlong, or that exceed the wall-clock or context budget, are
then excluded from optimization, so that tool-side and model-side failures do
not interfere with the main task signal.

\paragraph{Eligible-only group normalization.}
For each prompt $q_g$ in a batch \(\mathcal{B}=\{q_g\}_{g=1}^{B}\sim\mathcal{D}\), we draw a candidate group
\(\{o_{gi}\}_{i=1}^{G}\) online. After each rollout is recovered and scored, we dynamically construct
the eligible set
\begin{equation}
    \mathcal{E}_g =
    \left\{i\in\{1,\ldots,G\}\,\middle|\,
    o_{gi}\text{ remains trainable after runtime checks}\right\},
    \qquad n_g=\left|\mathcal{E}_g\right|.
    \label{eq:fagspo-eligible-set}
\end{equation}
The effective number of trajectories is therefore data-dependent. Only
trajectories in \(\mathcal{E}_g\) participate in group normalization:
\begin{equation}
    \bar{R}_g=\frac{1}{\max(n_g,1)}\sum_{i\in\mathcal{E}_g}R_{gi},\qquad
    \sigma_g=\sqrt{\frac{1}{\max(n_g-1,1)}
    \sum_{i\in\mathcal{E}_g}(R_{gi}-\bar{R}_g)^2}.
    \label{eq:fagspo-eligible-statistics}
\end{equation}
The reward is simple and is computed from the terminal answer:
\begin{equation}
    R_{gi}=0.9\,C_{gi}+0.1\,F_{gi},
    \qquad C_{gi},F_{gi}\in\{0,1\},
    \label{eq:fagspo-reward}
\end{equation}
where \(C_{gi}\) is the binary correctness judgment from the external answer
judge and \(F_{gi}\) indicates whether the final assistant turn obeys the
required response format. The eligible-only advantage is
\begin{equation}
    \widehat{A}_{gi} =
    \begin{cases}
        \displaystyle
        \frac{R_{gi}-\bar{R}_g}{\sigma_g+\delta},
        & i\in\mathcal{E}_g\text{ and }n_g\geq 2,\\[8pt]
        0,
        & i\notin\mathcal{E}_g\text{ or }n_g<2,
    \end{cases}
    \qquad \delta=10^{-6}.
    \label{eq:fagspo-filtered-advantage}
\end{equation}
Consequently, invalid trajectories contribute neither to the mean and
standard deviation nor to the policy gradient. A group with fewer than two
eligible trajectories is discarded from the current policy update because it
has no within-prompt preference signal.

\paragraph{Failure-Aware GSPO objective.}
We use the dynamically normalized advantages in the clipped, sequence-level
GSPO objective. For
\(\mathcal{B}=\{q_g\}_{g=1}^{B}\sim\mathcal{D}\), let
\(\mathbf{o}_g=\{o_{gi}\}_{i=1}^{G}\sim
\pi_{\theta_{\mathrm{old}}}(\cdot\mid q_g)\) and
\(\mathbf{o}=\{\mathbf{o}_g\}_{g=1}^{B}\). The expectation below is taken
over this joint sampling process:
\begin{equation}
\begin{aligned}
    \mathcal{J}_{\mathrm{FA\text{-}GSPO}}(\theta)
    &=\mathbb{E}_{\mathcal{B},\mathbf{o}}
    \left[
    \frac{1}{\max\!\left(\left|\mathcal{G}_{\mathrm{train}}\right|,1\right)}
    \sum_{g\in\mathcal{G}_{\mathrm{train}}}\frac{1}{n_g}
    \sum_{i\in\mathcal{E}_g}S_{gi}(\theta)
    \right],\\[-2pt]
    S_{gi}(\theta)
    &=\min\left(
        \rho_{gi}(\theta)\widehat{A}_{gi},\,
        \operatorname{clip}\!\left(
            \rho_{gi}(\theta),1-\epsilon_{\mathrm{lo}},1+\epsilon_{\mathrm{hi}}
        \right)\widehat{A}_{gi}
    \right),\\[-2pt]
    \rho_{gi}(\theta)
    &=
    \exp\left[
        \frac{1}{T_{gi}}\sum_{t=1}^{T_{gi}}
        \log\frac{\pi_{\theta}(a_{gi,t}\mid h_{gi,t})}
        {\pi_{\theta_{\mathrm{old}}}(a_{gi,t}\mid h_{gi,t})}
    \right],
\end{aligned}
\label{eq:fagspo-objective}
\end{equation}
where
\(\mathcal{G}_{\mathrm{train}}=\{g\in\{1,\ldots,B\}\mid n_g\geq 2\}\),
\(a_{gi,t}\) is the \(t\)-th model-generated action token in \(o_{gi}\),
\(T_{gi}\) is the number of such tokens, and \(h_{gi,t}\) contains the text,
images, and tool observations available before token \(t\).
The sequence-level ratio \(\rho_{gi}\) is the geometric mean of token-level
importance ratios.

\section{Experiments}

\subsection{Experimental setup}

We evaluate search ability on seven public benchmarks:
MMBrowseComp \citep{li2025mmbrowsecomp}, MMSearch
\citep{jiang2024mmsearch}, MMSearch-Plus \citep{tao2025mmsearchplus},
SimpleVQA \citep{cheng2025simplevqa}, VDR-Bench \citep{zeng2026vdrbench},
LiveVQA \citep{fu2025livevqa}, and FVQA \citep{wang2018fvqa}. We additionally
evaluate on \benchname. Results are scored by Kimi K2.6. Unless otherwise
stated, each evaluated configuration is run three times, with the mean and
standard deviation reported. Owing to the high inference cost, Gemini3.6-Flash,
Opus-5-High, and GPT-5.6-Sol-High are evaluated with a single run;
published open-source results are also point estimates.
The direct-answer baseline uses the released Qwen3-VL model
\citep{bai2025qwen3vl}.

Frontier baselines include Qwen3VL-30B-A3B\citep{bai2025qwen3vl}, Qwen3.6-35B-A3B \citep{qwen2026qwen36}, Qwen3.8-27B
\citep{qwen2026qwen38} Kimi K2.6\citep{moonshotai2026k26} and Qwen3.5-Plus
\citep{qwen2026qwen35}, as well as Gemini3.6-Flash\citep{google2026gemini36flash},
Opus-5-High\citep{anthropic2026opus5}, and GPT-5.6-Sol-High
\citep{openai2026gpt56sol}.
We evaluate the closed-source models Opus-5-High, GPT-5.6-Sol-High, and
Gemini3.6-Flash, together with Qwen3.5-Plus, via official APIs in thinking
mode. Kimi K2.6, Qwen3VL-30B-A3B, Qwen3.6-35B-A3B, and Qwen3.8-27B are self-deployed and
evaluated in non-thinking mode.
Lightweight search-agent baselines at a similar scale include
OpenSearch-VL-30B-A3B-Instruct
\citep{chen2026opensearchvl}, REDSearcher-MM-30B-A3B-Thinking
\citep{chu2026redsearcher}, and HyperEyes-30B-A3B-Instruct
\citep{li2026hypereyes}. We evaluate all frontier models with
\harnessname{} as their common harness. We take
published scores for lightweight search agents whenever available. Because
OpenSearch-VL releases its model and code, results absent from its report are
reproduced over three runs under our evaluation protocol and scored by Kimi K2.6.

\subsection{Main Results}

\begin{table*}[h]
\centering
\caption{Main comparison across seven public benchmarks and \benchname.
Reported baseline scores are point estimates; our results are means
\(\pm\) standard deviations over three runs unless marked with
\(^{\ddagger}\), which denotes a single run owing to high inference cost.
``Avg.'' covers all eight
benchmarks, and a dash denotes an unavailable result. \(^{*}\) denotes our
three-run reproduction; \(^{\dagger}\) denotes an unweighted average computed
from the per-benchmark means.
For MoE models, Params lists total parameters
with activated parameters in parentheses when disclosed.}
\label{tab:search-main}
\begingroup
\scriptsize
\setlength{\tabcolsep}{3.2pt}
\setlength{\arrayrulewidth}{0.3pt}
\setlength{\extrarowheight}{2.2pt}
\renewcommand{\arraystretch}{1.17}
\resizebox{\textwidth}{!}{
\begin{tabular}{lc|cccc|cccc|c}
\hline
& & \multicolumn{4}{c|}{\textbf{MMSearch-style Tasks}}
& \multicolumn{4}{c|}{\textbf{Multimodal Knowledge QA}} & \\
Model & Params & \shortstack{MMBrowse\\Comp} & MMSearch
& \shortstack{MMSearch\\Plus} & \benchname{} & \shortstack{Simple\\VQA}
& \shortstack{VDR\\Bench} & \shortstack{Live\\VQA}
& FVQA & \textbf{Avg.} \\
\hline
\rowcolor{gray!15}
\multicolumn{11}{c}{\textbf{Direct Answer}} \\
\hline
Qwen3-VL
& 30B (3B) & \result{7.66}{3.19} & \result{16.96}{2.68} & \result{2.36}{1.88}
& \result{0.00}{0.00} & \result{35.23}{0.29} & \result{2.02}{0.18}
& \result{37.50}{0.35} & \result{51.34}{0.83} & \bestresult{19.13}{0.47} \\
Kimi K2.6
& 1T (32B) & \result{2.53}{2.29} & \result{32.36}{5.27} & \result{4.72}{1.65}
& \result{1.99}{0.68} & \result{42.09}{0.72} & \result{4.00}{0.23}
& \result{50.15}{0.19} & \result{37.50}{0.83} & \bestresult{21.92}{0.75} \\
\hline
\rowcolor{gray!15}
\multicolumn{11}{c}{\textbf{Frontier Models with \harnessname}} \\
\hline
Gemini3.6-Flash\(^{\ddagger}\)
& -- & \pointresult{45.98} & \pointresult{80.12} & \pointresult{58.39}
& \pointresult{42.00} & \pointresult{80.20} & \pointresult{30.30}
& \pointresult{88.30} & \pointresult{80.78} & \bestpointresult{63.26} \\
GPT-5.6-Sol-High\(^{\ddagger}\)
& -- & \pointresult{49.11} & \pointresult{77.78} & \pointresult{51.61}
& \pointresult{49.33} & \pointresult{76.99} & \pointresult{31.85}
& \pointresult{88.00} & \pointresult{78.11} & \bestpointresult{62.85} \\
Opus-5-High\(^{\ddagger}\)
& -- & \pointresult{40.54} & \pointresult{80.70} & \pointresult{61.44}
& \pointresult{40.67} & -- & -- & -- & -- & -- \\
Kimi K2.6
& 1T (32B) & \result{33.63}{1.80} & \result{81.09}{2.05} & \result{49.08}{0.68}
& \result{33.78}{3.15} & \result{75.09}{0.54} & \result{30.43}{0.50}
& \result{85.00}{0.68} & \result{79.27}{0.58} & \bestresult{58.42}{0.66} \\
Qwen3.5-Plus
& 397B (17B) & \result{21.58}{1.36} & \result{71.73}{1.35} & \result{34.73}{1.34}
& \result{30.44}{0.77} & \result{73.59}{0.44} & \result{26.19}{0.25} & \result{81.29}{0.44}
& \result{73.05}{0.14} & \bestresult{{51.58}}{0.41} \\
Qwen3.6-35B-A3B
& 35B (3B) & \result{14.44}{0.68} & \result{66.28}{0.89} & \result{29.40}{0.17}
& \result{36.22}{0.38} & \result{65.79}{0.62} & \result{15.57}{0.15}
& \result{78.20}{0.38} & \result{66.09}{0.16} & \bestresult{46.50}{0.30} \\
Qwen3.8-27B
& 27B & \result{28.42}{0.25} & \result{69.59}{0.59} & \result{25.11}{0.60}
& \result{31.45}{1.07} & \result{73.17}{0.33} & \result{19.15}{0.62}
& \result{85.18}{0.26} & \result{70.43}{0.60} & \bestresult{50.31}{0.41} \\
\hline
\rowcolor{gray!15}
\multicolumn{11}{c}{\textbf{Lightweight Search Agents}} \\
\hline
OpenSearch-VL-30B
& 30B (3B) & \result{14.14}{0.93}\textsuperscript{*} & \pointresult{68.70}
& \result{28.94}{0.85}\textsuperscript{*}
& \result{16.89}{1.02}\textsuperscript{*} & \pointresult{74.90}
& \pointresult{33.50} & \pointresult{67.40} & \pointresult{73.20}
& \ensuremath{\mathbf{47.21}^{\dagger}} \\
REDSearcher-MM-30B
& 30B (3B) & \pointresult{23.50} & \pointresult{72.90}
& \pointresult{26.60} & -- & -- & -- & \pointresult{79.30} & -- & -- \\
HyperEyes-30B
& 30B (3B) & -- & \pointresult{86.90} & \pointresult{31.50}
& -- & -- & -- & \pointresult{81.60} & \pointresult{79.30} & -- \\
\hline
\rowcolor{gray!15}
\multicolumn{11}{c}{\textbf{Ours}} \\
\hline
\rowcolor{resultrow}
WeAgent-MMSearch-Base
& 30B (3B) & \result{5.80}{2.79} & \result{52.24}{3.33} & \result{12.98}{0.83}
& \result{8.00}{3.53} & \result{70.86}{0.84} & \result{18.65}{0.80}
& \result{62.21}{0.34} & \result{63.22}{0.83} & \bestresult{36.75}{1.01} \\
\rowcolor{resultrow}
WeAgent-MMSearch-SFT
& 30B (3B) & \result{21.87}{2.04} & \result{74.85}{1.55} & \result{31.00}{0.68}
& \result{24.00}{1.76} & \result{74.40}{0.24} & \result{28.93}{0.40}
& \result{80.61}{0.28} & \result{72.63}{0.50} & \bestresult{51.04}{0.28} \\
\rowcolor{resultrow}
WeAgent-MMSearch-RL
& 30B (3B) & \result{28.13}{0.73} & \result{77.78}{0.48} & \result{37.82}{0.26}
& \result{30.22}{0.32} & \result{77.73}{0.18} & \result{35.68}{0.25}
& \result{81.05}{0.06} & \result{79.33}{0.27} & \bestresult{55.97}{0.12} \\
\hline
\end{tabular}
}
\endgroup
\end{table*}

Table~\ref{tab:search-main} compares direct answering, frontier and lightweight
models using \harnessname, lightweight search agents, and our
WeAgent-MMSearch variants.
The substantial gains after SFT and RL demonstrate the effectiveness of both
the curated data and the post-training method. WeAgent-MMSearch-RL outperforms or
remains competitive with similarly scaled open-source search agents. Several
frontier models still retain an overall advantage, although post-training
narrows the gap and reaches comparable performance on several benchmarks.

\paragraph{\benchname{} results.}
\label{sec:vistarget-results}

\begin{figure*}[t]
\centering
\includegraphics[width=0.7\textwidth]{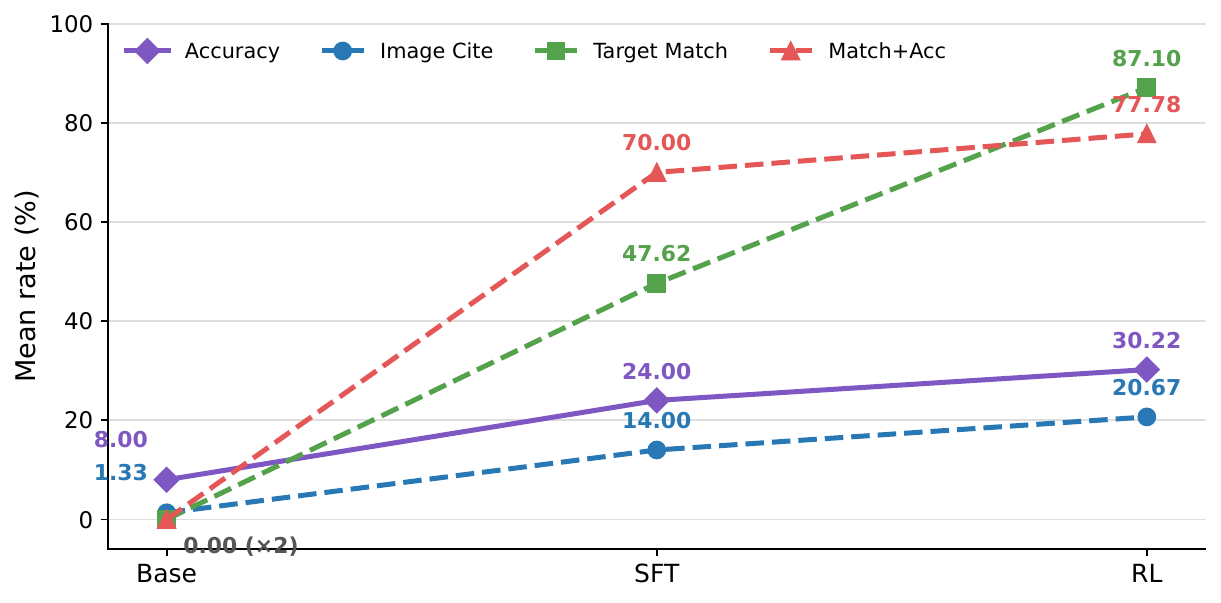}
\caption{Final-answer and stage-wise visual-evidence metrics on \benchname. All
systems use \harnessname. Accuracy and Image Cite are measured over all
responses, Target Match is conditioned on citing an image, and Match+Acc is
conditioned on a target match. Points show means over three runs.}
\label{fig:vistarget-evidence}
\end{figure*}

The no-tool diagnostic suggests that \benchname{} substantially reduces the
probability of answering correctly from parametric knowledge alone: Qwen3-VL
obtains \(0.00\pm0.00\%\) and Kimi K2.6 obtains
\(1.99\pm0.68\%\). Figure~\ref{fig:vistarget-evidence} compares final-answer
accuracy, image citation, target-image acquisition, and answer correctness
after target acquisition.

Post-training improves all four metrics. Final-answer accuracy rises from
\(8.00\%\) for WeAgent-MMSearch-Base to \(24.00\%\) after SFT and \(30.22\%\)
after RL, while image citation increases from \(1.33\%\) to \(14.00\%\) and
\(20.67\%\). Among responses with image citations, target matching reaches
\(47.62\%\) for SFT and \(87.10\%\) for RL. Once the target image is matched,
answer accuracy reaches \(70.00\%\) and \(77.78\%\), respectively, showing that
RL improves both visual-evidence acquisition and its use in final answering.

\subsection{Analysis Experiments}

\begin{table*}[t]
\centering
\caption{Harness-choice ablation for frontier models and \harnessname{} image
re-feed ablation for WeAgent-MMSearch-RL. Values are means \(\pm\) standard
deviations over three runs unless marked with \(^{\ddagger}\), which denotes a
single run owing to high inference cost. ``No image re-feed'' retains the
\harnessname{} tool interface but removes returned images from subsequent
model turns.}
\label{tab:harness-comparison}
\begingroup
\scriptsize
\setlength{\tabcolsep}{3.2pt}
\setlength{\arrayrulewidth}{0.3pt}
\setlength{\extrarowheight}{2.2pt}
\renewcommand{\arraystretch}{1.35}
\resizebox{\textwidth}{!}{
\begin{tabular}{l|cccc|cccc|c}
\hline
& \multicolumn{4}{c|}{\textbf{MMSearch-style Tasks}}
& \multicolumn{4}{c|}{\textbf{Multimodal Knowledge QA}} & \\
Configuration & \shortstack{MMBrowse\\Comp} & MMSearch
& \shortstack{MMSearch\\Plus} & \benchname{} & \shortstack{Simple\\VQA}
& \shortstack{VDR\\Bench} & \shortstack{Live\\VQA}
& FVQA & \textbf{Avg.} \\
\hline
\rowcolor{gray!15}
\multicolumn{10}{c}{\textbf{Harness Ablation}} \\
\hline
Kimi K2.6 + Hermes
& \result{23.39}{1.16} & \result{76.22}{4.11} & \result{50.96}{1.54}
& \result{20.01}{3.52} & \result{76.43}{0.40} & \result{29.83}{0.55}
& \result{84.47}{0.55} & \result{79.43}{0.42} & \bestresult{55.09}{0.18} \\
\rowcolor{resultrow}
Kimi K2.6 + \harnessname
& \result{33.63}{1.80} & \result{81.09}{2.05} & \result{49.08}{0.68}
& \result{33.78}{3.15} & \result{75.09}{0.54} & \result{30.43}{0.50}
& \result{85.00}{0.68} & \result{79.27}{0.58} & \bestresult{58.42}{0.66} \\
Gemini3.6-Flash\(^{\ddagger}\) + Hermes
& \pointresult{28.57} & \pointresult{82.46} & \pointresult{59.68}
& \pointresult{38.00} & \pointresult{80.05} & \pointresult{29.90}
& \pointresult{87.67} & \pointresult{81.28} & \bestpointresult{60.95} \\
\rowcolor{resultrow}
Gemini3.6-Flash\(^{\ddagger}\) + \harnessname
& \pointresult{45.98} & \pointresult{80.12} & \pointresult{58.39}
& \pointresult{42.00} & \pointresult{80.20} & \pointresult{30.30}
& \pointresult{88.30} & \pointresult{80.78} & \bestpointresult{63.26} \\
Qwen3.5-Plus + Hermes
& \result{22.02}{0.68} & \result{63.35}{0.89} & \result{37.34}{0.59}
& \result{21.55}{0.39} & \result{72.63}{0.47} & \result{26.47}{0.65}
& \result{82.43}{0.15} & \result{71.25}{0.23} & \bestresult{49.63}{0.26} \\
\rowcolor{resultrow}
Qwen3.5-Plus + \harnessname
& \result{21.58}{1.36} & \result{71.73}{1.35} & \result{34.73}{1.34}
& \result{30.44}{0.77} & \result{73.59}{0.44} & \result{26.19}{0.25}
& \result{81.29}{0.44} & \result{73.05}{0.14} & \bestresult{51.58}{0.41} \\
\hline
\rowcolor{gray!15}
\multicolumn{10}{c}{\textbf{\harnessname{} Image Re-feed Ablation}} \\
\hline
WeAgent-MMSearch-RL without image re-feed
& \result{13.69}{0.76} & \result{71.54}{0.73} & \result{23.26}{0.66}
& \result{10.44}{1.13} & \result{73.04}{0.53} & \result{29.97}{0.27}
& \result{79.43}{0.21} & \result{73.74}{0.23} & \bestresult{46.89}{0.27} \\
\rowcolor{resultrow}
WeAgent-MMSearch-RL with image re-feed
& \result{28.13}{0.73} & \result{77.78}{0.48} & \result{37.82}{0.26}
& \result{30.22}{0.32} & \result{77.73}{0.18} & \result{35.68}{0.25}
& \result{81.05}{0.06} & \result{79.33}{0.27} & \bestresult{55.97}{0.12} \\
\hline
\end{tabular}
}
\endgroup
\end{table*}

\paragraph{Harness ablation.}
Across the compared models, \harnessname{} is stronger than Hermes
\citep{nousresearch2026hermes}. For
Kimi K2.6 the eight-benchmark average rises from \(55.09\%\) to \(58.42\%\),
with the largest gains on MMBrowseComp and \benchname; for Gemini3.6-Flash
it rises from \(60.95\%\) to \(63.26\%\), and for Qwen3.5-Plus it rises from \(49.63\%\) to \(51.58\%\).
The advantage comes from richer multimodal tool observations and from
re-feeding retrieved images to later turns.
For WeAgent-MMSearch-RL, disabling image re-feeding causes a marked degradation
under the same \harnessname{} tool interface. Many evaluated benchmarks depend on
visual information returned during search; removing these images deprives the
policy of pixel-level evidence that cannot be recovered through text search
alone.

\begin{figure*}[t]
\centering
\includegraphics[width=0.9\textwidth]{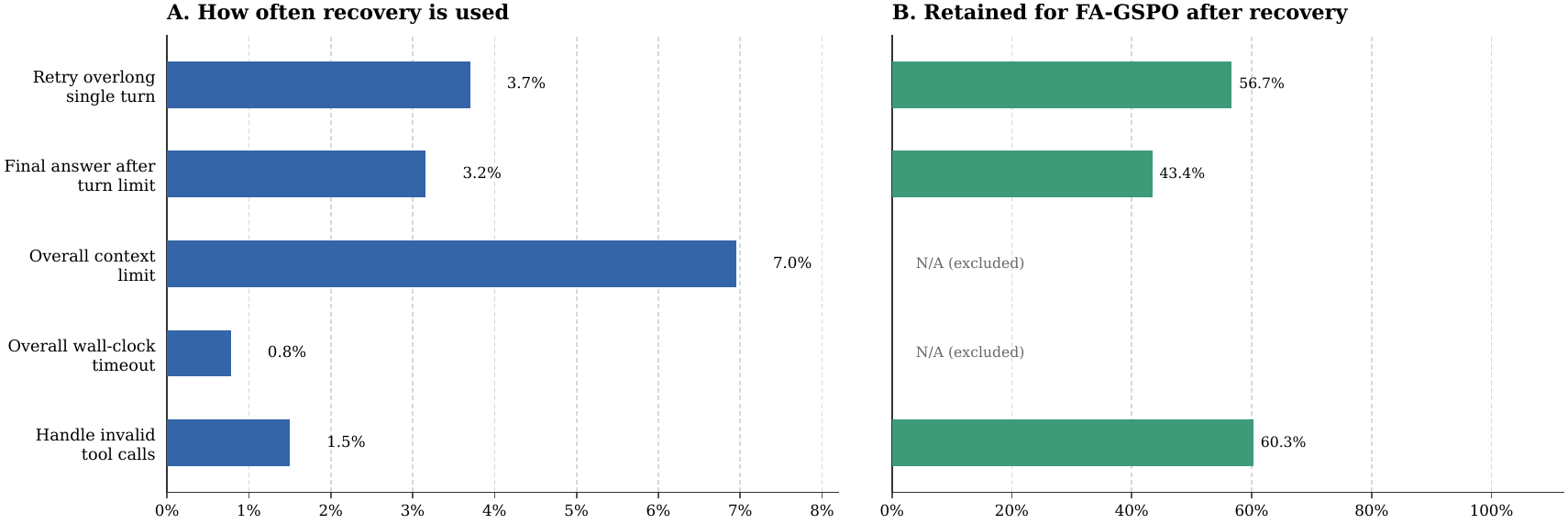}
\caption{Harness recovery on one RL run. \textbf{Left:} how often each
mechanism is triggered. \textbf{Right:} conditional on a trigger, the share
retained for FA-GSPO after recovery and runtime filtering. Overall context-limit
and wall-clock-timeout trajectories are N/A (excluded), because they are
filtered before this outcome is evaluated.}
\label{fig:harness-recovery}
\end{figure*}

\begin{table*}[t]
\centering
\caption{Data and optimization ablations. Raw Open-Source SFT uses only
open-source trajectories in their original formats; Normalized Open-Source SFT
converts the same data into the unified trajectory format; and Full SFT
additionally includes our harness-native data. Vanilla GSPO omits trajectory
self-recovery and invalid-trajectory filtering. Values are means \(\pm\)
standard deviations over three runs.}
\label{tab:data-rl-ablation}
\begingroup
\scriptsize
\setlength{\tabcolsep}{3.2pt}
\setlength{\arrayrulewidth}{0.3pt}
\setlength{\extrarowheight}{2.2pt}
\renewcommand{\arraystretch}{1.17}
\resizebox{\textwidth}{!}{
\begin{tabular}{l|cccc|cccc|c}
\hline
& \multicolumn{4}{c|}{\textbf{MMSearch-style Tasks}}
& \multicolumn{4}{c|}{\textbf{Multimodal Knowledge QA}} & \\
Configuration & \shortstack{MMBrowse\\Comp} & MMSearch
& \shortstack{MMSearch\\Plus} & \benchname{} & \shortstack{Simple\\VQA}
& \shortstack{VDR\\Bench} & \shortstack{Live\\VQA}
& FVQA & \textbf{Avg.} \\
\hline
\rowcolor{gray!15}
\multicolumn{10}{c}{\textbf{SFT Data Ablation}} \\
\hline
Raw Open-Source SFT
& \result{18.00}{1.36} & \result{67.25}{1.75} & \result{25.94}{1.13}
& \result{9.56}{2.69} & \result{66.93}{0.32} & \result{26.12}{0.33}
& \result{71.61}{0.12} & \result{67.15}{0.47} & \bestresult{44.07}{1.00} \\
Normalized Open-Source SFT
& \result{20.83}{1.12} & \result{71.73}{1.47} & \result{28.31}{1.15}
& \result{14.45}{4.34} & \result{73.12}{0.55} & \result{28.63}{0.33}
& \result{74.77}{0.06} & \result{72.67}{0.11} & \bestresult{48.06}{1.11} \\
\rowcolor{resultrow}
Full SFT
& \result{21.87}{2.04} & \result{74.85}{1.55} & \result{31.00}{0.68}
& \result{24.00}{1.76} & \result{74.40}{0.24} & \result{28.93}{0.40}
& \result{80.61}{0.28} & \result{72.63}{0.50} & \bestresult{51.04}{0.28} \\
\hline
\rowcolor{gray!15}
\multicolumn{10}{c}{\textbf{RL Optimization Ablation}} \\
\hline
Vanilla GSPO
& \result{21.58}{1.57} & \result{74.85}{2.68} & \result{36.81}{1.13}
& \result{9.78}{5.00} & \result{75.43}{0.52} & \result{32.35}{0.18}
& \result{80.31}{0.22} & \result{77.67}{0.39} & \bestresult{51.10}{1.41} \\
\rowcolor{resultrow}
Failure-Aware GSPO
& \result{28.13}{0.73} & \result{77.78}{0.48} & \result{37.82}{0.26}
& \result{30.22}{0.32} & \result{77.73}{0.18} & \result{35.68}{0.25}
& \result{81.05}{0.06} & \result{79.33}{0.27} & \bestresult{55.97}{0.12} \\
\hline
\end{tabular}
}
\endgroup
\end{table*}

\paragraph{Data and optimization ablations.}
Raw Open-Source SFT and Normalized Open-Source SFT use the same trajectories,
isolating the effect of format normalization. Converting these records into the
unified trajectory format raises average score from \(44.07\%\) to
\(48.06\%\). Adding our harness-native trajectories further raises it to
\(51.04\%\), with a larger
gain on \benchname{} from \(14.45\%\) to \(24.00\%\). These improvements show
that both a consistent trajectory grammar and demonstrations drawn from the
same tool-interaction distribution as \harnessname{} are important for agentic
SFT. Vanilla GSPO omits the proposed trajectory self-recovery and
invalid-trajectory filtering. The complete FA-GSPO recipe improves average
score from \(51.10\%\) to \(55.97\%\) and \benchname{} accuracy from
\(9.78\%\) to \(30.22\%\), demonstrating that the
proposed mechanisms protect learning from abnormal rollouts while preserving
recoverable training signal.

\paragraph{Trajectory recovery.}
Figure~\ref{fig:harness-recovery} reports recovery percentages from one RL
training run. For each mechanism, the trigger rate is its fraction of all
rollout trajectories. Conditional on a trigger, the retention rate is the share
of trajectories that remain in the FA-GSPO training set after recovery and
runtime filtering. Retrying an overlong single-turn output, forcing a final
answer after the regular turn limit, and returning invalid tool-call errors are
counted separately. We additionally report overall context-limit and
wall-clock-timeout triggers. These two trajectory-level violations are directly
excluded and are therefore reported as N/A (excluded). A trajectory may
activate more than one local mechanism, so trigger rates are not additive.

The recovery mechanisms retain a portion of locally abnormal executions for
optimization, avoiding the waste of otherwise usable rollout computation.
Directly excluding overall context-limit and wall-clock-timeout
trajectories prevents their length- and timeout-related artifacts from
introducing noise into policy optimization.

\section{Conclusion}

We presented \textbf{WeAgent-MMSearch}, an end-to-end multimodal search system
built around \harnessname. The harness preserves tool-retrieved images as
addressable policy state, re-feeds visual observations across turns, and
provides runtime recovery before local errors invalidate an entire trajectory.
The same interaction contract supports seed discovery, backward multi-hop task
construction, expert trajectory sampling, agentic SFT, FA-GSPO, and
inference.

The main experiments show that WeAgent-MMSearch-SFT and WeAgent-MMSearch-RL
improve the reported scores over WeAgent-MMSearch-Base, with the final model
reaching an eight-benchmark average of 55.97, a 19.22-point improvement over
the base checkpoint. In the reported comparison, it outperforms similarly sized
open-source models and is competitive with models having roughly ten times its
parameter count.
On \benchname,
Match+Acc does not achieve a full score even after the supporting image
matches the target image, showing that
retrieving the intended evidence does not guarantee successful visual
interpretation. WeAgent-MMSearch-RL improves this target-image-matched answer
accuracy over WeAgent-MMSearch-SFT.  Appendix~\ref{app:rl-cases} presents three RL rollouts and their decisive visual evidence.

The analysis experiments examine possible contributors to these gains. Across
the reported model comparisons, a text-only tool-return harness such as Hermes
is weaker than \harnessname, indicating that persistent image re-feeding can
improve the reported aggregate score. The recovery analysis records targeted
runtime interventions that can salvage locally abnormal executions, while
context-limit and wall-clock-timeout trajectories are excluded from
optimization. Together, persistent multimodal interaction, environment-aligned
learning, and runtime recovery form complementary components of the reported
system.

\clearpage

% \bibliographystyle{plainnat}
% \bibliography{main}

\clearpage

\appendix
\section{Implementation Details}
\label{app:implementation}

WeAgent-MMSearch-SFT updates all language-model and visual modules (vision
tower and projector) with a global batch size of 32 trajectories and learning
rates of $1.0\times10^{-5}$ for the language model and $1.0\times10^{-6}$ for
the vision tower. WeAgent-MMSearch-RL initializes from this SFT checkpoint,
freezes the vision tower and projector, and trains only the language-model
policy; it uses a prompt batch of 64 with 8 samples per prompt (global
trajectory batch 512), a constant learning rate of $5.0\times10^{-7}$, and
rollout temperature $1.0$.

\section{Tool Interfaces and Runtime Configuration}
\label{app:tool-config}

The reported system exposes five native tools.  A tool-returned image is saved
under a stable relative filename, registered in the episode evidence state, and
attached as visual input to the next model turn.  Table~\ref{tab:tool-interfaces}
defines the model-visible interfaces, and Table~\ref{tab:tool-runtime} gives the
episode budgets, recovery rules, and persistence policy.

\begin{table*}[ht]
\centering
\caption{Model-visible tool interfaces.  Limits are the active values in the
reported RL recipe.}
\label{tab:tool-interfaces}
\begingroup
\scriptsize
\setlength{\tabcolsep}{3pt}
\renewcommand{\arraystretch}{1.13}
\begin{tabular}{p{0.23\textwidth}p{0.32\textwidth}p{0.41\textwidth}}
\hline
\textbf{Tool} & \textbf{Arguments} & \textbf{Operation and returned evidence} \\
\hline
\texttt{web\_search} & \texttt{query}: string or up to 5 strings; \texttt{num}: 1--5 (default 3) & General web search. Returns titles, URLs, snippets, and publication times; query language selects the search locale automatically. \\
\texttt{web\_search\_for\_image} & \texttt{query}: string or up to 3 strings; \texttt{num}: 1--10 (default 5) & Text-to-image search. Localizes successful thumbnails, returns their titles/domains/relative paths, and directly re-feeds up to four images to the VLM. \\
\shortstack[l]{\texttt{web\_search\_based}\\[-1pt]\texttt{\_on\_image}} & \texttt{image}: a relative filename registered in the current rollout, or a list of up to 5 such filenames; \texttt{num}: 1--8 (default 5) & Reverse-image search. Returns related pages and an optional knowledge-graph match; result thumbnails are localized and up to five are re-fed at no more than 512 vision tokens each. \\
\texttt{web\_extract} & \texttt{urls}: up to 5 URLs; optional \texttt{goal} & Fetches HTML or PDF content and returns a goal-conditioned Markdown summary per URL. Pages are summarized by a LLM to at most 5,000 characters; up to five localized page images are also re-fed when available. \\
\texttt{execute\_code} & \texttt{code}: Python source & Runs Python in a persistent episode sandbox. Stdout/stderr are returned; newly generated raster images are registered and the first five are re-fed. \\
\hline
\end{tabular}
\endgroup
\end{table*}

\begin{table*}[t]
\centering
\caption{Harness budgets, recovery behavior, and system-level tool settings.}
\label{tab:tool-runtime}
\begingroup
\scriptsize
\setlength{\tabcolsep}{4pt}
\renewcommand{\arraystretch}{1.11}
\begin{tabular}{p{0.18\textwidth}p{0.29\textwidth}p{0.47\textwidth}}
\hline
\textbf{Category} & \textbf{Setting} & \textbf{Value / Behavior} \\
\hline
\textbf{Episode} & Turn / executed-call limits & 20 regular turns (plus at most one no-tool synthesis turn) / 32 calls \\
& Total context / wall-clock limit & 32,768 tokens / 900 seconds; trajectories crossing either limit are excluded from training \\
& Early final synthesis & Tool use is disabled once at most 4,096 context tokens remain; one no-tool answer turn is requested \\
\hline
\textbf{Recovery} & Argument repair & Deterministically repairs only unambiguous aliases, scalar-to-list conversions, and ignored extra fields; it never infers missing information \\
& Duplicate suppression & A call with the same tool name and canonicalized arguments as the immediately preceding executed call is not executed and instead receives a corrective observation \\
& Repeated format errors & After two consecutive identical invalid calls, a loop-guard observation instructs the model to change the call or answer with available evidence \\
& Overlong response & A single response exceeding the output-length limit is discarded and one replacement response is regenerated \\
& Max-turn finalization & After 20 regular turns, tools are disabled and one no-tool synthesis turn is allowed; it is retained only if it contains a complete terminal response \\
& Excluded trajectories & A trajectory hitting the context-length or wall-clock limit is excluded from training, irrespective of any subsequently generated answer \\
\hline
\textbf{Persistence} & General result threshold & Results above 16,384 characters are persisted with a 1,500-character in-context preview \\
& Web extraction & Length is controlled independently per URL; at most five 5,000-character summaries are returned \\
\hline
\end{tabular}
\endgroup
\end{table*}

\section{RL Prompts}
\label{app:rl-system-prompt}

Table~\ref{tab:rl-prompts} gives the system and user prompts used for all RL
rollouts and evaluations. At runtime, \texttt{CURRENT\_TIME} is instantiated at
hour granularity, while \texttt{TOOL\_LIST} is populated with the function
schemas.

\begin{table*}[p]
\centering
\caption{System and user prompts for RL rollouts and evaluations. Bracketed
fields are replaced by the corresponding runtime time, tool schemas, task,
image identifier, or filename.}
\label{tab:rl-prompts}
\begingroup
\scriptsize
\setlength{\tabcolsep}{4pt}
\renewcommand{\arraystretch}{1.16}
\begin{tabular}{p{0.15\textwidth}p{0.79\textwidth}}
\hline
\textbf{Prompt Type} & \textbf{Prompt} \\
\hline
\textbf{System Prompt} &
You are a helpful, knowledgeable assistant with strong multimodal
understanding. Use the available tools to solve the user's task. Be targeted
and efficient.

\medskip
\texttt{[Current date and time (Beijing Time, UTC+8): CURRENT\_TIME]}

\texttt{[Available tools and function schemas: TOOL\_LIST]}

\medskip
\textbf{Rules.}

\textbf{1.} Always use relative file paths in tool arguments. Do not hardcode
absolute paths.

\textbf{2.} Each turn: start with a short
\texttt{<think>...</think>} in the SAME language as the user's prompt, then
output EXACTLY ONE of the following: (i) a single tool call via the
function-calling protocol---do NOT write \texttt{<tool\_call>} JSON inside the
text content, and call at most one tool per turn; or (ii) a final answer wrapped
in \texttt{<response>...</response>}. Never emit a tool call, either textual or
structured, after \texttt{<response>}.

\textbf{3.} Uploaded images are shown to you directly AND are available as
files (\texttt{user\_image\_1.png}, \texttt{user\_image\_2.png}, \ldots); images
returned by tools (e.g., search thumbnails labeled \texttt{[Image N]}) are
likewise already visible. Do NOT call \texttt{execute\_code} just to open or
display an image you can already see---only run code on an image to crop or
zoom a small region or to measure or compute something you cannot judge by eye.
Pass relative paths to tools.

\textbf{4.} If a tool call fails or returns nothing useful, switch tactics by
using different arguments or a different tool. Retry the same kind of action at
most twice, then answer with the best available evidence.

\textbf{5.} Prefer a final reply that mixes text with supporting images when
image evidence exists. Inside \texttt{<response>...</response>}, if a claim is
backed by a saved image, embed that image inline with
\texttt{<IMG: filename>} (e.g., \texttt{<IMG: search\_*.jpg>}). Place each image
next to the claim it supports. Cite only the images that actually corroborate
the answer---do not dump every saved image. \\
\hline
\textbf{User Prompt} &
\texttt{[task instruction]}

\medskip
When input images are present, append their registered relative paths:\\
&\texttt{Registered image paths:}
\texttt{- user\_image\_1: [filename\_1]}, \ldots,
\texttt{- user\_image\_N: [filename\_N]}. \\
\hline
\end{tabular}
\endgroup
\end{table*}

\clearpage
\onecolumn
% Final-response comparison (no tool traces).
\definecolor{kimihead}{HTML}{B94A48}
\definecolor{kimibg}{HTML}{FDF2F2}
\definecolor{wehead}{HTML}{3D9B7A}
\definecolor{webg}{HTML}{EAF7F1}
\definecolor{querybg}{HTML}{F3F6F5}
\definecolor{queryrule}{HTML}{D5E2DC}

% Highlight styles for key question points and key response points.
\definecolor{hlquerybg}{HTML}{FFE599}   % soft yellow -- key question points
\definecolor{hlansbg}{HTML}{C9F0D5}     % soft green  -- key response points
\newcommand{\hlquery}[1]{\colorbox{hlquerybg}{#1}}
\newcommand{\hlans}[1]{\colorbox{hlquerybg}{#1}}

\long\def\respquery#1{%
  \begin{tcolorbox}[
      colback=querybg,
      colframe=queryrule,
      boxrule=0.45pt,
      leftrule=2.4pt,
      arc=1.2mm,
      left=7pt,
      right=6pt,
      top=4pt,
      bottom=4pt,
    ]
    \small\sffamily\noindent{\bfseries\color{wehead} Query.}~#1
  \end{tcolorbox}%
}

\newcommand{\resptarget}[2]{%
  \begin{center}
    {\sffamily\bfseries\small\color{wehead} Target image}%
    {\sffamily\small\color{black!50} (not provided to the model at query time)}\\[2pt]
    \includegraphics[width=#1,height=0.16\textheight,keepaspectratio]{#2}
  \end{center}%
}

\newcommand{\respcited}[2]{%
  \par\vspace{4pt}\centering
  \includegraphics[width=0.94\linewidth,height=0.15\textheight,keepaspectratio]{#1}\\[-1pt]
  {\sffamily\scriptsize\color{black!50}#2}%
}

\long\def\respcompare#1#2{%
  \begin{tcbraster}[
      raster columns=2,
      raster equal height,
      raster column skip=3.2mm,
      valign=top,
    ]
    \begin{tcolorbox}[
        colback=kimibg,
        colframe=kimihead,
        colbacktitle=kimihead,
        coltitle=white,
        boxrule=0.55pt,
        arc=1.8mm,
        left=6pt,
        right=6pt,
        top=6pt,
        bottom=7pt,
        title={\sffamily Kimi K2.6 + Hermes\hfill{\scriptsize Incorrect}},
        fonttitle=\sffamily\bfseries\small,
        fontupper=\footnotesize\sffamily,
        toptitle=1.5pt,
        bottomtitle=1.5pt,
      ]
      \setlength{\parskip}{0.32em}\setlength{\parindent}{0pt}#1
    \end{tcolorbox}
    \begin{tcolorbox}[
        colback=webg,
        colframe=wehead,
        colbacktitle=wehead,
        coltitle=white,
        boxrule=0.55pt,
        arc=1.8mm,
        left=6pt,
        right=6pt,
        top=6pt,
        bottom=7pt,
        title={\sffamily WeAgent-MMSearch-RL\hfill{\scriptsize Correct}},
        fonttitle=\sffamily\bfseries\small,
        fontupper=\footnotesize\sffamily,
        toptitle=1.5pt,
        bottomtitle=1.5pt,
      ]
      \setlength{\parskip}{0.32em}\setlength{\parindent}{0pt}#2
    \end{tcolorbox}
  \end{tcbraster}%
}

\section{Final-Response Comparison}
\label{app:response-compare}

This section compares the \emph{final reply} of WeAgent‑MMSearch‑RL and Kimi~K2.6 with Hermes on three samples. On these samples, WeAgent‑MMSearch‑RL is correct, and
Kimi~K2.6 with Hermes is not. Images cited in a reply are shown in that
column.

\begin{tcolorbox}[
  enhanced,
  colback=white,
  colframe=wehead,
  boxrule=0.7pt,
  arc=2.2mm,
  left=5pt,
  right=5pt,
  top=4pt,
  bottom=5pt,
  colbacktitle=wehead,
  coltitle=white,
  fonttitle=\sffamily\bfseries,
  title={Case 1\quad Membrane color in a Class~II cytokine-receptor schematic},
  before skip=0pt,
  after skip=0pt,
]
\respquery{This scholar has long taught immunology at the Catholic University of
Leuven in Belgium, is also affiliated with an international cancer research
institution, and is known for research on cytokines and cancer immunology.
In an article the scholar wrote for a leading immunology review journal in
2003, a schematic systematically depicting multiple members of receptor
families involved in antiviral and inflammatory regulation and the structures
of their transmembrane complexes shows the portions of the receptor complexes
embedded in the cell membrane \hlquery{mainly in what color?}}
\resptarget{0.42\linewidth}{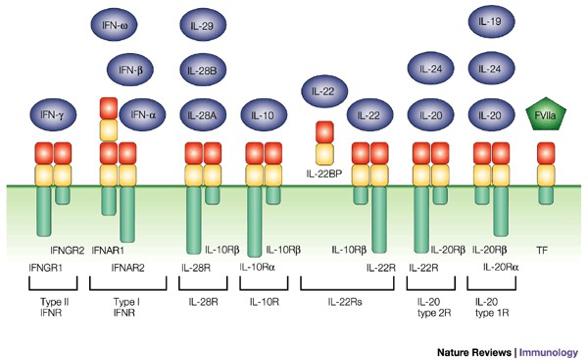}
\respcompare
{The membrane-embedded portions are shown in {\bfseries \hlans{gray}}.}
{Based on my research, the scholar is {\bfseries \hlans{Jean-Christophe Renauld}}, a
professor at the Catholic University of Leuven (KU~Leuven) in Belgium and a
researcher at the Ludwig Institute for Cancer Research. In 2003, he published
a review article titled ``Class~II cytokine receptors and their ligands: key
antiviral and inflammatory modulators'' in \emph{Nature Reviews Immunology}.

The schematic systematically depicts multiple members of Class~II cytokine
receptor families involved in antiviral and inflammatory regulation, together
with the structures of their transmembrane complexes. In this figure, the
portions of the receptor complexes embedded in the cell membrane are shown in
{\bfseries \hlans{green}} (or a greenish/teal color).

The figure includes receptors such as IFNGR2/IFNAR1, IFNAR2, IL-28RA,
IL-10R$\beta$, IL-10R$\alpha$, IL-22R, IL-20R$\beta$, IL-20R$\alpha$, and TF.
Extracellular domains are drawn in blue/purple, whereas the transmembrane
segments that cross the membrane bar are colored green.
\respcited{images/response_cases/case_0094_cited.jpg}{Retrieved receptor schematic}}
\end{tcolorbox}

\clearpage
\begin{tcolorbox}[
  enhanced,
  colback=white,
  colframe=wehead,
  boxrule=0.7pt,
  arc=2.2mm,
  left=5pt,
  right=5pt,
  top=4pt,
  bottom=5pt,
  colbacktitle=wehead,
  coltitle=white,
  fonttitle=\sffamily\bfseries,
  title={Case 2\quad Next Chinese equivalent after \emph{ma} in a Shui-script chart},
  before skip=0pt,
  after skip=0pt,
]
\respquery{In the Wanhu Shuizhai community on the outskirts of the county seat
of China's only Shui autonomous county, there is a museum that is the
country's first specialized exhibition space devoted to the culture of this
ethnic minority. On its first floor, a migration-route map and classical texts
show the development of an ancient script. A reference page for this script on
the Omniglot website contains a comparison chart. What is the next character
\hlquery{after \emph{ma} (`horse')} in the chart's third column?}
\resptarget{0.40\linewidth}{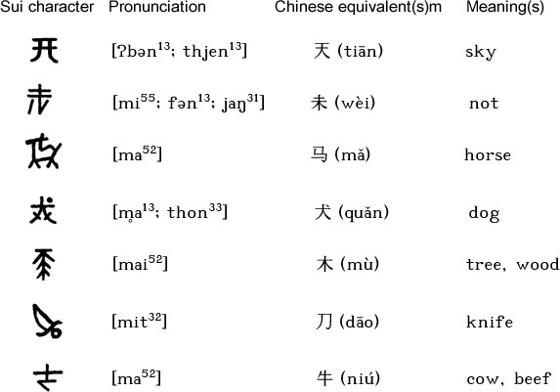}
\respcompare
{The next character is {\bfseries \emph{\hlans{yang}}} (`sheep').}
{Based on my investigation, the museum is the {\bfseries Guizhou Shui Culture
Museum}, located in Wanhu Shuizhai community, Fengyu Subdistrict, Sandu Shui
Autonomous County, Qiannan Buyei and Miao Autonomous Prefecture, Guizhou
Province. It is China's first museum devoted specifically to Shui culture.

The first floor displays a Shui migration-route map and classical texts that
present the history of the ancient Shui script (\emph{Sui} script).

The Omniglot reference page contains a comparison chart. In the third column
(Chinese equivalents) the sequence is
\emph{tian} (sky),
\emph{wei} (not),
\emph{ma} (horse),
{\bfseries \emph{\hlans{quan}} (dog)},
\emph{mu} (tree/wood),
\emph{dao} (knife),
\emph{niu} (cow).

Therefore, the next character after \emph{ma} is {\bfseries \emph{\hlans{quan}}
(`dog')}.
\respcited{images/response_cases/case_0117_cited.jpg}{Omniglot comparison chart}}
\end{tcolorbox}

\clearpage
\begin{tcolorbox}[
  enhanced,
  colback=white,
  colframe=wehead,
  boxrule=0.7pt,
  arc=2.2mm,
  left=5pt,
  right=5pt,
  top=4pt,
  bottom=5pt,
  colbacktitle=wehead,
  coltitle=white,
  fonttitle=\sffamily\bfseries,
  title={Case 3\quad Vertical text and rate letter on a 2015 Canada Picture Postage},
  before skip=0pt,
  after skip=0pt,
]
\respquery{A North American country and an East Asian country jointly issued a
se-tenant postal item to commemorate the 35th anniversary of their diplomatic
relations; it featured two kinds of big cat drawn separately by designers from
the two countries. Exactly ten years after this se-tenant postal item featuring
the two big cats was issued, the animal on the right was again selected by
that country's postal service as a postage design, this time shown as an
actual photograph on personalized postage. What fixed text is printed
vertically \hlquery{along the left edge of the postage?} What letter appears in the
\hlquery{postage symbol in the lower right?}}
\resptarget{0.28\linewidth}{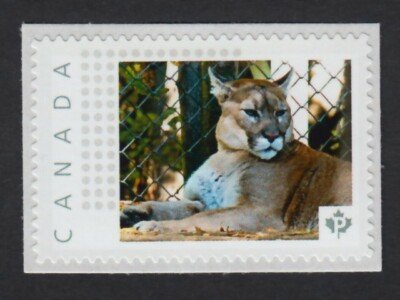}
\respcompare
{No final answer was produced. After a long search, the model hypothesized
that a 2015 Canada Post Picture Postage of a cougar might use a frame with
vertical ``CANADA'' and a Permanent ``P'' mark, then issued another web-search
call about frame designs and \hlans{stopped without boxing a response.}}
{The se-tenant item is the 2005 Canada--China joint issue commemorating the
35th anniversary of diplomatic relations, featuring a Canadian cougar (on the
right) and a Chinese leopard. Exactly ten years later, in 2015, Canada Post
issued a personalized Picture Postage stamp with a photograph of a cougar.

On this personalized postage stamp, the fixed text printed vertically along
the left edge is {\bfseries \hlans{CANADA}}, and the postage symbol in the
lower-right corner is a maple leaf containing the letter {\bfseries \hlans{P}}
(Permanent domestic rate).
\respcited{images/response_cases/case_0130_cited.jpg}{2015 Canada Picture Postage}}
\end{tcolorbox}

\clearpage

% Generated qualitative RL rollouts.
\definecolor{thinkbg}{HTML}{EEF3FA}
\definecolor{thinkhead}{HTML}{4A6FA5}
\definecolor{callbg}{HTML}{FFF6E0}
\definecolor{callhead}{HTML}{C4922A}
\definecolor{obsbg}{HTML}{E8F3EC}
\definecolor{obshead}{HTML}{2F7A5D}

\section{Qualitative RL Response Cases}
\label{app:rl-cases}

This appendix shows three WeAgent-MMSearch-RL rollouts. Tool
returns are shown as a short preview: at most three text hits, and at most
five images. Source URLs keep only the host and redact the path. The full payload
is omitted. The target image is held out at query time.

\phantomsection\label{app:rl-case-1}
\noindent{\sffamily\bfseries\large\color{wehead} Case 1: Cross-event athlete retrieval and suit-color reading}\par
\vspace{2pt}
\respquery{In the same event in which the athlete shown in this photograph competed, what were the two main colors of the racing suit worn by the host nation's best-performing athlete in a news photograph showing that athlete celebrating after crossing the finish line?}
{\small\sffamily\itshape Reference answer: Light blue and dark purple (blue and purple).}\par
\vspace{2pt}
\noindent\begin{minipage}[t]{0.48\linewidth}\centering
{\sffamily\bfseries\small\color{wehead} Input image}\\[-1pt]
{\sffamily\scriptsize\color{black!50} (provided with the query)}\\[3pt]
\includegraphics[width=0.92\linewidth,height=0.15\textheight,keepaspectratio]{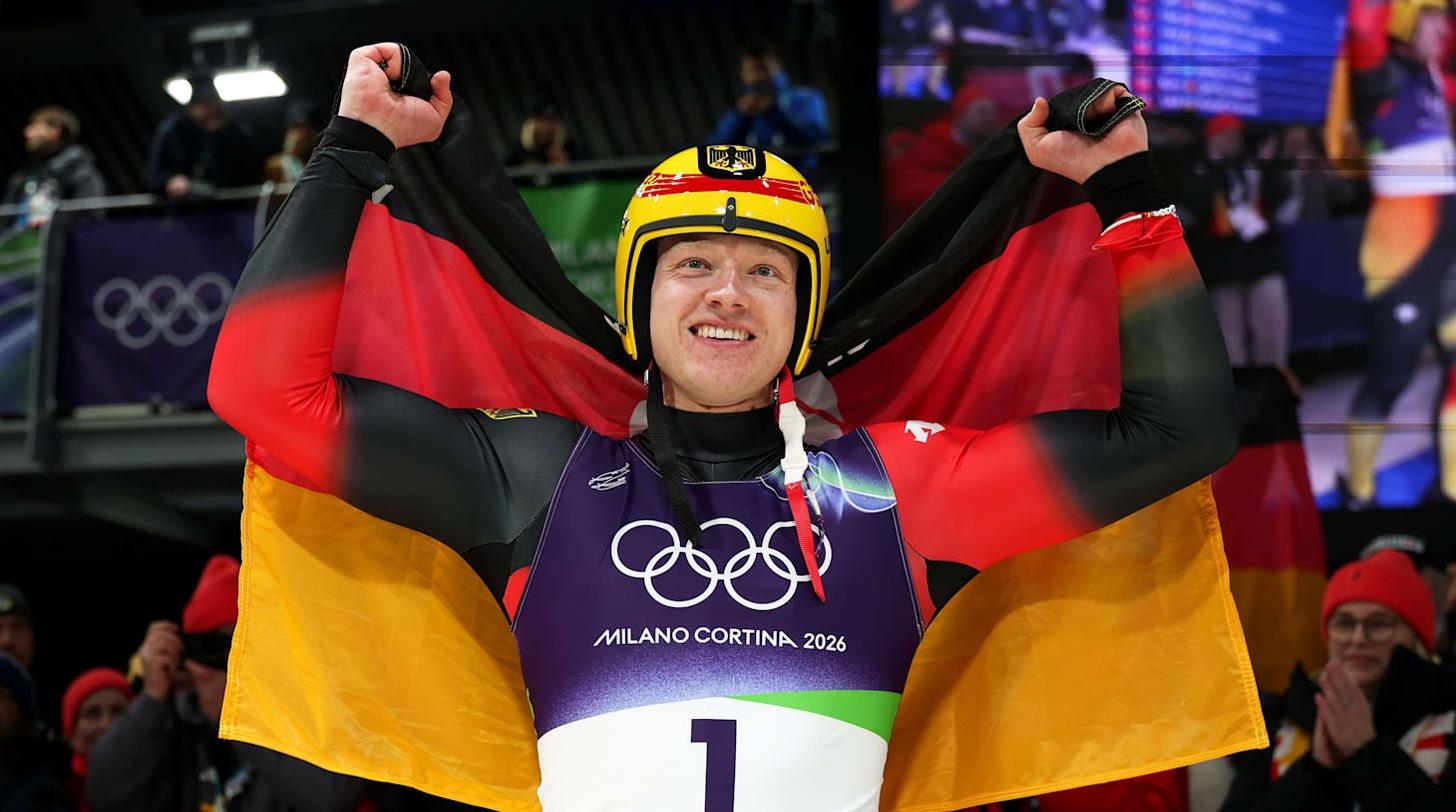}
\end{minipage}\hfill
\begin{minipage}[t]{0.48\linewidth}\centering
{\sffamily\bfseries\small\color{wehead} Target image}\\[-1pt]
{\sffamily\scriptsize\color{black!50} (not provided to the model at query time)}\\[3pt]
\includegraphics[width=0.92\linewidth,height=0.15\textheight,keepaspectratio]{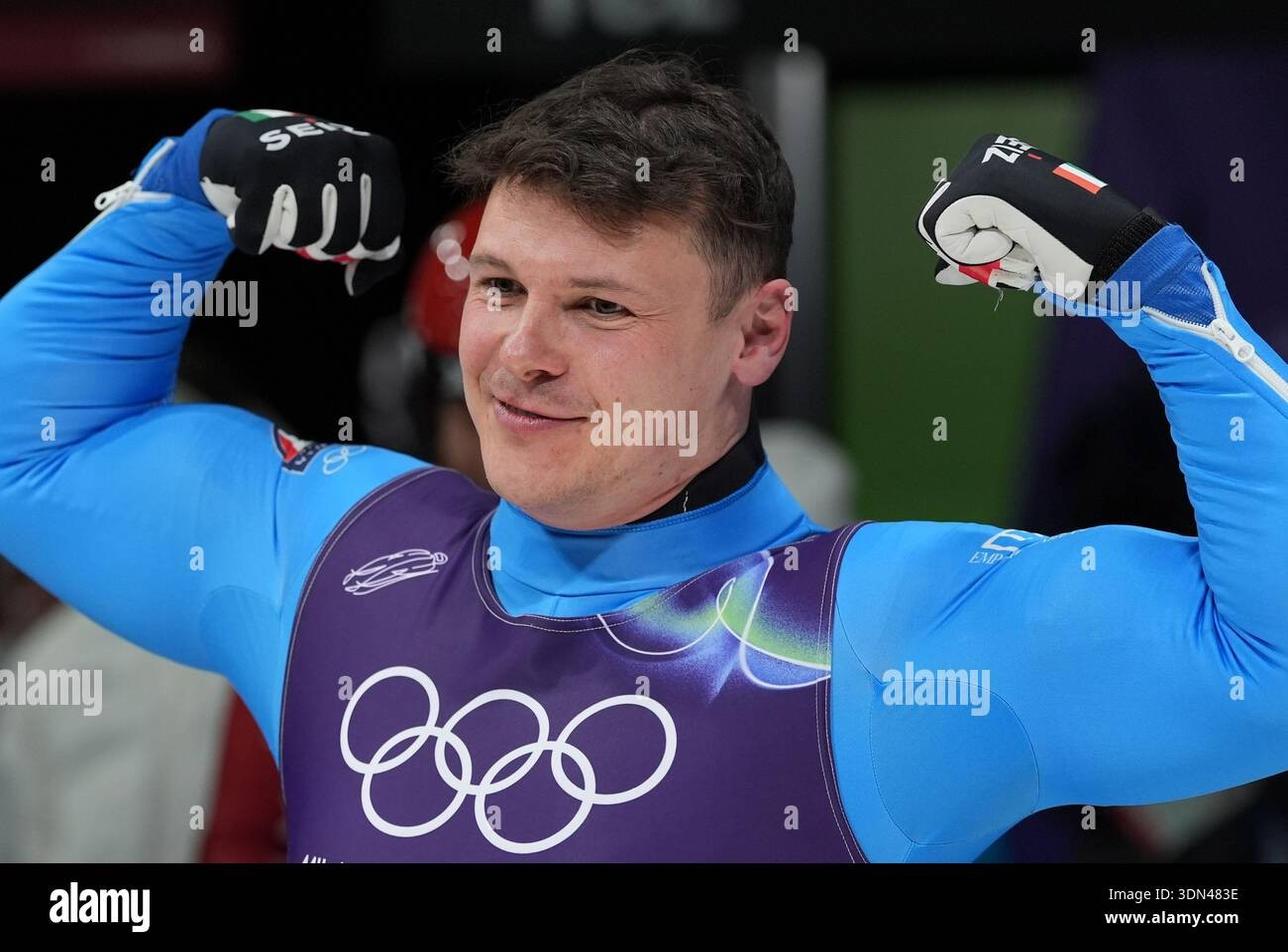}
\end{minipage}\par
\vspace{2pt}
\begin{tcolorbox}[
  colback=thinkbg, colframe=thinkhead, colbacktitle=thinkhead, coltitle=white,
  boxrule=0.45pt, arc=1.4mm, left=5pt,right=5pt,top=4pt,bottom=4pt,
  toptitle=1.2pt,bottomtitle=1.2pt, fonttitle=\sffamily\bfseries\small,
  fontupper=\footnotesize\sffamily, breakable,
  title={Think $|$ Round 1}, before skip=3.5pt, after skip=3.5pt,
]
We need to answer: In the same event in which the athlete shown in this photograph competed, what were the two main colors of the racing suit worn by the host nation's best-performing athlete in a news photograph showing that athlete celebrating after crossing the finish line?\par We have an image: user\_image\_1.png. We need to identify the athlete in the image, the event, then find the host nation's best-performing athlete in that event, and then find a news photograph of that athlete celebrating after crossing the finish line, and determine the two main colors of the racing suit worn by that athlete ...
\end{tcolorbox}
\begin{tcolorbox}[
  colback=callbg, colframe=callhead, colbacktitle=callhead, coltitle=white,
  boxrule=0.45pt, arc=1.4mm, left=5pt,right=5pt,top=4pt,bottom=4pt,
  toptitle=1.2pt,bottomtitle=1.2pt, fonttitle=\sffamily\bfseries\small,
  fontupper=\footnotesize\sffamily, breakable,
  title={Tool call $|$ web\_search\_based\_on\_image $|$ Round 1}, before skip=3.5pt, after skip=3.5pt,
]
{\ttfamily\small \{"image": "user\_image\_1.png", "num": 5\}}
\end{tcolorbox}
\begin{tcolorbox}[
  colback=obsbg, colframe=obshead, colbacktitle=obshead, coltitle=white,
  boxrule=0.45pt, arc=1.4mm, left=5pt,right=5pt,top=4pt,bottom=4pt,
  toptitle=1.2pt,bottomtitle=1.2pt, fonttitle=\sffamily\bfseries\small,
  fontupper=\footnotesize\sffamily, breakable,
  title={Tool response $|$ Round 1}, before skip=3.5pt, after skip=3.5pt,
]
{\sffamily\bfseries Summary preview}
\par{\sffamily\color{thinkhead}\bfseries Langenhan races to luge gold}
\par{\scriptsize\ttfamily\color{black!55} https://www.picture-alliance.com/***}
\par{\sffamily\color{thinkhead}\bfseries Olympia 2026: Die Winterspiele von Mailand und Cortina d'Ampezzo: Kuhne-Hellmessen, Ulrich, Ve ...}
\par{\scriptsize\ttfamily\color{black!55} https://www.amazon.ca/***}
\par{\sffamily\color{thinkhead}\bfseries Lieber Herr Merz - es tut mir super leid\textbackslash{}}
\par{\scriptsize\ttfamily\color{black!55} https://www.stern.de/***}
\par\vspace{3pt}\noindent \begin{minipage}[t]{0.16\linewidth}\centering\includegraphics[width=\linewidth,height=0.11\textheight,keepaspectratio]{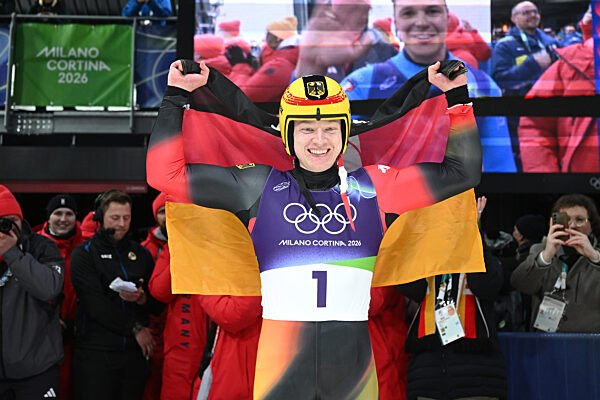}\\{\scriptsize\ttfamily search\_5785fff5.jpg}\end{minipage}\hfill \begin{minipage}[t]{0.16\linewidth}\centering\includegraphics[width=\linewidth,height=0.11\textheight,keepaspectratio]{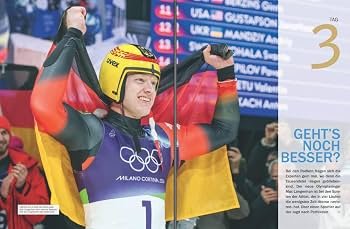}\\{\scriptsize\ttfamily search\_af8aa6c4.jpg}\end{minipage}\hfill \begin{minipage}[t]{0.16\linewidth}\centering\includegraphics[width=\linewidth,height=0.11\textheight,keepaspectratio]{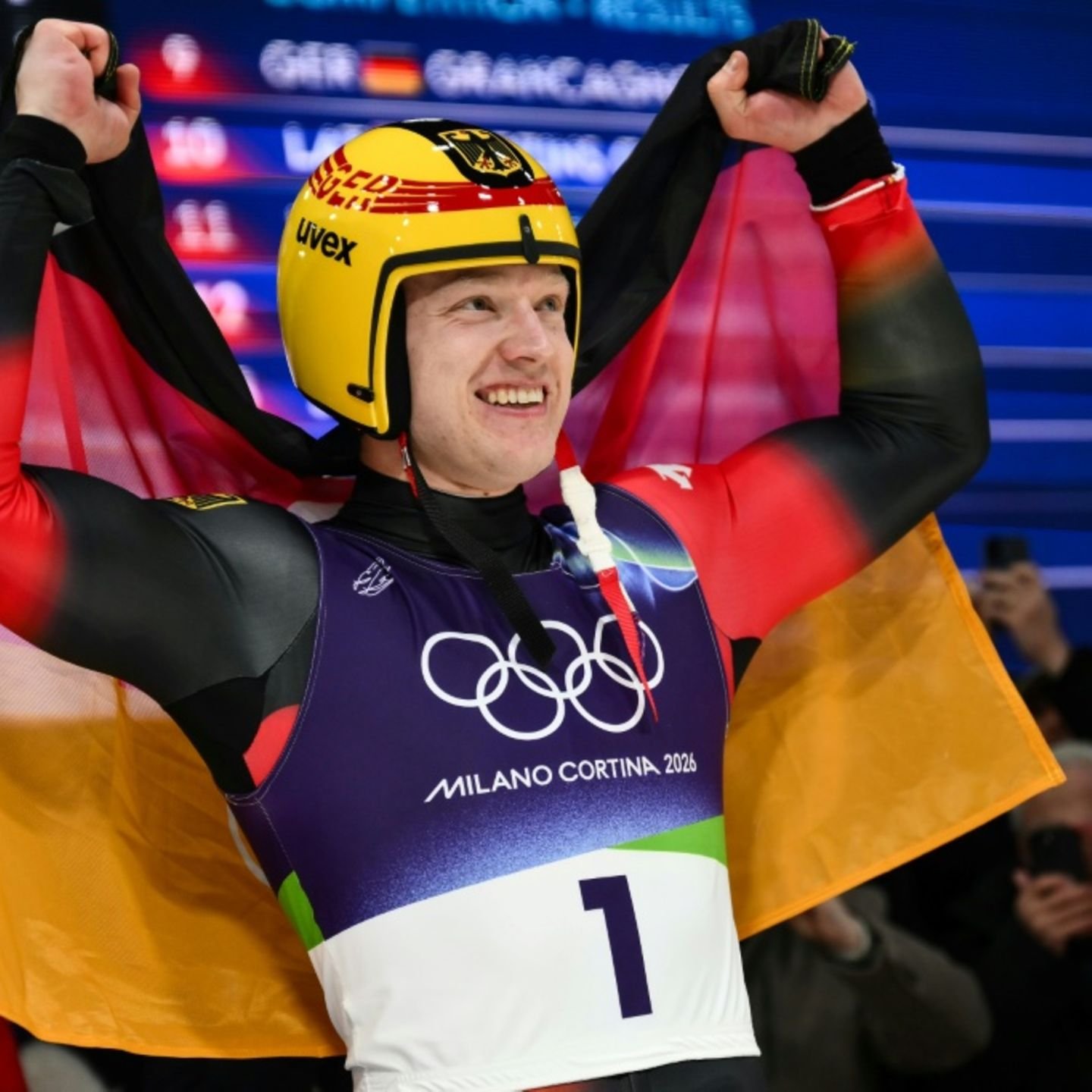}\\{\scriptsize\ttfamily search\_993575f1.jpg}\end{minipage}\hfill \begin{minipage}[t]{0.16\linewidth}\centering\includegraphics[width=\linewidth,height=0.11\textheight,keepaspectratio]{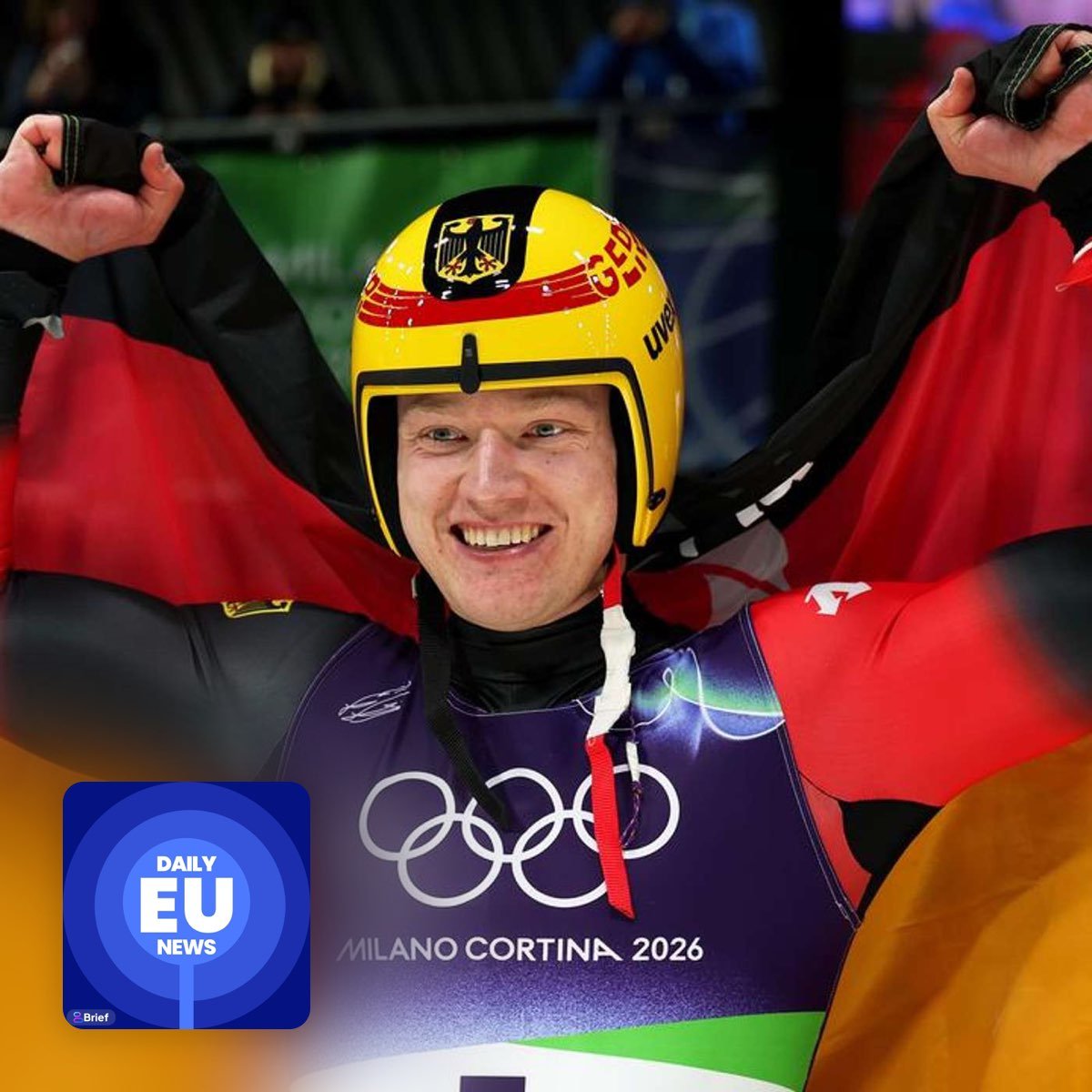}\\{\scriptsize\ttfamily search\_408aa347.jpg}\end{minipage}\hfill \begin{minipage}[t]{0.16\linewidth}\centering\includegraphics[width=\linewidth,height=0.11\textheight,keepaspectratio]{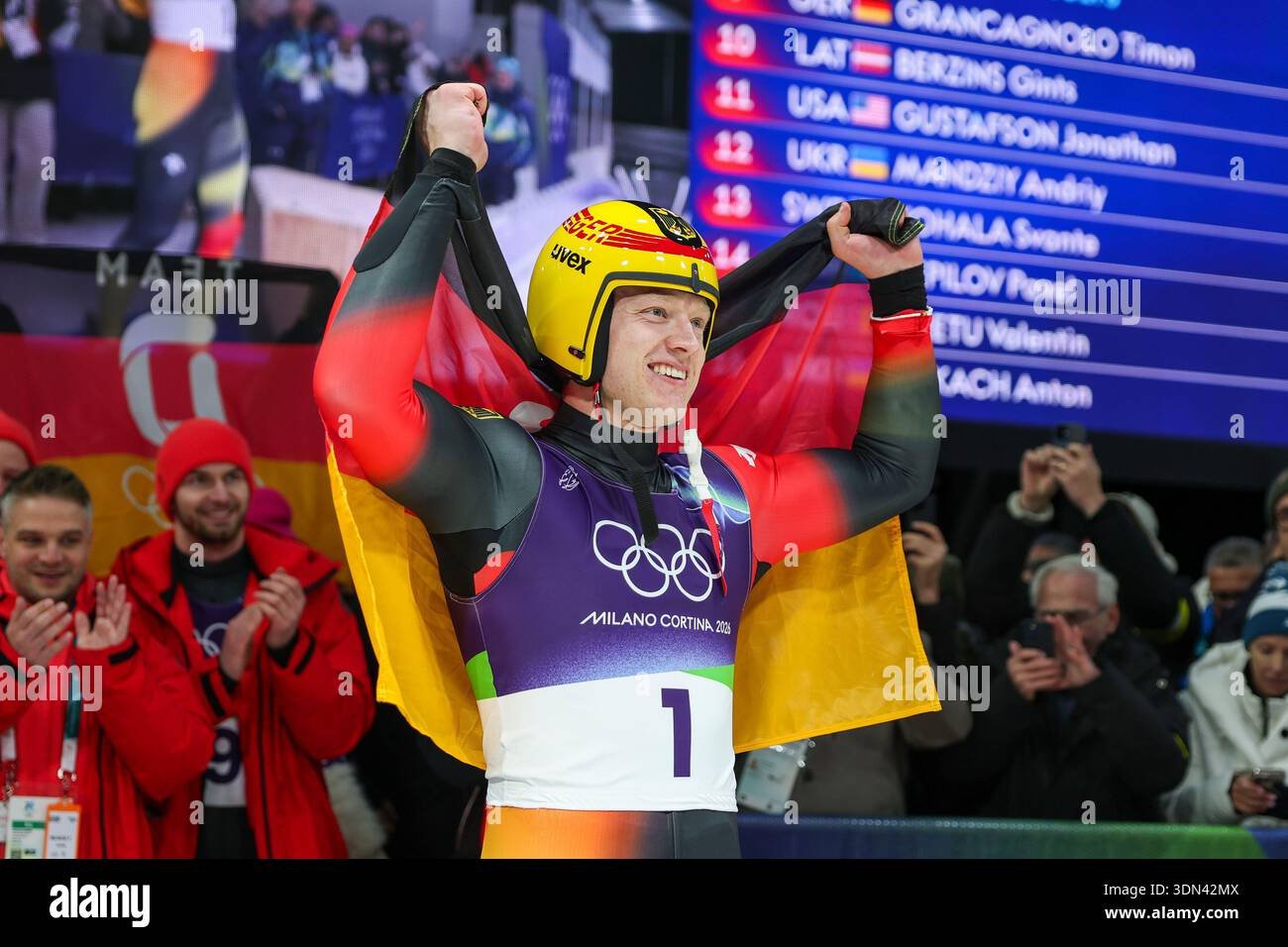}\\{\scriptsize\ttfamily search\_1a9c7c46.jpg}\end{minipage}\par\vspace{2pt}{\scriptsize\sffamily\color{black!50} $\triangleright$ Full tool return (2,743 characters)}
\end{tcolorbox}
\begin{tcolorbox}[
  colback=thinkbg, colframe=thinkhead, colbacktitle=thinkhead, coltitle=white,
  boxrule=0.45pt, arc=1.4mm, left=5pt,right=5pt,top=4pt,bottom=4pt,
  toptitle=1.2pt,bottomtitle=1.2pt, fonttitle=\sffamily\bfseries\small,
  fontupper=\footnotesize\sffamily, breakable,
  title={Think $|$ Round 2}, before skip=3.5pt, after skip=3.5pt,
]
The reverse image search results identify the athlete as Max Langenhan, a German luger who won gold at the Milano Cortina 2026 Winter Olympics in the men's singles luge event on February 8, 2026.\par The event is the Milano Cortina 2026 Winter Olympics (host nation: Italy). The host nation's best-performing athlete in this event? Wait, the event is luge men's singles. The host nation is Italy. What was Italy's best result in men's singles luge at Milano Cortina 2026? We need to find the host nation's best-performing athlete in the same event (men's singles luge). The host nation is Italy. We need to ...
\end{tcolorbox}
\begin{tcolorbox}[
  colback=callbg, colframe=callhead, colbacktitle=callhead, coltitle=white,
  boxrule=0.45pt, arc=1.4mm, left=5pt,right=5pt,top=4pt,bottom=4pt,
  toptitle=1.2pt,bottomtitle=1.2pt, fonttitle=\sffamily\bfseries\small,
  fontupper=\footnotesize\sffamily, breakable,
  title={Tool call $|$ web\_search $|$ Round 2}, before skip=3.5pt, after skip=3.5pt,
]
{\ttfamily\small \{"query": ["Milano Cortina 2026 luge men's singles Italy best result finish line celebration", "Italian luger Milano Cortina 2026 luge gold medal", "campioni olimpici Milano Cortina 2026 luge singolo maschile", "luge Milano Cortina 2026 Italia finish line celebrare"]\}}
\end{tcolorbox}
\begin{tcolorbox}[
  colback=obsbg, colframe=obshead, colbacktitle=obshead, coltitle=white,
  boxrule=0.45pt, arc=1.4mm, left=5pt,right=5pt,top=4pt,bottom=4pt,
  toptitle=1.2pt,bottomtitle=1.2pt, fonttitle=\sffamily\bfseries\small,
  fontupper=\footnotesize\sffamily, breakable,
  title={Tool response $|$ Round 2}, before skip=3.5pt, after skip=3.5pt,
]
{\sffamily\bfseries Summary preview}
\par{\sffamily\color{thinkhead}\bfseries An Olympic champion is crowned in men's singles luge ...}
\par{\scriptsize\ttfamily\color{black!55} https://www.facebook.com/***}
\par{\sffamily\color{thinkhead}\bfseries Olympic luge at Milan Cortina 2026: Germany dominates ...}
\par{\scriptsize\ttfamily\color{black!55} https://www.nbcolympics.com/***}
\par{\sffamily\color{thinkhead}\bfseries Focus on Cortina: Olympic stage for Luge in ...}
\par{\scriptsize\ttfamily\color{black!55} https://www.fil-luge.org/***}
\par\vspace{2pt}{\scriptsize\sffamily\color{black!50} $\triangleright$ Full tool return (5,910 characters)}
\end{tcolorbox}
\begin{tcolorbox}[
  colback=thinkbg, colframe=thinkhead, colbacktitle=thinkhead, coltitle=white,
  boxrule=0.45pt, arc=1.4mm, left=5pt,right=5pt,top=4pt,bottom=4pt,
  toptitle=1.2pt,bottomtitle=1.2pt, fonttitle=\sffamily\bfseries\small,
  fontupper=\footnotesize\sffamily, breakable,
  title={Think $|$ Round 3}, before skip=3.5pt, after skip=3.5pt,
]
We need to find the host nation (Italy) best-performing athlete in the same event (men's singles luge). The search results mention Germany dominated, Italy got gold in doubles, but what about men's singles? Let's check the Olympics.com results page for luge results. Also, we need a news photograph showing that athlete celebrating after crossing the finish line. Let's extract the Olympics.com results page and the NBC article.
\end{tcolorbox}
\begin{tcolorbox}[
  colback=callbg, colframe=callhead, colbacktitle=callhead, coltitle=white,
  boxrule=0.45pt, arc=1.4mm, left=5pt,right=5pt,top=4pt,bottom=4pt,
  toptitle=1.2pt,bottomtitle=1.2pt, fonttitle=\sffamily\bfseries\small,
  fontupper=\footnotesize\sffamily, breakable,
  title={Tool call $|$ web\_extract $|$ Round 3}, before skip=3.5pt, after skip=3.5pt,
]
{\ttfamily\small \{"urls": ["https://www.olympics.com/***", "https://www.nbcolympics.com/***"], "goal": "Find the results for men's singles luge at Milano Cortina 2026, spec ...}
\end{tcolorbox}
\begin{tcolorbox}[
  colback=obsbg, colframe=obshead, colbacktitle=obshead, coltitle=white,
  boxrule=0.45pt, arc=1.4mm, left=5pt,right=5pt,top=4pt,bottom=4pt,
  toptitle=1.2pt,bottomtitle=1.2pt, fonttitle=\sffamily\bfseries\small,
  fontupper=\footnotesize\sffamily, breakable,
  title={Tool response $|$ Round 3}, before skip=3.5pt, after skip=3.5pt,
]
{\sffamily\bfseries Summary preview}
\par{\sffamily\color{thinkhead}\bfseries Milano-Cortina 2026 Luge - Olympic Results by Discipline}
\par{\scriptsize\ttfamily\color{black!55} https://www.olympics.com/***}
\par{\sffamily\color{thinkhead}\bfseries Olympic luge at Milan Cortina 2026: Germany dominates, Farquharson gives U.S. its medal moment}
\par{\scriptsize\ttfamily\color{black!55} https://www.nbcolympics.com/***}
\par\vspace{3pt}\noindent \begin{minipage}[t]{0.16\linewidth}\centering\includegraphics[width=\linewidth,height=0.11\textheight,keepaspectratio]{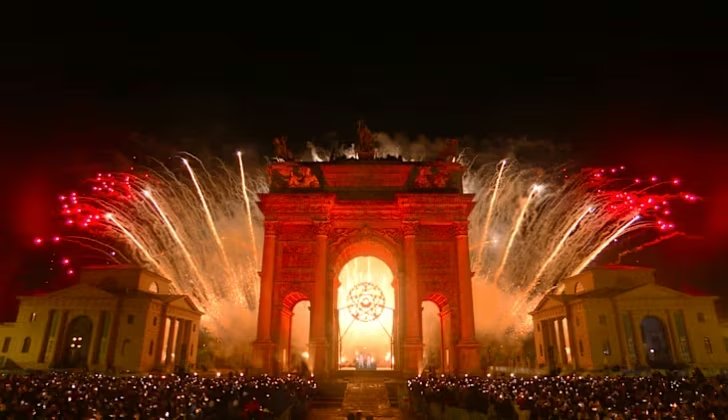}\\{\scriptsize\ttfamily extract\_73f08767.jpg}\end{minipage}\hfill \begin{minipage}[t]{0.16\linewidth}\centering\includegraphics[width=\linewidth,height=0.11\textheight,keepaspectratio]{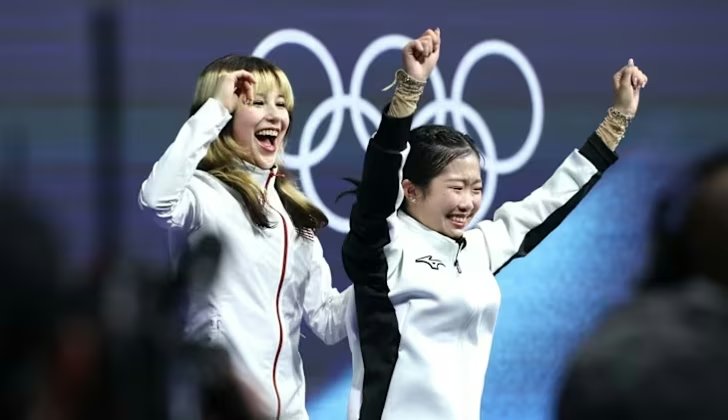}\\{\scriptsize\ttfamily extract\_0182f6e2.jpg}\end{minipage}\hfill \begin{minipage}[t]{0.16\linewidth}\centering\includegraphics[width=\linewidth,height=0.11\textheight,keepaspectratio]{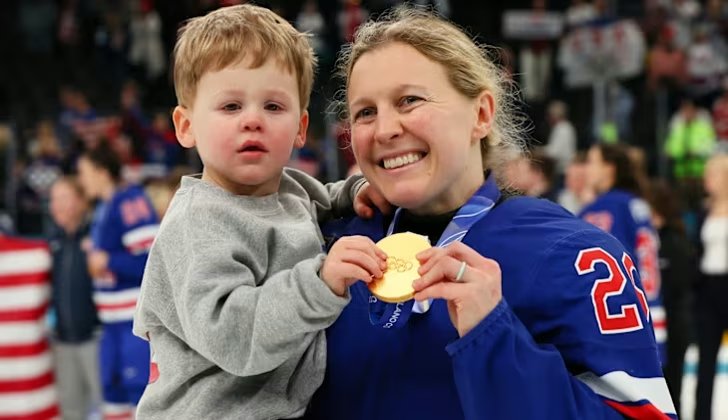}\\{\scriptsize\ttfamily extract\_2a0e719b.jpg}\end{minipage}\hfill \begin{minipage}[t]{0.16\linewidth}\centering\includegraphics[width=\linewidth,height=0.11\textheight,keepaspectratio]{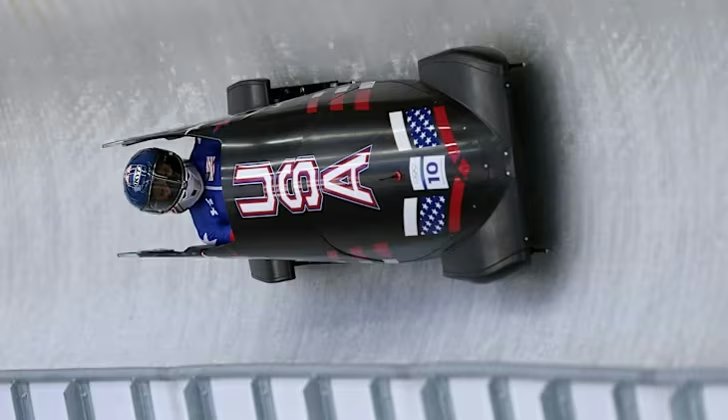}\\{\scriptsize\ttfamily extract\_420cb9c5.jpg}\end{minipage}\hfill \begin{minipage}[t]{0.16\linewidth}\centering\includegraphics[width=\linewidth,height=0.11\textheight,keepaspectratio]{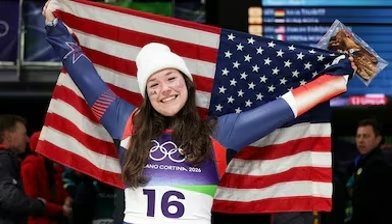}\\{\scriptsize\ttfamily extract\_077c4bf3.jpg}\end{minipage}\par\vspace{2pt}{\scriptsize\sffamily\color{black!50} $\triangleright$ Full tool return (3,102 characters)}
\end{tcolorbox}
\begin{tcolorbox}[
  colback=thinkbg, colframe=thinkhead, colbacktitle=thinkhead, coltitle=white,
  boxrule=0.45pt, arc=1.4mm, left=5pt,right=5pt,top=4pt,bottom=4pt,
  toptitle=1.2pt,bottomtitle=1.2pt, fonttitle=\sffamily\bfseries\small,
  fontupper=\footnotesize\sffamily, breakable,
  title={Think $|$ Round 4}, before skip=3.5pt, after skip=3.5pt,
]
The NBC article reveals that Italy's Dominik Fischnaller won bronze in men's singles luge at Milano Cortina 2026. He was the best-performing Italian in men's singles luge. We need a news photograph showing Dominik Fischnaller celebrating after crossing the finish line, and we need to determine the two main colors of the racing suit worn by him in that photo.\par Let's search for "Dominik Fischnaller Milano Cortina 2026 finish line celebration" or similar.
\end{tcolorbox}
\begin{tcolorbox}[
  colback=callbg, colframe=callhead, colbacktitle=callhead, coltitle=white,
  boxrule=0.45pt, arc=1.4mm, left=5pt,right=5pt,top=4pt,bottom=4pt,
  toptitle=1.2pt,bottomtitle=1.2pt, fonttitle=\sffamily\bfseries\small,
  fontupper=\footnotesize\sffamily, breakable,
  title={Tool call $|$ web\_search $|$ Round 4}, before skip=3.5pt, after skip=3.5pt,
]
{\ttfamily\small \{"query": ["Dominik Fischnaller Milano Cortina 2026 finish line celebration", "Dominik Fischnaller luge 2026 Italy finish line", "Dominik Fischnaller finish line photo 2026", "Luge Dominik Fischnaller Milano Cortina 2026"]\}}
\end{tcolorbox}
\begin{tcolorbox}[
  colback=obsbg, colframe=obshead, colbacktitle=obshead, coltitle=white,
  boxrule=0.45pt, arc=1.4mm, left=5pt,right=5pt,top=4pt,bottom=4pt,
  toptitle=1.2pt,bottomtitle=1.2pt, fonttitle=\sffamily\bfseries\small,
  fontupper=\footnotesize\sffamily, breakable,
  title={Tool response $|$ Round 4}, before skip=3.5pt, after skip=3.5pt,
]
{\sffamily\bfseries Summary preview}
\par{\sffamily\color{thinkhead}\bfseries Dominik Fischnaller races to bronze  at the Winter ...}
\par{\scriptsize\ttfamily\color{black!55} https://www.facebook.com/***}
\par{\sffamily\color{thinkhead}\bfseries Bravo, Dominik! Our luging star from Meransen ...}
\par{\scriptsize\ttfamily\color{black!55} https://www.instagram.com/***}
\par{\sffamily\color{thinkhead}\bfseries Dominik Fischnaller and the long road to Olympic dreams}
\par{\scriptsize\ttfamily\color{black!55} https://www.fil-luge.org/***}
\par\vspace{2pt}{\scriptsize\sffamily\color{black!50} $\triangleright$ Full tool return (5,641 characters)}
\end{tcolorbox}
\begin{tcolorbox}[
  colback=thinkbg, colframe=thinkhead, colbacktitle=thinkhead, coltitle=white,
  boxrule=0.45pt, arc=1.4mm, left=5pt,right=5pt,top=4pt,bottom=4pt,
  toptitle=1.2pt,bottomtitle=1.2pt, fonttitle=\sffamily\bfseries\small,
  fontupper=\footnotesize\sffamily, breakable,
  title={Think $|$ Round 5}, before skip=3.5pt, after skip=3.5pt,
]
We have a Facebook post from FILuge titled "Dominik Fischnaller races to bronze  at the Winter Olympics in Milano Cortina 2026!" which might contain a photo of him celebrating after crossing the finish line. Let's extract that page. Also, we might need to search for images of Dominik Fischnaller at the finish line. Let's extract the FILuge Facebook post and maybe the Olympics.com athlete page. Also, check Getty Images or other sources. Let's start with the Facebook post.
\end{tcolorbox}
\begin{tcolorbox}[
  colback=callbg, colframe=callhead, colbacktitle=callhead, coltitle=white,
  boxrule=0.45pt, arc=1.4mm, left=5pt,right=5pt,top=4pt,bottom=4pt,
  toptitle=1.2pt,bottomtitle=1.2pt, fonttitle=\sffamily\bfseries\small,
  fontupper=\footnotesize\sffamily, breakable,
  title={Tool call $|$ web\_extract $|$ Round 5}, before skip=3.5pt, after skip=3.5pt,
]
{\ttfamily\small \{"urls": ["https://www.facebook.com/***", "https://www.olympics.com/***", "https://en.wikipedia.org/***"], "goal": "Find a news photograph showing Dom ...}
\end{tcolorbox}
\begin{tcolorbox}[
  colback=obsbg, colframe=obshead, colbacktitle=obshead, coltitle=white,
  boxrule=0.45pt, arc=1.4mm, left=5pt,right=5pt,top=4pt,bottom=4pt,
  toptitle=1.2pt,bottomtitle=1.2pt, fonttitle=\sffamily\bfseries\small,
  fontupper=\footnotesize\sffamily, breakable,
  title={Tool response $|$ Round 5}, before skip=3.5pt, after skip=3.5pt,
]
{\sffamily\bfseries Summary preview}
\par{\sffamily\color{thinkhead}\bfseries International Luge Federation}
\par{\scriptsize\ttfamily\color{black!55} https://www.facebook.com/***}
\par{\sffamily\color{thinkhead}\bfseries Dominik FISCHNALLER}
\par{\scriptsize\ttfamily\color{black!55} https://www.olympics.com/***}
\par{\sffamily\color{thinkhead}\bfseries Dominik Fischnaller}
\par{\scriptsize\ttfamily\color{black!55} https://en.wikipedia.org/***}
\par\vspace{3pt}\noindent \begin{minipage}[t]{0.22\linewidth}\centering\includegraphics[width=\linewidth,height=0.11\textheight,keepaspectratio]{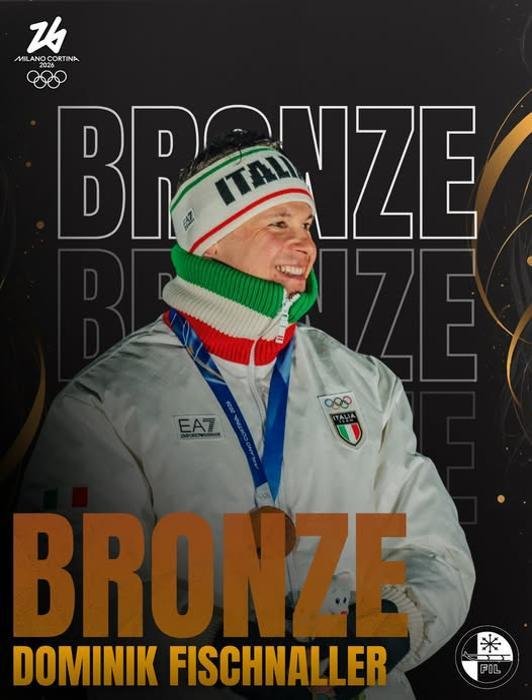}\\{\scriptsize\ttfamily extract\_2713a7e2.jpg}\end{minipage}\par\vspace{2pt}{\scriptsize\sffamily\color{black!50} $\triangleright$ Full tool return (3,590 characters)}
\end{tcolorbox}
\begin{tcolorbox}[
  colback=thinkbg, colframe=thinkhead, colbacktitle=thinkhead, coltitle=white,
  boxrule=0.45pt, arc=1.4mm, left=5pt,right=5pt,top=4pt,bottom=4pt,
  toptitle=1.2pt,bottomtitle=1.2pt, fonttitle=\sffamily\bfseries\small,
  fontupper=\footnotesize\sffamily, breakable,
  title={Think $|$ Round 6}, before skip=3.5pt, after skip=3.5pt,
]
The Facebook post images didn't show the actual photograph of Fischnaller celebrating at the finish line. The extracted images from the page were mostly emojis and a graphic. We need to locate the actual news photograph. Let's search for images of Dominik Fischnaller at the finish line. We can use web\_search\_for\_image with queries like "Dominik Fischnaller finish line Milano Cortina 2026" or "Dominik Fischnaller Milano Cortina 2026 luge". Let's try.
\end{tcolorbox}
\begin{tcolorbox}[
  colback=callbg, colframe=callhead, colbacktitle=callhead, coltitle=white,
  boxrule=0.45pt, arc=1.4mm, left=5pt,right=5pt,top=4pt,bottom=4pt,
  toptitle=1.2pt,bottomtitle=1.2pt, fonttitle=\sffamily\bfseries\small,
  fontupper=\footnotesize\sffamily, breakable,
  title={Tool call $|$ web\_search\_for\_image $|$ Round 6}, before skip=3.5pt, after skip=3.5pt,
]
{\ttfamily\small \{"query": ["Dominik Fischnaller finish line Milano Cortina 2026", "Dominik Fischnaller luge Milano Cortina 2026 finish line", "Dominik Fischnaller bronze medal Milano Cortina 2026 finish line"], "num": 10\}}
\end{tcolorbox}
\begin{tcolorbox}[
  colback=obsbg, colframe=obshead, colbacktitle=obshead, coltitle=white,
  boxrule=0.45pt, arc=1.4mm, left=5pt,right=5pt,top=4pt,bottom=4pt,
  toptitle=1.2pt,bottomtitle=1.2pt, fonttitle=\sffamily\bfseries\small,
  fontupper=\footnotesize\sffamily, breakable,
  title={Tool response $|$ Round 6}, before skip=3.5pt, after skip=3.5pt,
]
{\sffamily\bfseries Summary preview}
\par Tool returned a structured payload; only the preview is shown.
\par\vspace{3pt}\noindent \begin{minipage}[t]{0.16\linewidth}\centering\includegraphics[width=\linewidth,height=0.11\textheight,keepaspectratio]{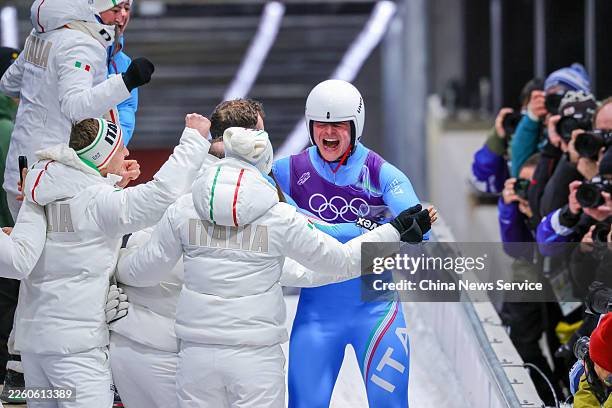}\\{\scriptsize\ttfamily search\_245c38a3.jpg}\end{minipage}\hfill \begin{minipage}[t]{0.16\linewidth}\centering\includegraphics[width=\linewidth,height=0.11\textheight,keepaspectratio]{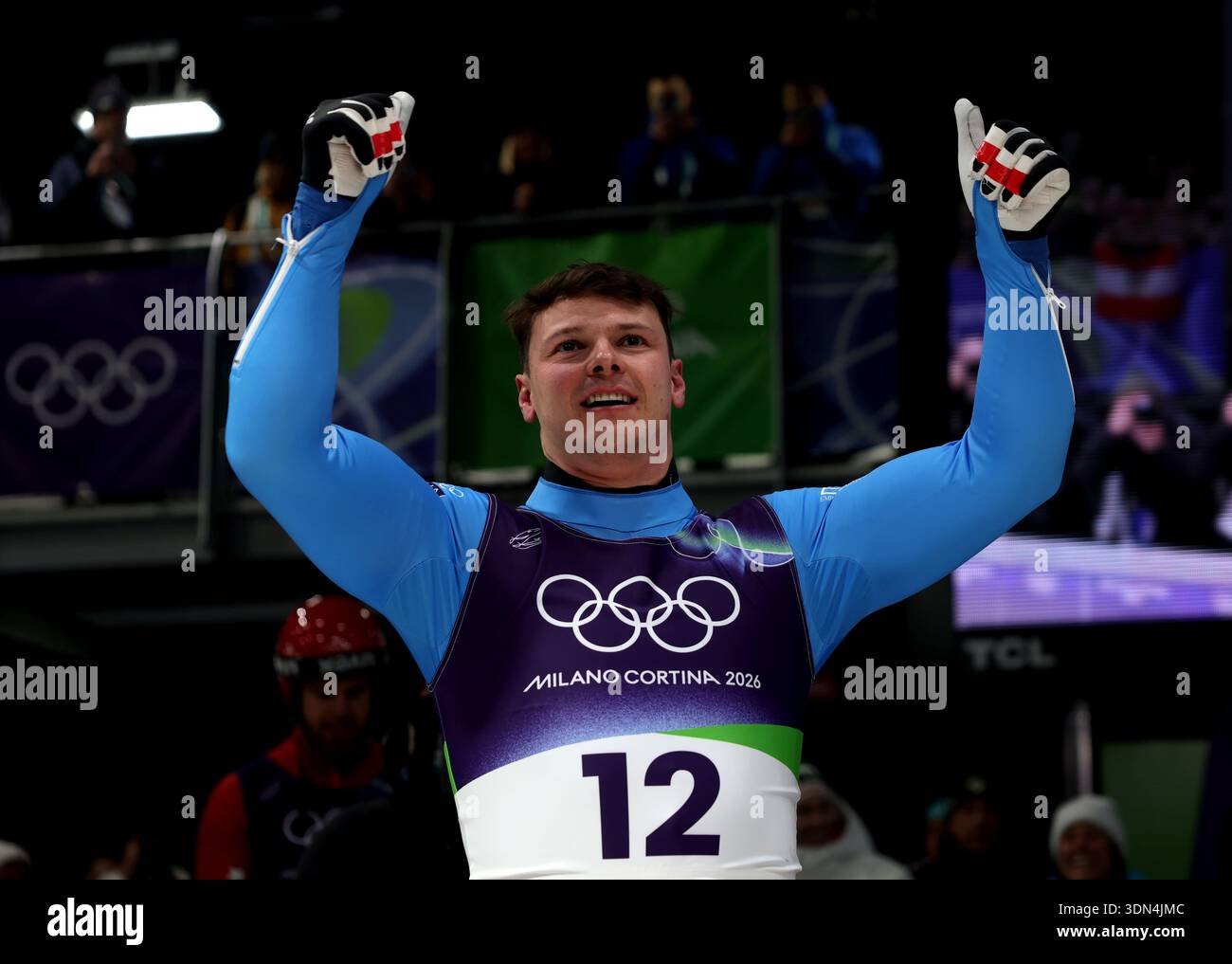}\\{\scriptsize\ttfamily search\_a0b48fcc.jpg}\end{minipage}\hfill \begin{minipage}[t]{0.16\linewidth}\centering\includegraphics[width=\linewidth,height=0.11\textheight,keepaspectratio]{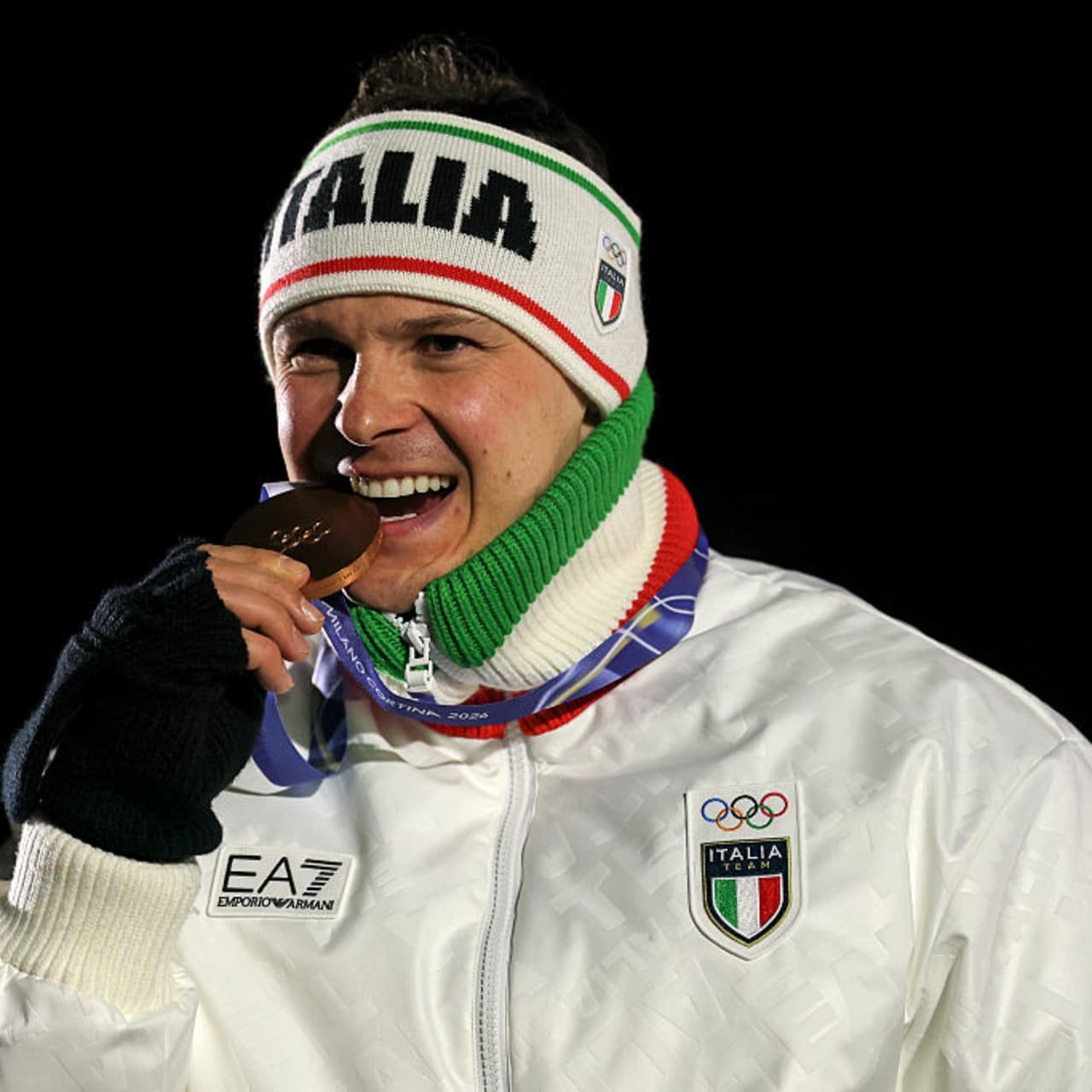}\\{\scriptsize\ttfamily search\_8be8ed1e.jpg}\end{minipage}\hfill \begin{minipage}[t]{0.16\linewidth}\centering\includegraphics[width=\linewidth,height=0.11\textheight,keepaspectratio]{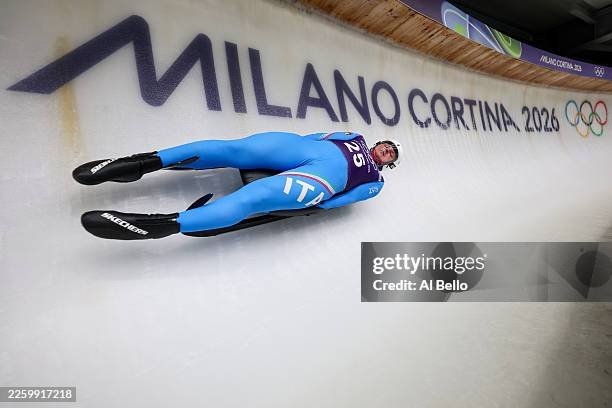}\\{\scriptsize\ttfamily search\_d707ce80.jpg}\end{minipage}\hfill \begin{minipage}[t]{0.16\linewidth}\centering\includegraphics[width=\linewidth,height=0.11\textheight,keepaspectratio]{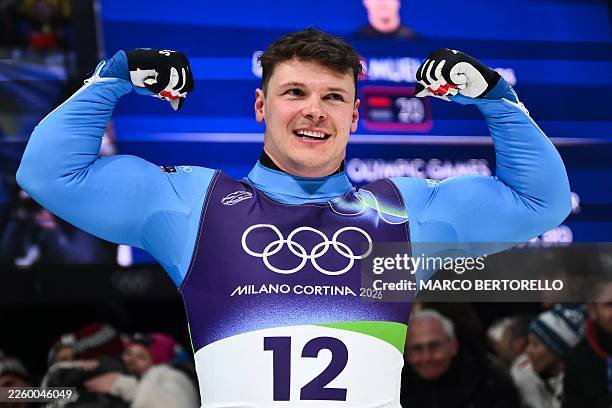}\\{\scriptsize\ttfamily search\_9c529dfd.jpg}\end{minipage}\par\vspace{2pt}{\scriptsize\sffamily\color{black!50} $\triangleright$ Full tool return (10,847 characters)}
\end{tcolorbox}
\begin{tcolorbox}[
  colback=thinkbg, colframe=thinkhead, colbacktitle=thinkhead, coltitle=white,
  boxrule=0.45pt, arc=1.4mm, left=5pt,right=5pt,top=4pt,bottom=4pt,
  toptitle=1.2pt,bottomtitle=1.2pt, fonttitle=\sffamily\bfseries\small,
  fontupper=\footnotesize\sffamily, breakable,
  title={Think $|$ Round 7}, before skip=3.5pt, after skip=3.5pt,
]
We have many images of Dominik Fischnaller. We need the one showing him celebrating after crossing the finish line. Let's look at the images carefully.\par Image 1 (Getty Images): Shows Fischnaller in a blue and purple suit being hugged by teammates in white "ITALIA" jackets. This is likely after winning bronze, but is it at the finish line? He is being hugged by teammates who are in white jackets, not racing suits. The caption says "Italy's Dominik Fischnaller celebrates winning the bronze medal as he arrives at the finish during a men's single luge run at the 2026 Winter Olympics, in Cortina". Wait ...
\end{tcolorbox}
\begin{tcolorbox}[
  colback=callbg, colframe=callhead, colbacktitle=callhead, coltitle=white,
  boxrule=0.45pt, arc=1.4mm, left=5pt,right=5pt,top=4pt,bottom=4pt,
  toptitle=1.2pt,bottomtitle=1.2pt, fonttitle=\sffamily\bfseries\small,
  fontupper=\footnotesize\sffamily, breakable,
  title={Tool call $|$ web\_search $|$ Round 7}, before skip=3.5pt, after skip=3.5pt,
]
{\ttfamily\small \{"query": ["Dominik Fischnaller suit color Milano Cortina 2026 luge", "Dominik Fischnaller racing suit blue purple", "Italy luge suit Milano Cortina 2026 colors"]\}}
\end{tcolorbox}
\begin{tcolorbox}[
  colback=obsbg, colframe=obshead, colbacktitle=obshead, coltitle=white,
  boxrule=0.45pt, arc=1.4mm, left=5pt,right=5pt,top=4pt,bottom=4pt,
  toptitle=1.2pt,bottomtitle=1.2pt, fonttitle=\sffamily\bfseries\small,
  fontupper=\footnotesize\sffamily, breakable,
  title={Tool response $|$ Round 7}, before skip=3.5pt, after skip=3.5pt,
]
{\sffamily\bfseries Summary preview}
\par{\sffamily\color{thinkhead}\bfseries Dominik Fischnaller (ITA) - Men's Singles Bronze - Luge}
\par{\scriptsize\ttfamily\color{black!55} https://www.olympics.com/***}
\par{\sffamily\color{thinkhead}\bfseries Winter Opening! \#fisiintour \#suit \#night \#milano ...}
\par{\scriptsize\ttfamily\color{black!55} https://www.instagram.com/***}
\par{\sffamily\color{thinkhead}\bfseries Dominik Fischnaller races to bronze  at the Winter ...}
\par{\scriptsize\ttfamily\color{black!55} https://www.facebook.com/***}
\par\vspace{2pt}{\scriptsize\sffamily\color{black!50} $\triangleright$ Full tool return (4,355 characters)}
\end{tcolorbox}
\begin{tcolorbox}[
  colback=thinkbg, colframe=thinkhead, colbacktitle=thinkhead, coltitle=white,
  boxrule=0.45pt, arc=1.4mm, left=5pt,right=5pt,top=4pt,bottom=4pt,
  toptitle=1.2pt,bottomtitle=1.2pt, fonttitle=\sffamily\bfseries\small,
  fontupper=\footnotesize\sffamily, breakable,
  title={Think $|$ Round 8}, before skip=3.5pt, after skip=3.5pt,
]
The search results don't specifically mention the suit colors beyond what we can see in the images. The images clearly show blue and purple. Let's verify by looking at the image more closely. We can use execute\_code to load and inspect the image if needed, but we can see it clearly. However, to be thorough, let's examine the Alamy image (search\_a0b48fcc.jpg) or the Getty image (search\_245c38a3.jpg) using code to confirm colors. But we can see them directly. The suit is predominantly blue (light blue) and purple (dark purple). The two main colors are blue and purple.\par Wait, the question says "two m ...
\end{tcolorbox}
\begin{tcolorbox}[
  colback=callbg, colframe=callhead, colbacktitle=callhead, coltitle=white,
  boxrule=0.45pt, arc=1.4mm, left=5pt,right=5pt,top=4pt,bottom=4pt,
  toptitle=1.2pt,bottomtitle=1.2pt, fonttitle=\sffamily\bfseries\small,
  fontupper=\footnotesize\sffamily, breakable,
  title={Tool call $|$ execute\_code $|$ Round 8}, before skip=3.5pt, after skip=3.5pt,
]
{\ttfamily\small \{"code": "from PIL import Image\textbackslash{}nimport matplotlib.pyplot as plt\textbackslash{}n\textbackslash{}n\# Load the Getty image showing Fischnaller celebrating at finish\textbackslash{}nimg1 = Image.open('search\_245c38a3.jpg')\textbackslash{}n\# Load the Alamy image\textbackslash{}nimg2 = Image.open('search\_a0b48fcc.jpg')\textbackslash{}n\textbackslash{}nfig, axes = plt.subplots(1, 2, figsize=(12, 6))\textbackslash{}naxes[0].imsho ...}
\end{tcolorbox}
\begin{tcolorbox}[
  colback=obsbg, colframe=obshead, colbacktitle=obshead, coltitle=white,
  boxrule=0.45pt, arc=1.4mm, left=5pt,right=5pt,top=4pt,bottom=4pt,
  toptitle=1.2pt,bottomtitle=1.2pt, fonttitle=\sffamily\bfseries\small,
  fontupper=\footnotesize\sffamily, breakable,
  title={Tool response $|$ Round 8}, before skip=3.5pt, after skip=3.5pt,
]
{\sffamily\bfseries Summary preview}
\par Tool returned a structured payload; only the preview is shown.
\par\vspace{3pt}\noindent \begin{minipage}[t]{0.58\linewidth}\centering\includegraphics[width=\linewidth,height=0.16\textheight,keepaspectratio]{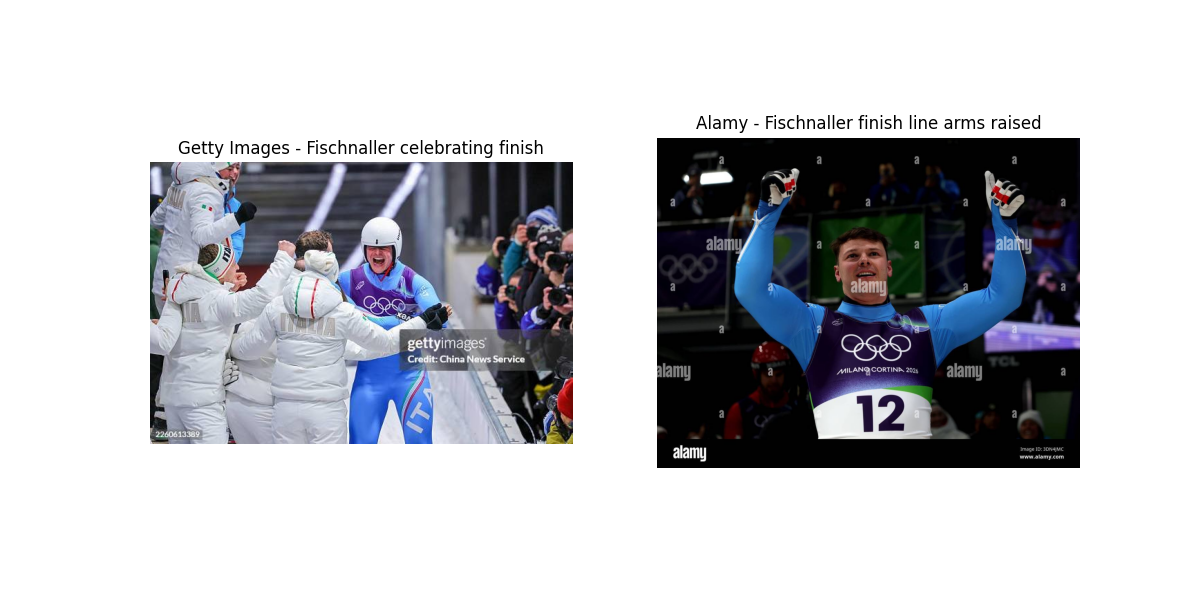}\\{\scriptsize\ttfamily fig\_001.png}\end{minipage}
\par\vspace{2pt}{\scriptsize\sffamily\color{black!50} $\triangleright$ Full tool return (171 characters)}
\end{tcolorbox}
\begin{tcolorbox}[
  colback=thinkbg, colframe=thinkhead, colbacktitle=thinkhead, coltitle=white,
  boxrule=0.45pt, arc=1.4mm, left=5pt,right=5pt,top=4pt,bottom=4pt,
  toptitle=1.2pt,bottomtitle=1.2pt, fonttitle=\sffamily\bfseries\small,
  fontupper=\footnotesize\sffamily, breakable,
  title={Think $|$ Round 9}, before skip=3.5pt, after skip=3.5pt,
]
The images confirm the racing suit colors: blue and purple. The suit has light blue sleeves and shoulders, and a dark purple torso with Olympic rings. The two main colors are blue and purple.\par Now we can formulate the final answer.\par But let's also verify that the host nation's best-performing athlete is indeed Fischnaller. We'll quickly extract the Olympics.com results page for men's singles luge to confirm his placement.
\end{tcolorbox}
\begin{tcolorbox}[
  colback=callbg, colframe=callhead, colbacktitle=callhead, coltitle=white,
  boxrule=0.45pt, arc=1.4mm, left=5pt,right=5pt,top=4pt,bottom=4pt,
  toptitle=1.2pt,bottomtitle=1.2pt, fonttitle=\sffamily\bfseries\small,
  fontupper=\footnotesize\sffamily, breakable,
  title={Tool call $|$ web\_extract $|$ Round 9}, before skip=3.5pt, after skip=3.5pt,
]
{\ttfamily\small \{"urls": ["https://www.olympics.com/***"], "goal": "Find the medalists for men's singles luge at Milano Cortina 2026 and confirm that Italy's best performer is Dominik Fischnaller with bronze."\}}
\end{tcolorbox}
\begin{tcolorbox}[
  colback=obsbg, colframe=obshead, colbacktitle=obshead, coltitle=white,
  boxrule=0.45pt, arc=1.4mm, left=5pt,right=5pt,top=4pt,bottom=4pt,
  toptitle=1.2pt,bottomtitle=1.2pt, fonttitle=\sffamily\bfseries\small,
  fontupper=\footnotesize\sffamily, breakable,
  title={Tool response $|$ Round 9}, before skip=3.5pt, after skip=3.5pt,
]
{\sffamily\bfseries Summary preview}
\par{\sffamily\color{thinkhead}\bfseries Milano-Cortina 2026 Luge - Olympic Results by Discipline}
\par{\scriptsize\ttfamily\color{black!55} https://www.olympics.com/***}
\par\vspace{3pt}\noindent \begin{minipage}[t]{0.16\linewidth}\centering\includegraphics[width=\linewidth,height=0.11\textheight,keepaspectratio]{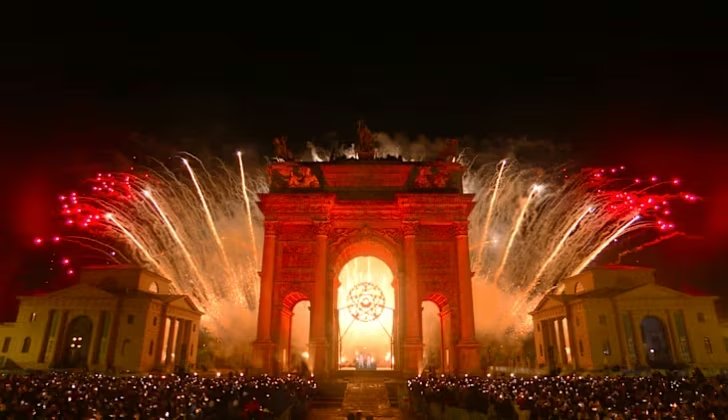}\\{\scriptsize\ttfamily extract\_c82923a6.jpg}\end{minipage}\hfill \begin{minipage}[t]{0.16\linewidth}\centering\includegraphics[width=\linewidth,height=0.11\textheight,keepaspectratio]{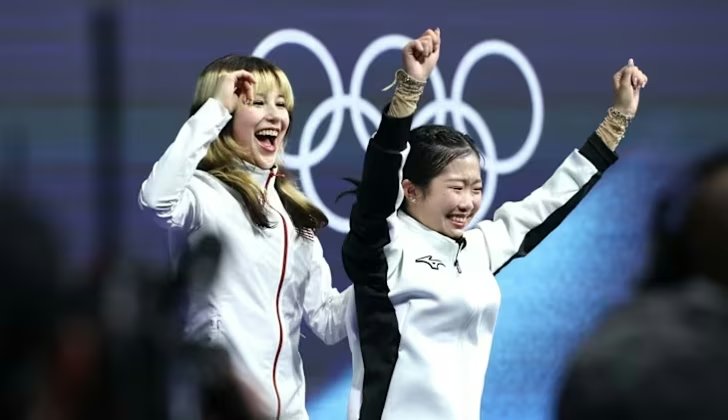}\\{\scriptsize\ttfamily extract\_e10f1b00.jpg}\end{minipage}\hfill \begin{minipage}[t]{0.16\linewidth}\centering\includegraphics[width=\linewidth,height=0.11\textheight,keepaspectratio]{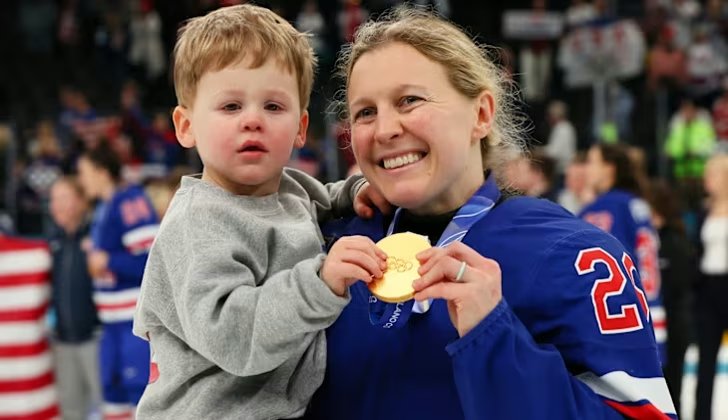}\\{\scriptsize\ttfamily extract\_009f4cf4.jpg}\end{minipage}\hfill \begin{minipage}[t]{0.16\linewidth}\centering\includegraphics[width=\linewidth,height=0.11\textheight,keepaspectratio]{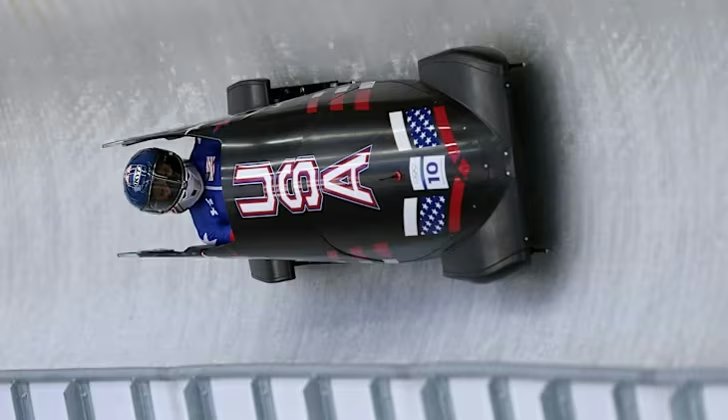}\\{\scriptsize\ttfamily extract\_79482084.jpg}\end{minipage}\par\vspace{2pt}{\scriptsize\sffamily\color{black!50} $\triangleright$ Full tool return (1,473 characters)}
\end{tcolorbox}
\begin{tcolorbox}[
  colback=thinkbg, colframe=thinkhead, colbacktitle=thinkhead, coltitle=white,
  boxrule=0.45pt, arc=1.4mm, left=5pt,right=5pt,top=4pt,bottom=4pt,
  toptitle=1.2pt,bottomtitle=1.2pt, fonttitle=\sffamily\bfseries\small,
  fontupper=\footnotesize\sffamily, breakable,
  title={Think $|$ Final}, before skip=3.5pt, after skip=3.5pt,
]
Use the verifiable evidence gathered in the trajectory and answer the question as requested.
\end{tcolorbox}
\begin{tcolorbox}[
  colback=webg, colframe=wehead, colbacktitle=wehead, coltitle=white,
  boxrule=0.45pt, arc=1.4mm, left=5pt,right=5pt,top=4pt,bottom=4pt,
  toptitle=1.2pt,bottomtitle=1.2pt, fonttitle=\sffamily\bfseries\small,
  fontupper=\footnotesize\sffamily, breakable,
  title={Final response}, before skip=3.5pt, after skip=3.5pt,
]
The athlete in the photograph is {\bfseries Max Langenhan} of Germany, celebrating his gold medal victory in men's singles luge at the {\bfseries Milano Cortina 2026 Winter Olympics}.\par The host nation of that event was Italy. Italy's best-performing athlete in men's singles luge was {\bfseries Dominik Fischnaller}, who won the bronze medal after four strong runs.\par In a news photograph showing him celebrating after crossing the finish line (e.g., the Getty Images photo titled "Fischnaller celebrates finish" or the Alamy photo "Cortina, Italia. 08th Feb, 2026. Italy's Dominik Fischnaller celebrates winning the bronze medal as he arrives at the finish..."), Fischnaller is wearing the standard Italian national luge racing suit.\par The {\bfseries two main colors of his racing suit} are:\par {\bfseries Blue and purple} -- the suit features light blue sleeves and shoulders with a dark purple torso adorned with the Olympic rings and Milano Cortina 2026 branding.
\par\vspace{4pt}\noindent \begin{minipage}[t]{0.46\linewidth}\centering\includegraphics[width=\linewidth,height=0.16\textheight,keepaspectratio]{images/rl_response_cases/case_0009_search_245c38a3.jpg}\\{\scriptsize\ttfamily search\_245c38a3.jpg}\end{minipage}\hfill \begin{minipage}[t]{0.46\linewidth}\centering\includegraphics[width=\linewidth,height=0.16\textheight,keepaspectratio]{images/rl_response_cases/case_0009_search_a0b48fcc.jpg}\\{\scriptsize\ttfamily search\_a0b48fcc.jpg}\end{minipage}
\end{tcolorbox}
\clearpage
\phantomsection\label{app:rl-case-2}
\noindent{\sffamily\bfseries\large\color{wehead} Case 2: Scholar, article, and receptor-schematic retrieval}\par
\vspace{2pt}
\respquery{This scholar has long taught immunology at the Catholic University of Leuven in Belgium, is also affiliated with an international cancer research institution, and is known for research on cytokines and cancer immunology. In an article the scholar wrote for a leading immunology review journal in 2003, a schematic systematically depicting multiple members of receptor families involved in antiviral and inflammatory regulation and the structures of their transmembrane complexes shows the portions of the receptor complexes embedded in the cell membrane mainly in what color?}
{\small\sffamily\itshape Reference answer: green.}\par
\begin{center}
{\sffamily\bfseries\small\color{wehead} Target image}\\[-1pt]
{\sffamily\scriptsize\color{black!50} (not provided to the model at query time)}\\[3pt]
\includegraphics[width=0.92\linewidth,height=0.15\textheight,keepaspectratio]{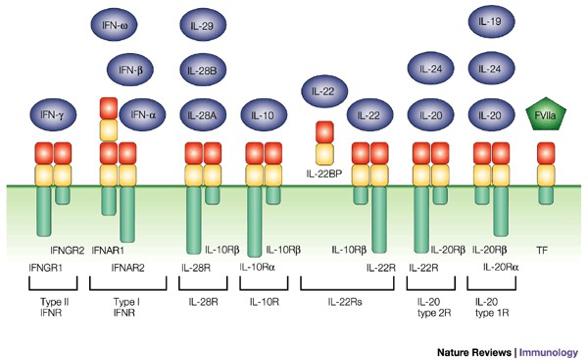}
\end{center}
\begin{tcolorbox}[
  colback=thinkbg, colframe=thinkhead, colbacktitle=thinkhead, coltitle=white,
  boxrule=0.45pt, arc=1.4mm, left=5pt,right=5pt,top=4pt,bottom=4pt,
  toptitle=1.2pt,bottomtitle=1.2pt, fonttitle=\sffamily\bfseries\small,
  fontupper=\footnotesize\sffamily, breakable,
  title={Think $|$ Round 1}, before skip=3.5pt, after skip=3.5pt,
]
The user is asking about a specific scholar who taught immunology at the Catholic University of Leuven in Belgium, is affiliated with an international cancer research institution, and is known for research on cytokines and cancer immunology. They wrote an article for a leading immunology review journal in 2003, and in that article there's a schematic showing receptor families involved in antiviral and inflammatory regulation. The question asks what color the portions of receptor complexes embedded in the cell membrane are depicted in.\par I need to identify this scholar first. Let me search for immun ...
\end{tcolorbox}
\begin{tcolorbox}[
  colback=callbg, colframe=callhead, colbacktitle=callhead, coltitle=white,
  boxrule=0.45pt, arc=1.4mm, left=5pt,right=5pt,top=4pt,bottom=4pt,
  toptitle=1.2pt,bottomtitle=1.2pt, fonttitle=\sffamily\bfseries\small,
  fontupper=\footnotesize\sffamily, breakable,
  title={Tool call $|$ web\_search $|$ Round 1}, before skip=3.5pt, after skip=3.5pt,
]
{\ttfamily\small \{"query": ["Catholic University of Leuven immunology cytokines cancer immuno", "KU Leuven immunology cytokines cancer immunology scholar 2003 review article", "Leuven immunology scholar cytokines cancer international cancer research institution", "2003 immunology review journal cytokine receptor schematic"]\}}
\end{tcolorbox}
\begin{tcolorbox}[
  colback=obsbg, colframe=obshead, colbacktitle=obshead, coltitle=white,
  boxrule=0.45pt, arc=1.4mm, left=5pt,right=5pt,top=4pt,bottom=4pt,
  toptitle=1.2pt,bottomtitle=1.2pt, fonttitle=\sffamily\bfseries\small,
  fontupper=\footnotesize\sffamily, breakable,
  title={Tool response $|$ Round 1}, before skip=3.5pt, after skip=3.5pt,
]
{\sffamily\bfseries Summary preview}
\par{\sffamily\color{thinkhead}\bfseries The role of cytokines in shaping the future of Cancer ...}
\par{\scriptsize\ttfamily\color{black!55} https://pubmed.ncbi.nlm.nih.gov/***}
\par{\sffamily\color{thinkhead}\bfseries The "double-edged sword" effect of cytokines in cancer}
\par{\scriptsize\ttfamily\color{black!55} https://www.frontiersin.org/***}
\par{\sffamily\color{thinkhead}\bfseries Research Portal - Laboratory of Cell Stress \& Immunity}
\par{\scriptsize\ttfamily\color{black!55} https://research.kuleuven.be/***}
\par\vspace{2pt}{\scriptsize\sffamily\color{black!50} $\triangleright$ Full tool return (5,747 characters)}
\end{tcolorbox}
\begin{tcolorbox}[
  colback=thinkbg, colframe=thinkhead, colbacktitle=thinkhead, coltitle=white,
  boxrule=0.45pt, arc=1.4mm, left=5pt,right=5pt,top=4pt,bottom=4pt,
  toptitle=1.2pt,bottomtitle=1.2pt, fonttitle=\sffamily\bfseries\small,
  fontupper=\footnotesize\sffamily, breakable,
  title={Think $|$ Round 2}, before skip=3.5pt, after skip=3.5pt,
]
The search results didn't directly identify the scholar. Let me try a more targeted search. The scholar might be affiliated with Ludwig Cancer Research (an international cancer research institution). Let me search for KU Leuven and Ludwig Cancer Research together.
\end{tcolorbox}
\begin{tcolorbox}[
  colback=callbg, colframe=callhead, colbacktitle=callhead, coltitle=white,
  boxrule=0.45pt, arc=1.4mm, left=5pt,right=5pt,top=4pt,bottom=4pt,
  toptitle=1.2pt,bottomtitle=1.2pt, fonttitle=\sffamily\bfseries\small,
  fontupper=\footnotesize\sffamily, breakable,
  title={Tool call $|$ web\_search $|$ Round 2}, before skip=3.5pt, after skip=3.5pt,
]
{\ttfamily\small \{"query": ["KU Leuven Ludwig Cancer Research immunology cytokines", "Leuven immunology scholar Ludwig Cancer Institute", "Ludwig Cancer Research KU Leuven cytokine receptor", "2003 Annual Review of Immunology cytokine receptor schematic"]\}}
\end{tcolorbox}
\begin{tcolorbox}[
  colback=obsbg, colframe=obshead, colbacktitle=obshead, coltitle=white,
  boxrule=0.45pt, arc=1.4mm, left=5pt,right=5pt,top=4pt,bottom=4pt,
  toptitle=1.2pt,bottomtitle=1.2pt, fonttitle=\sffamily\bfseries\small,
  fontupper=\footnotesize\sffamily, breakable,
  title={Tool response $|$ Round 2}, before skip=3.5pt, after skip=3.5pt,
]
{\sffamily\bfseries Summary preview}
\par{\sffamily\color{thinkhead}\bfseries The cytokine-serum amyloid A-chemokine network - PubMed}
\par{\scriptsize\ttfamily\color{black!55} https://pubmed.ncbi.nlm.nih.gov/***}
\par{\sffamily\color{thinkhead}\bfseries Ludwig Lausanne's}
\par{\scriptsize\ttfamily\color{black!55} https://www.ludwigcancerresearch.org/***}
\par{\sffamily\color{thinkhead}\bfseries Aminereactive OVA multimers for autovaccination against ...}
\par{\scriptsize\ttfamily\color{black!55} https://jlb.onlinelibrary.wiley.com/***}
\par\vspace{2pt}{\scriptsize\sffamily\color{black!50} $\triangleright$ Full tool return (5,555 characters)}
\end{tcolorbox}
\begin{tcolorbox}[
  colback=thinkbg, colframe=thinkhead, colbacktitle=thinkhead, coltitle=white,
  boxrule=0.45pt, arc=1.4mm, left=5pt,right=5pt,top=4pt,bottom=4pt,
  toptitle=1.2pt,bottomtitle=1.2pt, fonttitle=\sffamily\bfseries\small,
  fontupper=\footnotesize\sffamily, breakable,
  title={Think $|$ Round 3}, before skip=3.5pt, after skip=3.5pt,
]
I need to identify the specific scholar. Let me search for leading immunology review journals in 2003 that might have had an article about cytokine receptors. The Annual Review of Immunology was published in 2003. Let me search for that specific issue.
\end{tcolorbox}
\begin{tcolorbox}[
  colback=callbg, colframe=callhead, colbacktitle=callhead, coltitle=white,
  boxrule=0.45pt, arc=1.4mm, left=5pt,right=5pt,top=4pt,bottom=4pt,
  toptitle=1.2pt,bottomtitle=1.2pt, fonttitle=\sffamily\bfseries\small,
  fontupper=\footnotesize\sffamily, breakable,
  title={Tool call $|$ web\_search $|$ Round 3}, before skip=3.5pt, after skip=3.5pt,
]
{\ttfamily\small \{"query": ["Annual Review of Immunology 2003 cytokine receptors schematic", "Annual Review of Immunology 2003 receptor families antiviral inflammatory regulation", "2003 Immunology review cytokine receptor complex membrane embedded color"]\}}
\end{tcolorbox}
\begin{tcolorbox}[
  colback=obsbg, colframe=obshead, colbacktitle=obshead, coltitle=white,
  boxrule=0.45pt, arc=1.4mm, left=5pt,right=5pt,top=4pt,bottom=4pt,
  toptitle=1.2pt,bottomtitle=1.2pt, fonttitle=\sffamily\bfseries\small,
  fontupper=\footnotesize\sffamily, breakable,
  title={Tool response $|$ Round 3}, before skip=3.5pt, after skip=3.5pt,
]
{\sffamily\bfseries Summary preview}
\par{\sffamily\color{thinkhead}\bfseries Cytokine Receptor - an overview}
\par{\scriptsize\ttfamily\color{black!55} https://www.sciencedirect.com/***}
\par{\sffamily\color{thinkhead}\bfseries Mechanisms of Signal Transduction from Receptors ...}
\par{\scriptsize\ttfamily\color{black!55} https://www.tandfonline.com/***}
\par{\sffamily\color{thinkhead}\bfseries Cytokine Signaling Modules in Inflammatory Responses}
\par{\scriptsize\ttfamily\color{black!55} https://www.cell.com/***}
\par\vspace{2pt}{\scriptsize\sffamily\color{black!50} $\triangleright$ Full tool return (4,272 characters)}
\end{tcolorbox}
\begin{tcolorbox}[
  colback=thinkbg, colframe=thinkhead, colbacktitle=thinkhead, coltitle=white,
  boxrule=0.45pt, arc=1.4mm, left=5pt,right=5pt,top=4pt,bottom=4pt,
  toptitle=1.2pt,bottomtitle=1.2pt, fonttitle=\sffamily\bfseries\small,
  fontupper=\footnotesize\sffamily, breakable,
  title={Think $|$ Round 4}, before skip=3.5pt, after skip=3.5pt,
]
Let me search more specifically for the 2003 Annual Review of Immunology issue and look for an article about cytokine receptor families, antiviral, and inflammatory regulation. I need to identify the author.
\end{tcolorbox}
\begin{tcolorbox}[
  colback=callbg, colframe=callhead, colbacktitle=callhead, coltitle=white,
  boxrule=0.45pt, arc=1.4mm, left=5pt,right=5pt,top=4pt,bottom=4pt,
  toptitle=1.2pt,bottomtitle=1.2pt, fonttitle=\sffamily\bfseries\small,
  fontupper=\footnotesize\sffamily, breakable,
  title={Tool call $|$ web\_search $|$ Round 4}, before skip=3.5pt, after skip=3.5pt,
]
{\ttfamily\small \{"query": ["2003 Annual Review of Immunology volume 21 cytokine receptors", "Annual Review of Immunology 2003 \textbackslash{}"interferon\textbackslash{}" interleukin receptor families", "2003 Immunology review journal \textbackslash{}"schematic\textbackslash{}" \textbackslash{}"receptor families\textbackslash{}" antiviral"]\}}
\end{tcolorbox}
\begin{tcolorbox}[
  colback=obsbg, colframe=obshead, colbacktitle=obshead, coltitle=white,
  boxrule=0.45pt, arc=1.4mm, left=5pt,right=5pt,top=4pt,bottom=4pt,
  toptitle=1.2pt,bottomtitle=1.2pt, fonttitle=\sffamily\bfseries\small,
  fontupper=\footnotesize\sffamily, breakable,
  title={Tool response $|$ Round 4}, before skip=3.5pt, after skip=3.5pt,
]
{\sffamily\bfseries Summary preview}
\par{\sffamily\color{thinkhead}\bfseries Annual Review of Immunology - Volume 21, 2003}
\par{\scriptsize\ttfamily\color{black!55} https://www.annualreviews.org/***}
\par{\sffamily\color{thinkhead}\bfseries Annual review of immunology}
\par{\scriptsize\ttfamily\color{black!55} https://www.scielo.br/***}
\par{\sffamily\color{thinkhead}\bfseries (PDF) Annual review of immunology}
\par{\scriptsize\ttfamily\color{black!55} https://www.researchgate.net/***}
\par\vspace{2pt}{\scriptsize\sffamily\color{black!50} $\triangleright$ Full tool return (4,330 characters)}
\end{tcolorbox}
\begin{tcolorbox}[
  colback=thinkbg, colframe=thinkhead, colbacktitle=thinkhead, coltitle=white,
  boxrule=0.45pt, arc=1.4mm, left=5pt,right=5pt,top=4pt,bottom=4pt,
  toptitle=1.2pt,bottomtitle=1.2pt, fonttitle=\sffamily\bfseries\small,
  fontupper=\footnotesize\sffamily, breakable,
  title={Think $|$ Round 5}, before skip=3.5pt, after skip=3.5pt,
]
I need to be more systematic. Let me search for the specific article in the 2003 Annual Review of Immunology about cytokine receptors and their families involved in antiviral and inflammatory regulation. The description mentions "multiple members of receptor families involved in antiviral and inflammatory regulation" and a "schematic systematically depicting" them. This sounds like it could be about the interferon receptor family or interleukin receptor family.\par Let me search for 2003 Annual Review of Immunology articles specifically about cytokine receptors and receptor families.
\end{tcolorbox}
\begin{tcolorbox}[
  colback=callbg, colframe=callhead, colbacktitle=callhead, coltitle=white,
  boxrule=0.45pt, arc=1.4mm, left=5pt,right=5pt,top=4pt,bottom=4pt,
  toptitle=1.2pt,bottomtitle=1.2pt, fonttitle=\sffamily\bfseries\small,
  fontupper=\footnotesize\sffamily, breakable,
  title={Tool call $|$ web\_search $|$ Round 5}, before skip=3.5pt, after skip=3.5pt,
]
{\ttfamily\small \{"query": ["2003 Annual Review of Immunology cytokine receptor families interferon interleukin", "Annual Review of Immunology 2003 \textbackslash{}"cytokine\textbackslash{}" \textbackslash{}"receptor\textbackslash{}" family schematic", "2003 \textbackslash{}"Annual Review of Immunology\textbackslash{}" \textbackslash{}"interferon receptor\textbackslash{}" family", "2003 \textbackslash{}"Annual Review of Immunology\textbackslash{}" \textbackslash{}"interleukin receptor\textbackslash{}" family"]\}}
\end{tcolorbox}
\begin{tcolorbox}[
  colback=obsbg, colframe=obshead, colbacktitle=obshead, coltitle=white,
  boxrule=0.45pt, arc=1.4mm, left=5pt,right=5pt,top=4pt,bottom=4pt,
  toptitle=1.2pt,bottomtitle=1.2pt, fonttitle=\sffamily\bfseries\small,
  fontupper=\footnotesize\sffamily, breakable,
  title={Tool response $|$ Round 5}, before skip=3.5pt, after skip=3.5pt,
]
{\sffamily\bfseries Summary preview}
\par{\sffamily\color{thinkhead}\bfseries Review The interleukin-10 family of cytokines}
\par{\scriptsize\ttfamily\color{black!55} https://www.sciencedirect.com/***}
\par{\sffamily\color{thinkhead}\bfseries Insights into Cytokine--Receptor Interactions from ... - PMC - NIH}
\par{\scriptsize\ttfamily\color{black!55} https://pmc.ncbi.nlm.nih.gov/***}
\par{\sffamily\color{thinkhead}\bfseries A Life with Cytokines, Accompanied by Friends and Music}
\par{\scriptsize\ttfamily\color{black!55} https://www.annualreviews.org/***}
\par\vspace{2pt}{\scriptsize\sffamily\color{black!50} $\triangleright$ Full tool return (5,979 characters)}
\end{tcolorbox}
\begin{tcolorbox}[
  colback=thinkbg, colframe=thinkhead, colbacktitle=thinkhead, coltitle=white,
  boxrule=0.45pt, arc=1.4mm, left=5pt,right=5pt,top=4pt,bottom=4pt,
  toptitle=1.2pt,bottomtitle=1.2pt, fonttitle=\sffamily\bfseries\small,
  fontupper=\footnotesize\sffamily, breakable,
  title={Think $|$ Round 6}, before skip=3.5pt, after skip=3.5pt,
]
The search results mention that the 2002 article by H Fickenscher in Annual Review of Immunology discusses the CRF2 (cytokine receptor family type 2) which comprises interferons. But I need the 2003 article. Let me think about this differently.\par The description says: "a schematic systematically depicting multiple members of receptor families involved in antiviral and inflammatory regulation and the structures of their transmembrane complexes shows the portions of the receptor complexes embedded in the cell membrane mainly in what color?"\par This sounds like a specific schematic from a review article ...
\end{tcolorbox}
\begin{tcolorbox}[
  colback=callbg, colframe=callhead, colbacktitle=callhead, coltitle=white,
  boxrule=0.45pt, arc=1.4mm, left=5pt,right=5pt,top=4pt,bottom=4pt,
  toptitle=1.2pt,bottomtitle=1.2pt, fonttitle=\sffamily\bfseries\small,
  fontupper=\footnotesize\sffamily, breakable,
  title={Tool call $|$ web\_search $|$ Round 6}, before skip=3.5pt, after skip=3.5pt,
]
{\ttfamily\small \{"query": ["KU Leuven Ludwig Cancer Research immunology professor", "Catholic University of Leuven Ludwig cancer research senior scientist", "KU Leuven immunology Ludwig Institute for Cancer Research"]\}}
\end{tcolorbox}
\begin{tcolorbox}[
  colback=obsbg, colframe=obshead, colbacktitle=obshead, coltitle=white,
  boxrule=0.45pt, arc=1.4mm, left=5pt,right=5pt,top=4pt,bottom=4pt,
  toptitle=1.2pt,bottomtitle=1.2pt, fonttitle=\sffamily\bfseries\small,
  fontupper=\footnotesize\sffamily, breakable,
  title={Tool response $|$ Round 6}, before skip=3.5pt, after skip=3.5pt,
]
{\sffamily\bfseries Summary preview}
\par{\sffamily\color{thinkhead}\bfseries Benoit Van den Eynde}
\par{\scriptsize\ttfamily\color{black!55} https://www.ludwigcancerresearch.org/***}
\par{\sffamily\color{thinkhead}\bfseries Ludwig Lausanne's Mikael Pittet}
\par{\scriptsize\ttfamily\color{black!55} https://www.ludwigcancerresearch.org/***}
\par{\sffamily\color{thinkhead}\bfseries Laboratory for Tumor Immunology and Immunotherapy}
\par{\scriptsize\ttfamily\color{black!55} https://research.kuleuven.be/***}
\par\vspace{2pt}{\scriptsize\sffamily\color{black!50} $\triangleright$ Full tool return (4,098 characters)}
\end{tcolorbox}
\begin{tcolorbox}[
  colback=thinkbg, colframe=thinkhead, colbacktitle=thinkhead, coltitle=white,
  boxrule=0.45pt, arc=1.4mm, left=5pt,right=5pt,top=4pt,bottom=4pt,
  toptitle=1.2pt,bottomtitle=1.2pt, fonttitle=\sffamily\bfseries\small,
  fontupper=\footnotesize\sffamily, breakable,
  title={Think $|$ Round 7}, before skip=3.5pt, after skip=3.5pt,
]
I see Benoit Van den Eynde is mentioned as a Member of the Ludwig Institute for Cancer Research and co-director of the de Duve Institute. He's a tumor immunologist. Let me search for him specifically, as he might have published in Annual Review of Immunology in 2003.
\end{tcolorbox}
\begin{tcolorbox}[
  colback=callbg, colframe=callhead, colbacktitle=callhead, coltitle=white,
  boxrule=0.45pt, arc=1.4mm, left=5pt,right=5pt,top=4pt,bottom=4pt,
  toptitle=1.2pt,bottomtitle=1.2pt, fonttitle=\sffamily\bfseries\small,
  fontupper=\footnotesize\sffamily, breakable,
  title={Tool call $|$ web\_search $|$ Round 7}, before skip=3.5pt, after skip=3.5pt,
]
{\ttfamily\small \{"query": ["Benoit Van den Eynde Annual Review of Immunology 2003", "Benoit Van den Eynde cytokine receptor schematic", "Benoit Van den Eynde 2003 immunology review article"]\}}
\end{tcolorbox}
\begin{tcolorbox}[
  colback=obsbg, colframe=obshead, colbacktitle=obshead, coltitle=white,
  boxrule=0.45pt, arc=1.4mm, left=5pt,right=5pt,top=4pt,bottom=4pt,
  toptitle=1.2pt,bottomtitle=1.2pt, fonttitle=\sffamily\bfseries\small,
  fontupper=\footnotesize\sffamily, breakable,
  title={Tool response $|$ Round 7}, before skip=3.5pt, after skip=3.5pt,
]
{\sffamily\bfseries Summary preview}
\par{\sffamily\color{thinkhead}\bfseries Tumor Antigens Recognized by T Lymphocytes}
\par{\scriptsize\ttfamily\color{black!55} https://www.annualreviews.org/***}
\par{\sffamily\color{thinkhead}\bfseries The tumor defense dismantler}
\par{\scriptsize\ttfamily\color{black!55} https://www.ludwigcancerresearch.org/***}
\par{\sffamily\color{thinkhead}\bfseries Tumour immunology}
\par{\scriptsize\ttfamily\color{black!55} https://pubmed.ncbi.nlm.nih.gov/***}
\par\vspace{2pt}{\scriptsize\sffamily\color{black!50} $\triangleright$ Full tool return (4,133 characters)}
\end{tcolorbox}
\begin{tcolorbox}[
  colback=thinkbg, colframe=thinkhead, colbacktitle=thinkhead, coltitle=white,
  boxrule=0.45pt, arc=1.4mm, left=5pt,right=5pt,top=4pt,bottom=4pt,
  toptitle=1.2pt,bottomtitle=1.2pt, fonttitle=\sffamily\bfseries\small,
  fontupper=\footnotesize\sffamily, breakable,
  title={Think $|$ Round 8}, before skip=3.5pt, after skip=3.5pt,
]
Benoit Van den Eynde is a tumor immunologist at the Ludwig Institute, but he's more associated with tumor immunology and antigens rather than cytokines and receptor families specifically. Let me search for other KU Leuven scholars affiliated with Ludwig Cancer Research who focus on cytokines and receptor families. Maybe it's a different person.\par Let me search for "KU Leuven cytokine receptor family" directly.
\end{tcolorbox}
\begin{tcolorbox}[
  colback=callbg, colframe=callhead, colbacktitle=callhead, coltitle=white,
  boxrule=0.45pt, arc=1.4mm, left=5pt,right=5pt,top=4pt,bottom=4pt,
  toptitle=1.2pt,bottomtitle=1.2pt, fonttitle=\sffamily\bfseries\small,
  fontupper=\footnotesize\sffamily, breakable,
  title={Tool call $|$ web\_search $|$ Round 8}, before skip=3.5pt, after skip=3.5pt,
]
{\ttfamily\small \{"query": ["KU Leuven cytokine receptor family", "Catholic University of Leuven cytokine receptor", "KU Leuven immunology cytokine receptor family professor", "KU Leuven cancer immunology cytokine"]\}}
\end{tcolorbox}
\begin{tcolorbox}[
  colback=obsbg, colframe=obshead, colbacktitle=obshead, coltitle=white,
  boxrule=0.45pt, arc=1.4mm, left=5pt,right=5pt,top=4pt,bottom=4pt,
  toptitle=1.2pt,bottomtitle=1.2pt, fonttitle=\sffamily\bfseries\small,
  fontupper=\footnotesize\sffamily, breakable,
  title={Tool response $|$ Round 8}, before skip=3.5pt, after skip=3.5pt,
]
{\sffamily\bfseries Summary preview}
\par{\sffamily\color{thinkhead}\bfseries Structural Biology of Shared Cytokine Receptors - PMC}
\par{\scriptsize\ttfamily\color{black!55} https://pmc.ncbi.nlm.nih.gov/***}
\par{\sffamily\color{thinkhead}\bfseries Laboratory of Molecular Immunology}
\par{\scriptsize\ttfamily\color{black!55} https://rega.kuleuven.be/***}
\par{\sffamily\color{thinkhead}\bfseries Paul Proost KU Leuven}
\par{\scriptsize\ttfamily\color{black!55} https://www.researchgate.net/***}
\par\vspace{2pt}{\scriptsize\sffamily\color{black!50} $\triangleright$ Full tool return (5,485 characters)}
\end{tcolorbox}
\begin{tcolorbox}[
  colback=thinkbg, colframe=thinkhead, colbacktitle=thinkhead, coltitle=white,
  boxrule=0.45pt, arc=1.4mm, left=5pt,right=5pt,top=4pt,bottom=4pt,
  toptitle=1.2pt,bottomtitle=1.2pt, fonttitle=\sffamily\bfseries\small,
  fontupper=\footnotesize\sffamily, breakable,
  title={Think $|$ Round 9}, before skip=3.5pt, after skip=3.5pt,
]
I found Ghislain OPDENAKKER listed as Professor of Immunology, Head of Immunobiology, focusing on cytokines, chemokines, adhesion molecules and other membrane receptors at KU Leuven. He also has an MD from Catholic University of Leuven. Let me search for him specifically.
\end{tcolorbox}
\begin{tcolorbox}[
  colback=callbg, colframe=callhead, colbacktitle=callhead, coltitle=white,
  boxrule=0.45pt, arc=1.4mm, left=5pt,right=5pt,top=4pt,bottom=4pt,
  toptitle=1.2pt,bottomtitle=1.2pt, fonttitle=\sffamily\bfseries\small,
  fontupper=\footnotesize\sffamily, breakable,
  title={Tool call $|$ web\_search $|$ Round 9}, before skip=3.5pt, after skip=3.5pt,
]
{\ttfamily\small \{"query": ["Ghislain Opdenakker Annual Review of Immunology 2003", "Ghislain Opdenakker cytokine receptor schematic", "Opdenakker 2003 cytokine receptor review"]\}}
\end{tcolorbox}
\begin{tcolorbox}[
  colback=obsbg, colframe=obshead, colbacktitle=obshead, coltitle=white,
  boxrule=0.45pt, arc=1.4mm, left=5pt,right=5pt,top=4pt,bottom=4pt,
  toptitle=1.2pt,bottomtitle=1.2pt, fonttitle=\sffamily\bfseries\small,
  fontupper=\footnotesize\sffamily, breakable,
  title={Tool response $|$ Round 9}, before skip=3.5pt, after skip=3.5pt,
]
{\sffamily\bfseries Summary preview}
\par{\sffamily\color{thinkhead}\bfseries Interferons and other cytokines, genetics and beyond in ... - PMC}
\par{\scriptsize\ttfamily\color{black!55} https://pmc.ncbi.nlm.nih.gov/***}
\par{\sffamily\color{thinkhead}\bfseries Ghislain Opdenakker - Loop}
\par{\scriptsize\ttfamily\color{black!55} https://loop.frontiersin.org/***}
\par{\sffamily\color{thinkhead}\bfseries Ghislain OPDENAKKER | Professor of Immunology, Head ...}
\par{\scriptsize\ttfamily\color{black!55} https://www.researchgate.net/***}
\par\vspace{2pt}{\scriptsize\sffamily\color{black!50} $\triangleright$ Full tool return (4,229 characters)}
\end{tcolorbox}
\begin{tcolorbox}[
  colback=thinkbg, colframe=thinkhead, colbacktitle=thinkhead, coltitle=white,
  boxrule=0.45pt, arc=1.4mm, left=5pt,right=5pt,top=4pt,bottom=4pt,
  toptitle=1.2pt,bottomtitle=1.2pt, fonttitle=\sffamily\bfseries\small,
  fontupper=\footnotesize\sffamily, breakable,
  title={Think $|$ Round 10}, before skip=3.5pt, after skip=3.5pt,
]
Ghislain Opdenakker is a professor at KU Leuven focusing on cytokines, but I'm not sure if he published in Annual Review of Immunology in 2003. Let me search more systematically for 2003 Annual Review of Immunology articles about cytokine receptors and their families. The description mentions "multiple members of receptor families involved in antiviral and inflammatory regulation" - this sounds like it could be about the interferon receptor family or interleukin receptor family.\par Let me try to find the 2003 Annual Review of Immunology table of contents or specific chapter about cytokine receptors.
\end{tcolorbox}
\begin{tcolorbox}[
  colback=callbg, colframe=callhead, colbacktitle=callhead, coltitle=white,
  boxrule=0.45pt, arc=1.4mm, left=5pt,right=5pt,top=4pt,bottom=4pt,
  toptitle=1.2pt,bottomtitle=1.2pt, fonttitle=\sffamily\bfseries\small,
  fontupper=\footnotesize\sffamily, breakable,
  title={Tool call $|$ web\_search $|$ Round 10}, before skip=3.5pt, after skip=3.5pt,
]
{\ttfamily\small \{"query": ["2003 Annual Review of Immunology table of contents cytokine receptors", "Annual Review of Immunology 2003 \textbackslash{}"cytokine receptors\textbackslash{}" \textbackslash{}"receptor family\textbackslash{}"", "2003 Annual Review of Immunology \textbackslash{}"interferon\textbackslash{}" \textbackslash{}"receptor\textbackslash{}" schematic", "2003 Annual Review of Immunology \textbackslash{}"interleukin\textbackslash{}" \textbackslash{}"receptor\textbackslash{}""]\}}
\end{tcolorbox}
\begin{tcolorbox}[
  colback=obsbg, colframe=obshead, colbacktitle=obshead, coltitle=white,
  boxrule=0.45pt, arc=1.4mm, left=5pt,right=5pt,top=4pt,bottom=4pt,
  toptitle=1.2pt,bottomtitle=1.2pt, fonttitle=\sffamily\bfseries\small,
  fontupper=\footnotesize\sffamily, breakable,
  title={Tool response $|$ Round 10}, before skip=3.5pt, after skip=3.5pt,
]
{\sffamily\bfseries Summary preview}
\par{\sffamily\color{thinkhead}\bfseries Structural Biology of Shared Cytokine Receptors}
\par{\scriptsize\ttfamily\color{black!55} https://www.annualreviews.org/***}
\par{\sffamily\color{thinkhead}\bfseries Interleukin-10 and Related Cytokines and Receptors}
\par{\scriptsize\ttfamily\color{black!55} https://www.annualreviews.org/***}
\par{\sffamily\color{thinkhead}\bfseries Cytokine-mediated communications: a quantitative appraisal ...}
\par{\scriptsize\ttfamily\color{black!55} https://pmc.ncbi.nlm.nih.gov/***}
\par\vspace{2pt}{\scriptsize\sffamily\color{black!50} $\triangleright$ Full tool return (6,021 characters)}
\end{tcolorbox}
\begin{tcolorbox}[
  colback=thinkbg, colframe=thinkhead, colbacktitle=thinkhead, coltitle=white,
  boxrule=0.45pt, arc=1.4mm, left=5pt,right=5pt,top=4pt,bottom=4pt,
  toptitle=1.2pt,bottomtitle=1.2pt, fonttitle=\sffamily\bfseries\small,
  fontupper=\footnotesize\sffamily, breakable,
  title={Think $|$ Round 11}, before skip=3.5pt, after skip=3.5pt,
]
Let me try to find the specific 2003 Annual Review of Immunology article about multiple receptor families involved in antiviral and inflammatory regulation. The description mentions "schematic systematically depicting multiple members of receptor families" - this sounds like it could be a figure showing the classification of cytokine receptors.\par Let me search for "2003 Annual Review of Immunology cytokine receptor families schematic figure" or try to access the actual PDF of the 2003 issue.
\end{tcolorbox}
\begin{tcolorbox}[
  colback=callbg, colframe=callhead, colbacktitle=callhead, coltitle=white,
  boxrule=0.45pt, arc=1.4mm, left=5pt,right=5pt,top=4pt,bottom=4pt,
  toptitle=1.2pt,bottomtitle=1.2pt, fonttitle=\sffamily\bfseries\small,
  fontupper=\footnotesize\sffamily, breakable,
  title={Tool call $|$ web\_search $|$ Round 11}, before skip=3.5pt, after skip=3.5pt,
]
{\ttfamily\small \{"query": ["2003 Annual Review of Immunology cytokine receptor families figure schematic", "Annual Review of Immunology 2003 cytokine receptor classification", "2003 Annual Reviews immunology cytokine receptor family diagram color", "2003 Annual Review of Immunology \textbackslash{}"interferon\textbackslash{}" \textbackslash{}"interleukin\textbackslash{}" receptor family"]\}}
\end{tcolorbox}
\begin{tcolorbox}[
  colback=obsbg, colframe=obshead, colbacktitle=obshead, coltitle=white,
  boxrule=0.45pt, arc=1.4mm, left=5pt,right=5pt,top=4pt,bottom=4pt,
  toptitle=1.2pt,bottomtitle=1.2pt, fonttitle=\sffamily\bfseries\small,
  fontupper=\footnotesize\sffamily, breakable,
  title={Tool response $|$ Round 11}, before skip=3.5pt, after skip=3.5pt,
]
{\sffamily\bfseries Summary preview}
\par{\sffamily\color{thinkhead}\bfseries The Common Cytokine Receptor  Chain Family of ... - PMC}
\par{\scriptsize\ttfamily\color{black!55} https://pmc.ncbi.nlm.nih.gov/***}
\par{\sffamily\color{thinkhead}\bfseries Cytokine Receptor - an overview}
\par{\scriptsize\ttfamily\color{black!55} https://www.sciencedirect.com/***}
\par{\sffamily\color{thinkhead}\bfseries The c Family of Cytokines: Basic Biology to Therapeutic ...}
\par{\scriptsize\ttfamily\color{black!55} https://www.cell.com/***}
\par\vspace{2pt}{\scriptsize\sffamily\color{black!50} $\triangleright$ Full tool return (5,968 characters)}
\end{tcolorbox}
\begin{tcolorbox}[
  colback=thinkbg, colframe=thinkhead, colbacktitle=thinkhead, coltitle=white,
  boxrule=0.45pt, arc=1.4mm, left=5pt,right=5pt,top=4pt,bottom=4pt,
  toptitle=1.2pt,bottomtitle=1.2pt, fonttitle=\sffamily\bfseries\small,
  fontupper=\footnotesize\sffamily, breakable,
  title={Think $|$ Round 12}, before skip=3.5pt, after skip=3.5pt,
]
Let me try a different approach. The scholar might be someone like Willem J. (WIJ) J. (maybe a professor at KU Leuven). Let me search for "KU Leuven immunology professor Ludwig Institute cytokine" more broadly. Also, maybe the scholar is a Belgian immunologist who has published in Annual Review of Immunology about cytokine receptors. Let me search for "Annual Review of Immunology 2003" with more specific terms about receptor families.\par Actually, the description mentions "multiple members of receptor families involved in antiviral and inflammatory regulation" - this sounds like it could be about th ...
\end{tcolorbox}
\begin{tcolorbox}[
  colback=callbg, colframe=callhead, colbacktitle=callhead, coltitle=white,
  boxrule=0.45pt, arc=1.4mm, left=5pt,right=5pt,top=4pt,bottom=4pt,
  toptitle=1.2pt,bottomtitle=1.2pt, fonttitle=\sffamily\bfseries\small,
  fontupper=\footnotesize\sffamily, breakable,
  title={Tool call $|$ web\_search $|$ Round 12}, before skip=3.5pt, after skip=3.5pt,
]
{\ttfamily\small \{"query": ["2003 Annual Review of Immunology cytokine receptors review article", "2003 Annual Review of Immunology \textbackslash{}"cytokine\textbackslash{}" \textbackslash{}"receptor\textbackslash{}" \textbackslash{}"families\textbackslash{}"", "2003 Annual Review of Immunology cytokine receptor family review", "Annual Review of Immunology 2003 \textbackslash{}"receptor\textbackslash{}" \textbackslash{}"cytokine\textbackslash{}" \textbackslash{}"inflammatory\textbackslash{}""]\}}
\end{tcolorbox}
\begin{tcolorbox}[
  colback=obsbg, colframe=obshead, colbacktitle=obshead, coltitle=white,
  boxrule=0.45pt, arc=1.4mm, left=5pt,right=5pt,top=4pt,bottom=4pt,
  toptitle=1.2pt,bottomtitle=1.2pt, fonttitle=\sffamily\bfseries\small,
  fontupper=\footnotesize\sffamily, breakable,
  title={Tool response $|$ Round 12}, before skip=3.5pt, after skip=3.5pt,
]
{\sffamily\bfseries Summary preview}
\par{\sffamily\color{thinkhead}\bfseries Structural Biology of Shared Cytokine Receptors}
\par{\scriptsize\ttfamily\color{black!55} https://www.annualreviews.org/***}
\par{\sffamily\color{thinkhead}\bfseries Interleukin-10 and Related Cytokines and Receptors}
\par{\scriptsize\ttfamily\color{black!55} https://www.annualreviews.org/***}
\par{\sffamily\color{thinkhead}\bfseries Historical Review of Cytokines - PMC - NIH}
\par{\scriptsize\ttfamily\color{black!55} https://pmc.ncbi.nlm.nih.gov/***}
\par\vspace{2pt}{\scriptsize\sffamily\color{black!50} $\triangleright$ Full tool return (5,922 characters)}
\end{tcolorbox}
\begin{tcolorbox}[
  colback=thinkbg, colframe=thinkhead, colbacktitle=thinkhead, coltitle=white,
  boxrule=0.45pt, arc=1.4mm, left=5pt,right=5pt,top=4pt,bottom=4pt,
  toptitle=1.2pt,bottomtitle=1.2pt, fonttitle=\sffamily\bfseries\small,
  fontupper=\footnotesize\sffamily, breakable,
  title={Think $|$ Round 13}, before skip=3.5pt, after skip=3.5pt,
]
The search results are not giving me the specific article. Let me try to find the 2003 Annual Review of Immunology table of contents or specific chapter. The description mentions "multiple members of receptor families involved in antiviral and inflammatory regulation" - this could be referring to a figure showing the classification of cytokine receptors across different families. Let me search for "Annual Review of Immunology 2003 cytokine receptors schematic figure color" or try to find the actual PDF of the 2003 issue.\par Actually, let me think about this differently. The scholar taught at Catholi ...
\end{tcolorbox}
\begin{tcolorbox}[
  colback=callbg, colframe=callhead, colbacktitle=callhead, coltitle=white,
  boxrule=0.45pt, arc=1.4mm, left=5pt,right=5pt,top=4pt,bottom=4pt,
  toptitle=1.2pt,bottomtitle=1.2pt, fonttitle=\sffamily\bfseries\small,
  fontupper=\footnotesize\sffamily, breakable,
  title={Tool call $|$ web\_search $|$ Round 13}, before skip=3.5pt, after skip=3.5pt,
]
{\ttfamily\small \{"query": ["Ludwig Institute for Cancer Research KU Leuven professor cytokine", "Ludwig cancer research KU Leuven immunology professor", "KU Leuven Ludwig Institute for Cancer Research immunologist"]\}}
\end{tcolorbox}
\begin{tcolorbox}[
  colback=obsbg, colframe=obshead, colbacktitle=obshead, coltitle=white,
  boxrule=0.45pt, arc=1.4mm, left=5pt,right=5pt,top=4pt,bottom=4pt,
  toptitle=1.2pt,bottomtitle=1.2pt, fonttitle=\sffamily\bfseries\small,
  fontupper=\footnotesize\sffamily, breakable,
  title={Tool response $|$ Round 13}, before skip=3.5pt, after skip=3.5pt,
]
{\sffamily\bfseries Summary preview}
\par{\sffamily\color{thinkhead}\bfseries Ludwig Cancer Research and Oxford University welcome ...}
\par{\scriptsize\ttfamily\color{black!55} https://www.ludwigcancerresearch.org/***}
\par{\sffamily\color{thinkhead}\bfseries Stefan Constantinescu}
\par{\scriptsize\ttfamily\color{black!55} https://www.deduveinstitute.be/***}
\par{\sffamily\color{thinkhead}\bfseries Jean-Christophe Renauld - Professor at the Universite ...}
\par{\scriptsize\ttfamily\color{black!55} https://be.linkedin.com/***}
\par\vspace{2pt}{\scriptsize\sffamily\color{black!50} $\triangleright$ Full tool return (4,112 characters)}
\end{tcolorbox}
\begin{tcolorbox}[
  colback=thinkbg, colframe=thinkhead, colbacktitle=thinkhead, coltitle=white,
  boxrule=0.45pt, arc=1.4mm, left=5pt,right=5pt,top=4pt,bottom=4pt,
  toptitle=1.2pt,bottomtitle=1.2pt, fonttitle=\sffamily\bfseries\small,
  fontupper=\footnotesize\sffamily, breakable,
  title={Think $|$ Round 14}, before skip=3.5pt, after skip=3.5pt,
]
I found Ghislain OPDENAKKER listed as Professor of Immunology and Head of Immunobiology at KU Leuven. He might be the one. But I need to confirm if he published in Annual Review of Immunology in 2003. Let me search more specifically for him and Annual Review of Immunology 2003. Actually, the 2003 Annual Review of Immunology is Volume 21. Let me search for Volume 21 of Annual Review of Immunology and look for a chapter about cytokine receptors.\par Actually, let me reconsider the scholar. The description says "long taught immunology at the Catholic University of Leuven in Belgium" - this could be some ...
\end{tcolorbox}
\begin{tcolorbox}[
  colback=callbg, colframe=callhead, colbacktitle=callhead, coltitle=white,
  boxrule=0.45pt, arc=1.4mm, left=5pt,right=5pt,top=4pt,bottom=4pt,
  toptitle=1.2pt,bottomtitle=1.2pt, fonttitle=\sffamily\bfseries\small,
  fontupper=\footnotesize\sffamily, breakable,
  title={Tool call $|$ web\_search $|$ Round 14}, before skip=3.5pt, after skip=3.5pt,
]
{\ttfamily\small \{"query": ["Jean-Christophe Renauld Ludwig Institute cancer immunology", "Jean-Christophe Renauld Annual Review of Immunology 2003", "Jean-Christophe Renauld cytokine receptor"]\}}
\end{tcolorbox}
\begin{tcolorbox}[
  colback=obsbg, colframe=obshead, colbacktitle=obshead, coltitle=white,
  boxrule=0.45pt, arc=1.4mm, left=5pt,right=5pt,top=4pt,bottom=4pt,
  toptitle=1.2pt,bottomtitle=1.2pt, fonttitle=\sffamily\bfseries\small,
  fontupper=\footnotesize\sffamily, breakable,
  title={Tool response $|$ Round 14}, before skip=3.5pt, after skip=3.5pt,
]
{\sffamily\bfseries Summary preview}
\par{\sffamily\color{thinkhead}\bfseries Jean-Christophe Renauld - Professor at the Universite ...}
\par{\scriptsize\ttfamily\color{black!55} https://be.linkedin.com/***}
\par{\sffamily\color{thinkhead}\bfseries Jean-Christophe RENAULD | Professor | MD, PhD}
\par{\scriptsize\ttfamily\color{black!55} https://www.researchgate.net/***}
\par{\sffamily\color{thinkhead}\bfseries Masters of the JAKs}
\par{\scriptsize\ttfamily\color{black!55} https://www.ludwigcancerresearch.org/***}
\par\vspace{2pt}{\scriptsize\sffamily\color{black!50} $\triangleright$ Full tool return (4,270 characters)}
\end{tcolorbox}
\begin{tcolorbox}[
  colback=thinkbg, colframe=thinkhead, colbacktitle=thinkhead, coltitle=white,
  boxrule=0.45pt, arc=1.4mm, left=5pt,right=5pt,top=4pt,bottom=4pt,
  toptitle=1.2pt,bottomtitle=1.2pt, fonttitle=\sffamily\bfseries\small,
  fontupper=\footnotesize\sffamily, breakable,
  title={Think $|$ Round 15}, before skip=3.5pt, after skip=3.5pt,
]
Great! I found Jean-Christophe Renauld. The PubMed result shows: "Class II cytokine receptors and their ligands: key antiviral and inflammatory modulators" by JC Renauld, 2003, Rev Immunol. 2003 Aug;3(8):667-76. But wait, the user said "leading immunology review journal" and "2003". This is "Reviews in Immunology" which might be a review journal. But the user mentioned "leading immunology review journal" - could be Annual Review of Immunology, Nature Immunology, Immunity, or Reviews in Immunology.\par Wait, let me look at the search result more carefully. It says "by JC Renauld  2003  Cited by 395 ...
\end{tcolorbox}
\begin{tcolorbox}[
  colback=callbg, colframe=callhead, colbacktitle=callhead, coltitle=white,
  boxrule=0.45pt, arc=1.4mm, left=5pt,right=5pt,top=4pt,bottom=4pt,
  toptitle=1.2pt,bottomtitle=1.2pt, fonttitle=\sffamily\bfseries\small,
  fontupper=\footnotesize\sffamily, breakable,
  title={Tool call $|$ web\_extract $|$ Round 15}, before skip=3.5pt, after skip=3.5pt,
]
{\ttfamily\small \{"urls": ["https://pubmed.ncbi.nlm.nih.gov/***"], "goal": "Find details about the article by Jean-Christophe Renauld from 2003, including journal name, title, and any descriptions of figures showing receptor families and schematic diagrams. Also check if this is the article described in the user's question."\}}
\end{tcolorbox}
\begin{tcolorbox}[
  colback=obsbg, colframe=obshead, colbacktitle=obshead, coltitle=white,
  boxrule=0.45pt, arc=1.4mm, left=5pt,right=5pt,top=4pt,bottom=4pt,
  toptitle=1.2pt,bottomtitle=1.2pt, fonttitle=\sffamily\bfseries\small,
  fontupper=\footnotesize\sffamily, breakable,
  title={Tool response $|$ Round 15}, before skip=3.5pt, after skip=3.5pt,
]
{\sffamily\bfseries Summary preview}
\par{\sffamily\color{thinkhead}\bfseries Class II cytokine receptors and their ligands: key antiviral and inflammatory modulators - PubMed}
\par{\scriptsize\ttfamily\color{black!55} https://pubmed.ncbi.nlm.nih.gov/***}
\par\vspace{2pt}{\scriptsize\sffamily\color{black!50} $\triangleright$ Full tool return (1,236 characters)}
\end{tcolorbox}
\begin{tcolorbox}[
  colback=thinkbg, colframe=thinkhead, colbacktitle=thinkhead, coltitle=white,
  boxrule=0.45pt, arc=1.4mm, left=5pt,right=5pt,top=4pt,bottom=4pt,
  toptitle=1.2pt,bottomtitle=1.2pt, fonttitle=\sffamily\bfseries\small,
  fontupper=\footnotesize\sffamily, breakable,
  title={Think $|$ Round 16}, before skip=3.5pt, after skip=3.5pt,
]
The article is "Class II cytokine receptors and their ligands: key antiviral and inflammatory modulators" published in Nature Reviews Immunology in August 2003. This matches the description: Jean-Christophe Renauld is a professor at the Universite catholique de Louvain (KU Leuven, Catholic University of Leuven) and is affiliated with the Ludwig Institute for Cancer Research. He has published reviews on cytokine receptors.\par Now I need to find the specific figure from this article that shows "multiple members of receptor families involved in antiviral and inflammatory regulation" and "structures of ...
\end{tcolorbox}
\begin{tcolorbox}[
  colback=callbg, colframe=callhead, colbacktitle=callhead, coltitle=white,
  boxrule=0.45pt, arc=1.4mm, left=5pt,right=5pt,top=4pt,bottom=4pt,
  toptitle=1.2pt,bottomtitle=1.2pt, fonttitle=\sffamily\bfseries\small,
  fontupper=\footnotesize\sffamily, breakable,
  title={Tool call $|$ web\_search $|$ Round 16}, before skip=3.5pt, after skip=3.5pt,
]
{\ttfamily\small \{"query": ["Class II cytokine receptors and their ligands: key antiviral and inflammatory modulators figure schematic", "Renauld Nature Reviews Immunology 2003 receptor families schematic", "Nature Reviews Immunology 2003 \textbackslash{}"Class II cytokine receptors\textbackslash{}" figure", "Renauld 2003 Nature Reviews Immunology cytok ...}
\end{tcolorbox}
\begin{tcolorbox}[
  colback=obsbg, colframe=obshead, colbacktitle=obshead, coltitle=white,
  boxrule=0.45pt, arc=1.4mm, left=5pt,right=5pt,top=4pt,bottom=4pt,
  toptitle=1.2pt,bottomtitle=1.2pt, fonttitle=\sffamily\bfseries\small,
  fontupper=\footnotesize\sffamily, breakable,
  title={Tool response $|$ Round 16}, before skip=3.5pt, after skip=3.5pt,
]
{\sffamily\bfseries Summary preview}
\par{\sffamily\color{thinkhead}\bfseries Class II cytokine receptors and their ligands: key antiviral ...}
\par{\scriptsize\ttfamily\color{black!55} https://pubmed.ncbi.nlm.nih.gov/***}
\par{\sffamily\color{thinkhead}\bfseries (PDF) Class II cytokine receptors and their ligands}
\par{\scriptsize\ttfamily\color{black!55} https://www.researchgate.net/***}
\par{\sffamily\color{thinkhead}\bfseries Class II cytokine receptors and their ligands: Key antiviral ...}
\par{\scriptsize\ttfamily\color{black!55} https://www.scilit.com/***}
\par\vspace{2pt}{\scriptsize\sffamily\color{black!50} $\triangleright$ Full tool return (5,916 characters)}
\end{tcolorbox}
\begin{tcolorbox}[
  colback=thinkbg, colframe=thinkhead, colbacktitle=thinkhead, coltitle=white,
  boxrule=0.45pt, arc=1.4mm, left=5pt,right=5pt,top=4pt,bottom=4pt,
  toptitle=1.2pt,bottomtitle=1.2pt, fonttitle=\sffamily\bfseries\small,
  fontupper=\footnotesize\sffamily, breakable,
  title={Think $|$ Round 17}, before skip=3.5pt, after skip=3.5pt,
]
I need to find the actual figures from this article. Let me try to access the PDF or the full text through ResearchGate or other sources. I should look for the figure showing receptor families and transmembrane complexes. Let me search for images or the specific figure from this paper.
\end{tcolorbox}
\begin{tcolorbox}[
  colback=callbg, colframe=callhead, colbacktitle=callhead, coltitle=white,
  boxrule=0.45pt, arc=1.4mm, left=5pt,right=5pt,top=4pt,bottom=4pt,
  toptitle=1.2pt,bottomtitle=1.2pt, fonttitle=\sffamily\bfseries\small,
  fontupper=\footnotesize\sffamily, breakable,
  title={Tool call $|$ web\_search $|$ Round 17}, before skip=3.5pt, after skip=3.5pt,
]
{\ttfamily\small \{"query": ["site:researchgate.net \textbackslash{}"Class II cytokine receptors\textbackslash{}" figure", "Renauld 2003 Nature Reviews Immunology schematic receptor families", "Nature Reviews Immunology 2003 figure cytokine receptor complex membrane color"]\}}
\end{tcolorbox}
\begin{tcolorbox}[
  colback=obsbg, colframe=obshead, colbacktitle=obshead, coltitle=white,
  boxrule=0.45pt, arc=1.4mm, left=5pt,right=5pt,top=4pt,bottom=4pt,
  toptitle=1.2pt,bottomtitle=1.2pt, fonttitle=\sffamily\bfseries\small,
  fontupper=\footnotesize\sffamily, breakable,
  title={Tool response $|$ Round 17}, before skip=3.5pt, after skip=3.5pt,
]
{\sffamily\bfseries Summary preview}
\par{\sffamily\color{thinkhead}\bfseries (PDF) Class II cytokine receptors and their ligands}
\par{\scriptsize\ttfamily\color{black!55} https://www.researchgate.net/***}
\par{\sffamily\color{thinkhead}\bfseries Phylogenetic tree of class II cytokines and receptors.The...}
\par{\scriptsize\ttfamily\color{black!55} https://www.researchgate.net/***}
\par{\sffamily\color{thinkhead}\bfseries Expression pattern for the classII helical cytokine receptor ...}
\par{\scriptsize\ttfamily\color{black!55} https://www.researchgate.net/***}
\par\vspace{2pt}{\scriptsize\sffamily\color{black!50} $\triangleright$ Full tool return (4,586 characters)}
\end{tcolorbox}
\begin{tcolorbox}[
  colback=thinkbg, colframe=thinkhead, colbacktitle=thinkhead, coltitle=white,
  boxrule=0.45pt, arc=1.4mm, left=5pt,right=5pt,top=4pt,bottom=4pt,
  toptitle=1.2pt,bottomtitle=1.2pt, fonttitle=\sffamily\bfseries\small,
  fontupper=\footnotesize\sffamily, breakable,
  title={Think $|$ Round 18}, before skip=3.5pt, after skip=3.5pt,
]
Let me try to access the ResearchGate publication directly to see the figures. I need to find the specific figure that shows receptor families and transmembrane complexes with colored membrane-embedded portions.
\end{tcolorbox}
\begin{tcolorbox}[
  colback=callbg, colframe=callhead, colbacktitle=callhead, coltitle=white,
  boxrule=0.45pt, arc=1.4mm, left=5pt,right=5pt,top=4pt,bottom=4pt,
  toptitle=1.2pt,bottomtitle=1.2pt, fonttitle=\sffamily\bfseries\small,
  fontupper=\footnotesize\sffamily, breakable,
  title={Tool call $|$ web\_extract $|$ Round 18}, before skip=3.5pt, after skip=3.5pt,
]
{\ttfamily\small \{"urls": ["https://www.researchgate.net/***"], "goal": "Find any figures from this article, especially a schematic showing multiple receptor families and transmembrane complexes with colored portions of ...}
\end{tcolorbox}
\begin{tcolorbox}[
  colback=obsbg, colframe=obshead, colbacktitle=obshead, coltitle=white,
  boxrule=0.45pt, arc=1.4mm, left=5pt,right=5pt,top=4pt,bottom=4pt,
  toptitle=1.2pt,bottomtitle=1.2pt, fonttitle=\sffamily\bfseries\small,
  fontupper=\footnotesize\sffamily, breakable,
  title={Tool response $|$ Round 18}, before skip=3.5pt, after skip=3.5pt,
]
{\sffamily\bfseries Summary preview}
\par{\sffamily\color{thinkhead}\bfseries Just a moment...}
\par{\scriptsize\ttfamily\color{black!55} https://www.researchgate.net/***}
\par\vspace{2pt}{\scriptsize\sffamily\color{black!50} $\triangleright$ Full tool return (626 characters)}
\end{tcolorbox}
\begin{tcolorbox}[
  colback=thinkbg, colframe=thinkhead, colbacktitle=thinkhead, coltitle=white,
  boxrule=0.45pt, arc=1.4mm, left=5pt,right=5pt,top=4pt,bottom=4pt,
  toptitle=1.2pt,bottomtitle=1.2pt, fonttitle=\sffamily\bfseries\small,
  fontupper=\footnotesize\sffamily, breakable,
  title={Think $|$ Round 19}, before skip=3.5pt, after skip=3.5pt,
]
The ResearchGate page is blocked. Let me try to find the figure through other sources. I can search for the specific figure or try to access the PDF through the Springer link or other repositories. Let me also search for images from this paper.
\end{tcolorbox}
\begin{tcolorbox}[
  colback=callbg, colframe=callhead, colbacktitle=callhead, coltitle=white,
  boxrule=0.45pt, arc=1.4mm, left=5pt,right=5pt,top=4pt,bottom=4pt,
  toptitle=1.2pt,bottomtitle=1.2pt, fonttitle=\sffamily\bfseries\small,
  fontupper=\footnotesize\sffamily, breakable,
  title={Tool call $|$ web\_search\_for\_image $|$ Round 19}, before skip=3.5pt, after skip=3.5pt,
]
{\ttfamily\small \{"query": ["Class II cytokine receptors and their ligands Nature Reviews Immunology 2003 figure", "Renauld 2003 cytokine receptor schematic", "Nature Reviews Immunology 2003 receptor families diagram"]\}}
\end{tcolorbox}
\begin{tcolorbox}[
  colback=obsbg, colframe=obshead, colbacktitle=obshead, coltitle=white,
  boxrule=0.45pt, arc=1.4mm, left=5pt,right=5pt,top=4pt,bottom=4pt,
  toptitle=1.2pt,bottomtitle=1.2pt, fonttitle=\sffamily\bfseries\small,
  fontupper=\footnotesize\sffamily, breakable,
  title={Tool response $|$ Round 19}, before skip=3.5pt, after skip=3.5pt,
]
{\sffamily\bfseries Summary preview}
\par Tool returned a structured payload; only the preview is shown.
\par\vspace{3pt}\noindent \begin{minipage}[t]{0.28\linewidth}\centering\includegraphics[width=\linewidth,height=0.11\textheight,keepaspectratio]{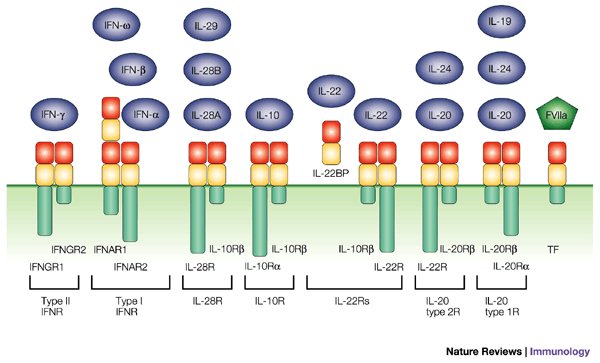}\\{\scriptsize\ttfamily search\_21d0f33d.jpg}\end{minipage}\par\vspace{2pt}{\scriptsize\sffamily\color{black!50} $\triangleright$ Full tool return (7,498 characters)}
\end{tcolorbox}
\begin{tcolorbox}[
  colback=thinkbg, colframe=thinkhead, colbacktitle=thinkhead, coltitle=white,
  boxrule=0.45pt, arc=1.4mm, left=5pt,right=5pt,top=4pt,bottom=4pt,
  toptitle=1.2pt,bottomtitle=1.2pt, fonttitle=\sffamily\bfseries\small,
  fontupper=\footnotesize\sffamily, breakable,
  title={Think $|$ Final}, before skip=3.5pt, after skip=3.5pt,
]
I found the key figure! The second image (search\_21d0f33d.jpg) shows exactly what the user described: a schematic systematically depicting multiple members of receptor families involved in antiviral and inflammatory regulation, and the structures of their transmembrane complexes.\par Looking at this figure, the portions of the receptor complexes embedded in the cell membrane are shown in {\bfseries green} color ...
\end{tcolorbox}
\begin{tcolorbox}[
  colback=webg, colframe=wehead, colbacktitle=wehead, coltitle=white,
  boxrule=0.45pt, arc=1.4mm, left=5pt,right=5pt,top=4pt,bottom=4pt,
  toptitle=1.2pt,bottomtitle=1.2pt, fonttitle=\sffamily\bfseries\small,
  fontupper=\footnotesize\sffamily, breakable,
  title={Final response}, before skip=3.5pt, after skip=3.5pt,
]
Based on my research, the scholar is {\bfseries Jean-Christophe Renauld}, a professor at the Catholic University of Leuven (KU Leuven) in Belgium and a researcher at the Ludwig Institute for Cancer Research. In 2003, he published a review article titled "Class II cytokine receptors and their ligands: key antiviral and inflammatory modulators" in *Nature Reviews Immunology*.\par The schematic you're referring to is a figure from this article that systematically depicts multiple members of Class II cytokine receptor families (involved in antiviral and inflammatory regulation) and the structures of their transmembrane complexes. In this figure, the portions of the receptor complexes embedded in the cell membrane are shown in {\bfseries green} (or a greenish/\allowbreak{}teal color).\par The figure shows various Class II cytokine receptors (such as IFNGR2/\allowbreak{}IFNAR1, IFNAR2, IL-28RA, IL-10R, IL-10R, IL-22R, IL-20R, IL-20R, and TF) with their characteristic transmembrane domains passing through the cell membrane, which are depicted in green. The extracellular receptor domains are shown in blue/\allowbreak{}purple, while the transmembrane portions are colored green.
\par\vspace{4pt}\noindent \begin{minipage}[t]{0.55\linewidth}\centering\includegraphics[width=\linewidth,height=0.18\textheight,keepaspectratio]{images/rl_response_cases/case_0094_search_21d0f33d.jpg}\\{\scriptsize\ttfamily search\_21d0f33d.jpg}\end{minipage}
\end{tcolorbox}
\clearpage
\phantomsection\label{app:rl-case-3}
\noindent{\sffamily\bfseries\large\color{wehead} Case 3: Game-lineage, award, and screenshot reasoning}\par
\vspace{2pt}
\respquery{This two-dimensional puzzle game, released in 2019, uses impossible architecture as its core visual style; its game world creates the illusion of infinite extension by looping and folding through three-dimensional space. The development team later released a new VR title that prioritizes a room-scale experience. What award did the new title receive in February 2026? In addition, in an official screenshot from the new title, a large, fluid-like bright-red object stands in the center of a hall; what shape is the light-blue decoration to its left?}
{\small\sffamily\itshape Reference answer: In February 2026, it received: the 29th Annual D.I.C.E. Award for Immersive Reality Technical Achievement; the shape of the light-blue decoration on the left is: a water-drop shape.}\par
\begin{center}
{\sffamily\bfseries\small\color{wehead} Target image}\\[-1pt]
{\sffamily\scriptsize\color{black!50} (not provided to the model at query time)}\\[3pt]
\includegraphics[width=0.92\linewidth,height=0.15\textheight,keepaspectratio]{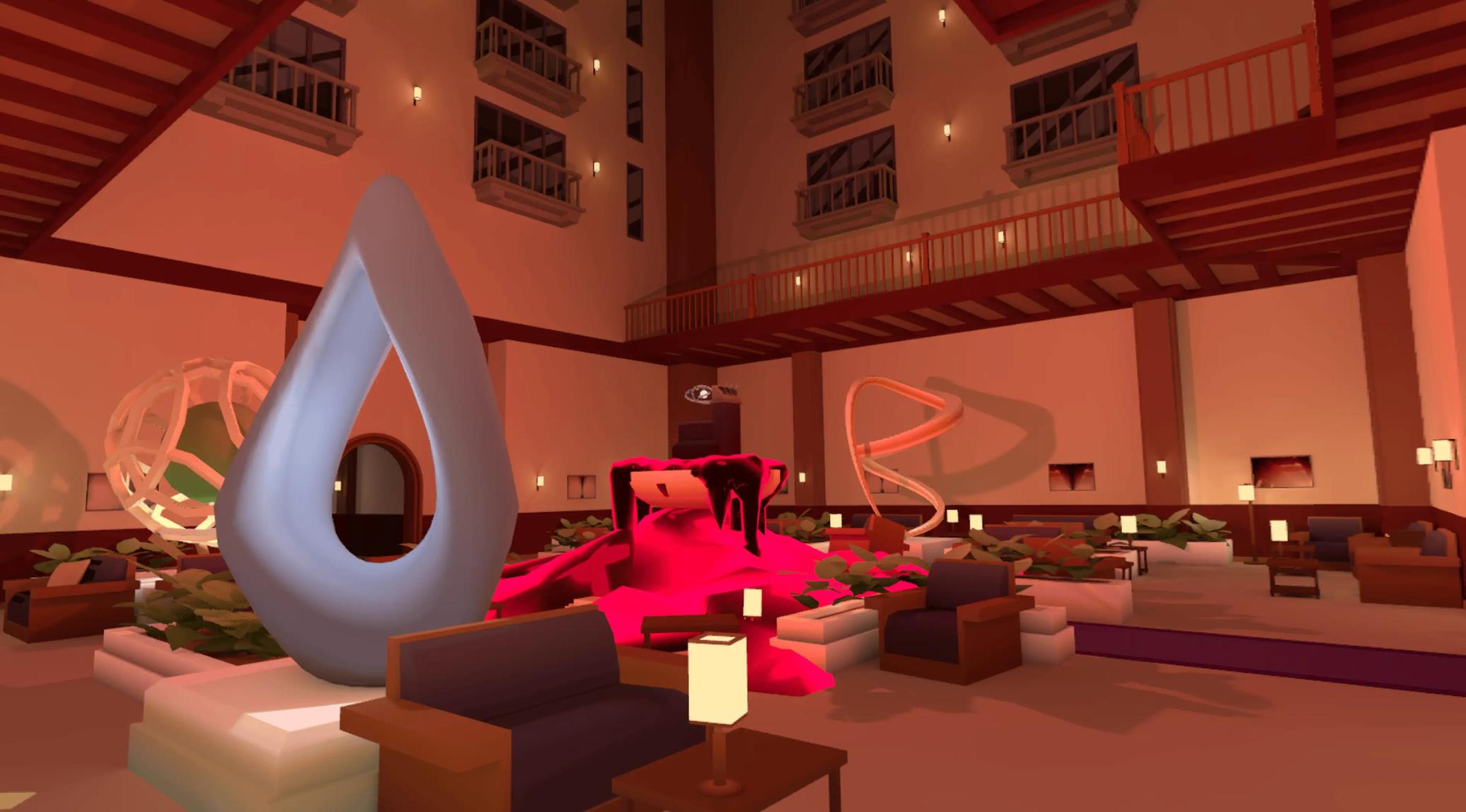}
\end{center}
\begin{tcolorbox}[
  colback=thinkbg, colframe=thinkhead, colbacktitle=thinkhead, coltitle=white,
  boxrule=0.45pt, arc=1.4mm, left=5pt,right=5pt,top=4pt,bottom=4pt,
  toptitle=1.2pt,bottomtitle=1.2pt, fonttitle=\sffamily\bfseries\small,
  fontupper=\footnotesize\sffamily, breakable,
  title={Think $|$ Round 1}, before skip=3.5pt, after skip=3.5pt,
]
The user is asking about a specific puzzle game released in 2019 with impossible architecture as its core visual style, which creates the illusion of infinite extension by looping and folding through three-dimensional space. The development team later released a new VR title that prioritizes a room-scale experience. I need to identify: 1. The award the new title received in February 2026 2. In an official screenshot from the new title, a large, fluid-like bright-red object stands in the center of a hall; what shape is the light-blue decoration to its left?\par First, I need to identify the game. Let ...
\end{tcolorbox}
\begin{tcolorbox}[
  colback=callbg, colframe=callhead, colbacktitle=callhead, coltitle=white,
  boxrule=0.45pt, arc=1.4mm, left=5pt,right=5pt,top=4pt,bottom=4pt,
  toptitle=1.2pt,bottomtitle=1.2pt, fonttitle=\sffamily\bfseries\small,
  fontupper=\footnotesize\sffamily, breakable,
  title={Tool call $|$ web\_search $|$ Round 1}, before skip=3.5pt, after skip=3.5pt,
]
{\ttfamily\small \{"query": "puzzle game 2019 impossible architecture infinite extension looping folding VR sequel room-scale", "num": 5\}}
\end{tcolorbox}
\begin{tcolorbox}[
  colback=obsbg, colframe=obshead, colbacktitle=obshead, coltitle=white,
  boxrule=0.45pt, arc=1.4mm, left=5pt,right=5pt,top=4pt,bottom=4pt,
  toptitle=1.2pt,bottomtitle=1.2pt, fonttitle=\sffamily\bfseries\small,
  fontupper=\footnotesize\sffamily, breakable,
  title={Tool response $|$ Round 1}, before skip=3.5pt, after skip=3.5pt,
]
{\sffamily\bfseries Summary preview}
\par{\sffamily\color{thinkhead}\bfseries Infinite Hallways in Roomscale VR | Hotel Infinity Devlog \#2}
\par{\scriptsize\ttfamily\color{black!55} https://www.youtube.com/***}
\par{\sffamily\color{thinkhead}\bfseries Roomscale VR Has NEVER Been This Wild! Hotel Infinity}
\par{\scriptsize\ttfamily\color{black!55} https://www.youtube.com/***}
\par{\sffamily\color{thinkhead}\bfseries Building an Impossible VR Game | Hotel Infinity Devlog \#1}
\par{\scriptsize\ttfamily\color{black!55} https://www.youtube.com/***}
\par\vspace{2pt}{\scriptsize\sffamily\color{black!50} $\triangleright$ Full tool return (2,168 characters)}
\end{tcolorbox}
\begin{tcolorbox}[
  colback=thinkbg, colframe=thinkhead, colbacktitle=thinkhead, coltitle=white,
  boxrule=0.45pt, arc=1.4mm, left=5pt,right=5pt,top=4pt,bottom=4pt,
  toptitle=1.2pt,bottomtitle=1.2pt, fonttitle=\sffamily\bfseries\small,
  fontupper=\footnotesize\sffamily, breakable,
  title={Think $|$ Round 2}, before skip=3.5pt, after skip=3.5pt,
]
The search results point to "Hotel Infinity". Let me search for more details about Hotel Infinity, including its 2019 release and the VR title that prioritizes room-scale experience, as well as awards in February 2026.
\end{tcolorbox}
\begin{tcolorbox}[
  colback=callbg, colframe=callhead, colbacktitle=callhead, coltitle=white,
  boxrule=0.45pt, arc=1.4mm, left=5pt,right=5pt,top=4pt,bottom=4pt,
  toptitle=1.2pt,bottomtitle=1.2pt, fonttitle=\sffamily\bfseries\small,
  fontupper=\footnotesize\sffamily, breakable,
  title={Tool call $|$ web\_search $|$ Round 2}, before skip=3.5pt, after skip=3.5pt,
]
{\ttfamily\small \{"query": "Hotel Infinity VR game 2019 impossible architecture award February 2026", "num": 5\}}
\end{tcolorbox}
\begin{tcolorbox}[
  colback=obsbg, colframe=obshead, colbacktitle=obshead, coltitle=white,
  boxrule=0.45pt, arc=1.4mm, left=5pt,right=5pt,top=4pt,bottom=4pt,
  toptitle=1.2pt,bottomtitle=1.2pt, fonttitle=\sffamily\bfseries\small,
  fontupper=\footnotesize\sffamily, breakable,
  title={Tool response $|$ Round 2}, before skip=3.5pt, after skip=3.5pt,
]
{\sffamily\bfseries Summary preview}
\par{\sffamily\color{thinkhead}\bfseries Building an Impossible VR Game | Hotel Infinity Devlog \#1}
\par{\scriptsize\ttfamily\color{black!55} https://www.youtube.com/***}
\par{\sffamily\color{thinkhead}\bfseries Hotel Infinity  We Hope Your Stay is Eternal  Studio Chyr}
\par{\scriptsize\ttfamily\color{black!55} https://hotelinfinityvr.com/***}
\par{\sffamily\color{thinkhead}\bfseries Hotel Infinity review: mind-bending corridors, hollow ...}
\par{\scriptsize\ttfamily\color{black!55} https://www.thevrcritic.com/***}
\par\vspace{2pt}{\scriptsize\sffamily\color{black!50} $\triangleright$ Full tool return (2,256 characters)}
\end{tcolorbox}
\begin{tcolorbox}[
  colback=thinkbg, colframe=thinkhead, colbacktitle=thinkhead, coltitle=white,
  boxrule=0.45pt, arc=1.4mm, left=5pt,right=5pt,top=4pt,bottom=4pt,
  toptitle=1.2pt,bottomtitle=1.2pt, fonttitle=\sffamily\bfseries\small,
  fontupper=\footnotesize\sffamily, breakable,
  title={Think $|$ Round 3}, before skip=3.5pt, after skip=3.5pt,
]
Hotel Infinity is definitely the game. Now I need to find: 1. What award did the new title receive in February 2026? 2. In an official screenshot from the new title, a large, fluid-like bright-red object stands in the center of a hall; what shape is the light-blue decoration to its left?\par Let me search for more specific information about awards in February 2026 and also look for official screenshots.
\end{tcolorbox}
\begin{tcolorbox}[
  colback=callbg, colframe=callhead, colbacktitle=callhead, coltitle=white,
  boxrule=0.45pt, arc=1.4mm, left=5pt,right=5pt,top=4pt,bottom=4pt,
  toptitle=1.2pt,bottomtitle=1.2pt, fonttitle=\sffamily\bfseries\small,
  fontupper=\footnotesize\sffamily, breakable,
  title={Tool call $|$ web\_search $|$ Round 3}, before skip=3.5pt, after skip=3.5pt,
]
{\ttfamily\small \{"query": "Hotel Infinity award February 2026", "num": 5\}}
\end{tcolorbox}
\begin{tcolorbox}[
  colback=obsbg, colframe=obshead, colbacktitle=obshead, coltitle=white,
  boxrule=0.45pt, arc=1.4mm, left=5pt,right=5pt,top=4pt,bottom=4pt,
  toptitle=1.2pt,bottomtitle=1.2pt, fonttitle=\sffamily\bfseries\small,
  fontupper=\footnotesize\sffamily, breakable,
  title={Tool response $|$ Round 3}, before skip=3.5pt, after skip=3.5pt,
]
{\sffamily\bfseries Summary preview}
\par{\sffamily\color{thinkhead}\bfseries 42nd Annual ICP Infinity Awards}
\par{\scriptsize\ttfamily\color{black!55} https://www.icp.org/***}
\par{\sffamily\color{thinkhead}\bfseries 2026 Infinity Awards}
\par{\scriptsize\ttfamily\color{black!55} https://givebutter.com/***}
\par{\sffamily\color{thinkhead}\bfseries Receiving the 2026 ICP Infinity Award for Documentary and ...}
\par{\scriptsize\ttfamily\color{black!55} https://hsakag.substack.com/***}
\par\vspace{2pt}{\scriptsize\sffamily\color{black!50} $\triangleright$ Full tool return (2,085 characters)}
\end{tcolorbox}
\begin{tcolorbox}[
  colback=thinkbg, colframe=thinkhead, colbacktitle=thinkhead, coltitle=white,
  boxrule=0.45pt, arc=1.4mm, left=5pt,right=5pt,top=4pt,bottom=4pt,
  toptitle=1.2pt,bottomtitle=1.2pt, fonttitle=\sffamily\bfseries\small,
  fontupper=\footnotesize\sffamily, breakable,
  title={Think $|$ Round 4}, before skip=3.5pt, after skip=3.5pt,
]
The search didn't find the specific award. Let me try different queries - maybe it's an award at a specific event like Gamescom, The Game Awards, or something like that. Also the user mentions "new VR title that prioritizes a room-scale experience" - this could be Hotel Infinity itself, or maybe a sequel? Let me search more specifically.
\end{tcolorbox}
\begin{tcolorbox}[
  colback=callbg, colframe=callhead, colbacktitle=callhead, coltitle=white,
  boxrule=0.45pt, arc=1.4mm, left=5pt,right=5pt,top=4pt,bottom=4pt,
  toptitle=1.2pt,bottomtitle=1.2pt, fonttitle=\sffamily\bfseries\small,
  fontupper=\footnotesize\sffamily, breakable,
  title={Tool call $|$ web\_search $|$ Round 4}, before skip=3.5pt, after skip=3.5pt,
]
{\ttfamily\small \{"query": ["Hotel Infinity VR award 2026", "Hotel Infinity Steam Awards", "Hotel Infinity accolades 2026"], "num": 5\}}
\end{tcolorbox}
\begin{tcolorbox}[
  colback=obsbg, colframe=obshead, colbacktitle=obshead, coltitle=white,
  boxrule=0.45pt, arc=1.4mm, left=5pt,right=5pt,top=4pt,bottom=4pt,
  toptitle=1.2pt,bottomtitle=1.2pt, fonttitle=\sffamily\bfseries\small,
  fontupper=\footnotesize\sffamily, breakable,
  title={Tool response $|$ Round 4}, before skip=3.5pt, after skip=3.5pt,
]
{\sffamily\bfseries Summary preview}
\par{\sffamily\color{thinkhead}\bfseries HOTEL INFINITY by William Chyr - XRGuide}
\par{\scriptsize\ttfamily\color{black!55} https://xrguide.app/***}
\par{\sffamily\color{thinkhead}\bfseries William Chyr's Post}
\par{\scriptsize\ttfamily\color{black!55} https://www.linkedin.com/***}
\par{\sffamily\color{thinkhead}\bfseries Hotel Infinity review: mind-bending corridors, hollow ...}
\par{\scriptsize\ttfamily\color{black!55} https://www.thevrcritic.com/***}
\par\vspace{2pt}{\scriptsize\sffamily\color{black!50} $\triangleright$ Full tool return (5,723 characters)}
\end{tcolorbox}
\begin{tcolorbox}[
  colback=thinkbg, colframe=thinkhead, colbacktitle=thinkhead, coltitle=white,
  boxrule=0.45pt, arc=1.4mm, left=5pt,right=5pt,top=4pt,bottom=4pt,
  toptitle=1.2pt,bottomtitle=1.2pt, fonttitle=\sffamily\bfseries\small,
  fontupper=\footnotesize\sffamily, breakable,
  title={Think $|$ Round 5}, before skip=3.5pt, after skip=3.5pt,
]
I found some relevant information: 1. Hotel Infinity won the "Immersive Reality Technical Achievement" at the DICE Awards (29th DICE Awards) 2. It won the "Auggie Award Winner --- Best Indie Creator(s)" at AWE USA 2026\par But the user specifically asked about an award in February 2026. Let me search more specifically for February 2026 awards and also look for official screenshots with the described scene.
\end{tcolorbox}
\begin{tcolorbox}[
  colback=callbg, colframe=callhead, colbacktitle=callhead, coltitle=white,
  boxrule=0.45pt, arc=1.4mm, left=5pt,right=5pt,top=4pt,bottom=4pt,
  toptitle=1.2pt,bottomtitle=1.2pt, fonttitle=\sffamily\bfseries\small,
  fontupper=\footnotesize\sffamily, breakable,
  title={Tool call $|$ web\_search $|$ Round 5}, before skip=3.5pt, after skip=3.5pt,
]
{\ttfamily\small \{"query": ["Hotel Infinity February 2026 award", "Hotel Infinity DICE Awards 2026 February", "Hotel Infinity Steam Awards February 2026"], "num": 5\}}
\end{tcolorbox}
\begin{tcolorbox}[
  colback=obsbg, colframe=obshead, colbacktitle=obshead, coltitle=white,
  boxrule=0.45pt, arc=1.4mm, left=5pt,right=5pt,top=4pt,bottom=4pt,
  toptitle=1.2pt,bottomtitle=1.2pt, fonttitle=\sffamily\bfseries\small,
  fontupper=\footnotesize\sffamily, breakable,
  title={Tool response $|$ Round 5}, before skip=3.5pt, after skip=3.5pt,
]
{\sffamily\bfseries Summary preview}
\par{\sffamily\color{thinkhead}\bfseries Only a few days are left to cast your votes in this year's @ ...}
\par{\scriptsize\ttfamily\color{black!55} https://www.instagram.com/***}
\par{\sffamily\color{thinkhead}\bfseries AIDAC INFINITY AWARD 2026 Hospitality Architecture}
\par{\scriptsize\ttfamily\color{black!55} https://www.instagram.com/***}
\par{\sffamily\color{thinkhead}\bfseries Celebrity Infinity Member Review}
\par{\scriptsize\ttfamily\color{black!55} https://www.cruisecritic.com/***}
\par\vspace{2pt}{\scriptsize\sffamily\color{black!50} $\triangleright$ Full tool return (4,803 characters)}
\end{tcolorbox}
\begin{tcolorbox}[
  colback=thinkbg, colframe=thinkhead, colbacktitle=thinkhead, coltitle=white,
  boxrule=0.45pt, arc=1.4mm, left=5pt,right=5pt,top=4pt,bottom=4pt,
  toptitle=1.2pt,bottomtitle=1.2pt, fonttitle=\sffamily\bfseries\small,
  fontupper=\footnotesize\sffamily, breakable,
  title={Think $|$ Round 6}, before skip=3.5pt, after skip=3.5pt,
]
Great! I found that Hotel Infinity won the "Immersive Reality Technical Achievement" at the 29th D.I.C.E. Awards on February 12, 2026. That's the award in February 2026.\par Now I need to find the official screenshot with the described scene: a large, fluid-like bright-red object in the center of a hall, and a light-blue decoration to its left. Let me search for official screenshots of Hotel Infinity.
\end{tcolorbox}
\begin{tcolorbox}[
  colback=callbg, colframe=callhead, colbacktitle=callhead, coltitle=white,
  boxrule=0.45pt, arc=1.4mm, left=5pt,right=5pt,top=4pt,bottom=4pt,
  toptitle=1.2pt,bottomtitle=1.2pt, fonttitle=\sffamily\bfseries\small,
  fontupper=\footnotesize\sffamily, breakable,
  title={Tool call $|$ web\_search\_for\_image $|$ Round 6}, before skip=3.5pt, after skip=3.5pt,
]
{\ttfamily\small \{"query": "Hotel Infinity official screenshot hall red object blue decoration", "num": 10\}}
\end{tcolorbox}
\begin{tcolorbox}[
  colback=obsbg, colframe=obshead, colbacktitle=obshead, coltitle=white,
  boxrule=0.45pt, arc=1.4mm, left=5pt,right=5pt,top=4pt,bottom=4pt,
  toptitle=1.2pt,bottomtitle=1.2pt, fonttitle=\sffamily\bfseries\small,
  fontupper=\footnotesize\sffamily, breakable,
  title={Tool response $|$ Round 6}, before skip=3.5pt, after skip=3.5pt,
]
{\sffamily\bfseries Summary preview}
\par Tool returned a structured payload; only the preview is shown.
\par\vspace{3pt}\noindent \begin{minipage}[t]{0.28\linewidth}\centering\includegraphics[width=\linewidth,height=0.11\textheight,keepaspectratio]{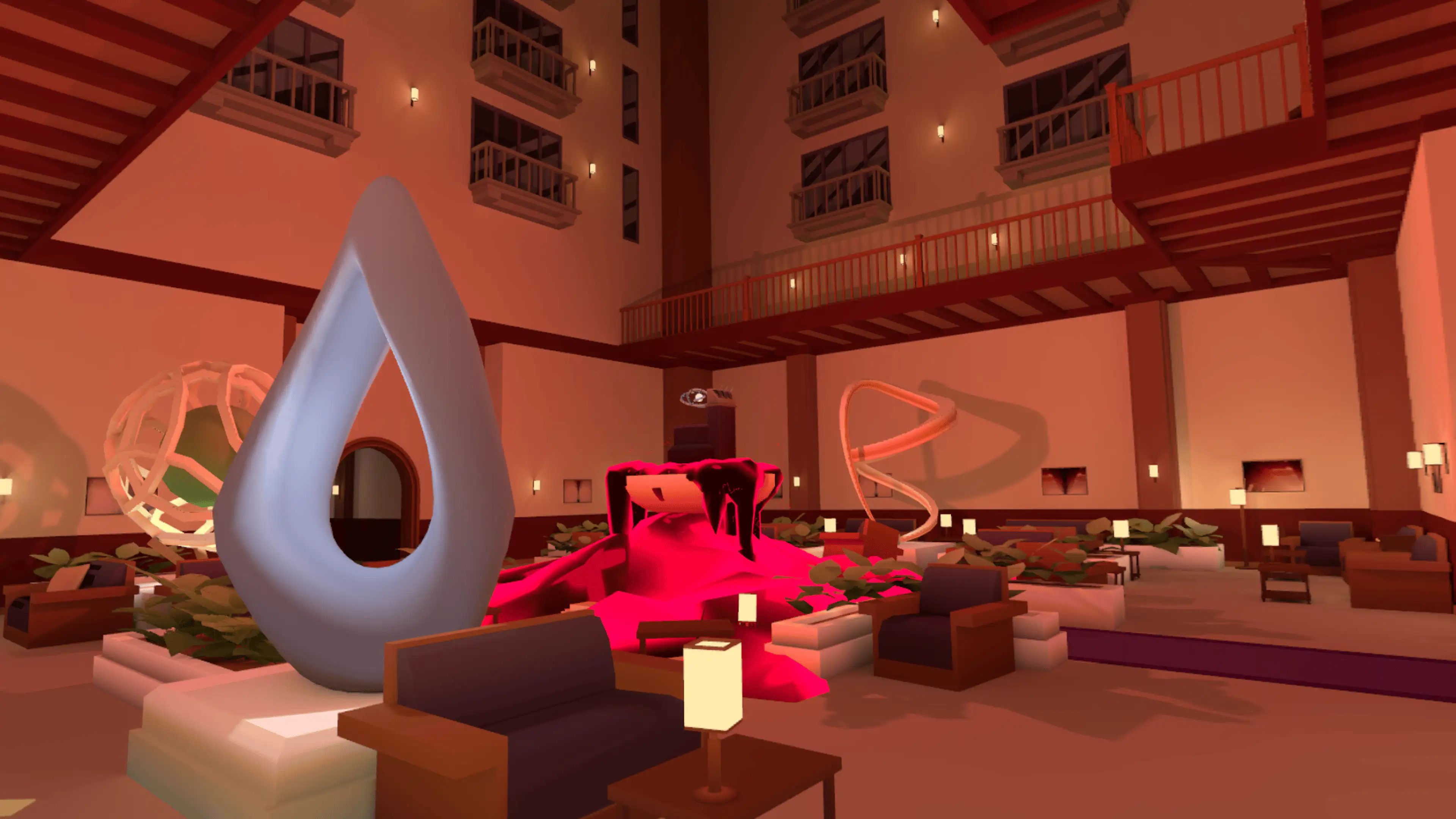}\\{\scriptsize\ttfamily search\_b5af4025.jpg}\end{minipage}\hfill \begin{minipage}[t]{0.28\linewidth}\centering\includegraphics[width=\linewidth,height=0.11\textheight,keepaspectratio]{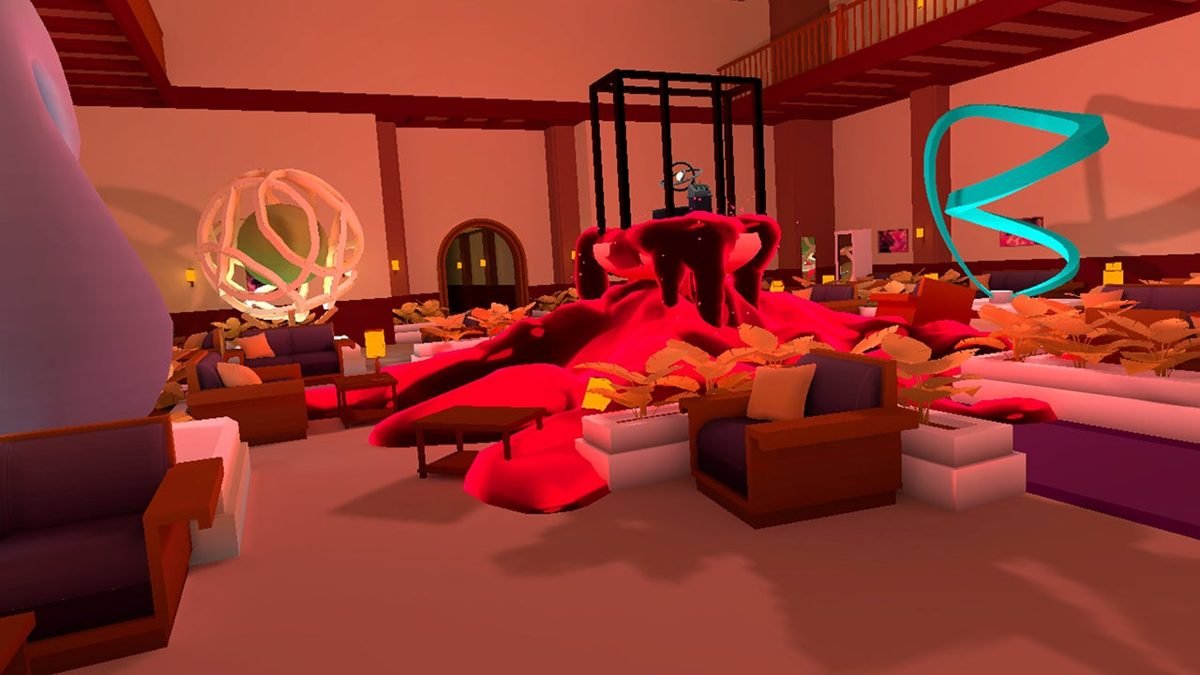}\\{\scriptsize\ttfamily search\_1a3eb628.jpg}\end{minipage}\par\vspace{2pt}{\scriptsize\sffamily\color{black!50} $\triangleright$ Full tool return (2,460 characters)}
\end{tcolorbox}
\begin{tcolorbox}[
  colback=thinkbg, colframe=thinkhead, colbacktitle=thinkhead, coltitle=white,
  boxrule=0.45pt, arc=1.4mm, left=5pt,right=5pt,top=4pt,bottom=4pt,
  toptitle=1.2pt,bottomtitle=1.2pt, fonttitle=\sffamily\bfseries\small,
  fontupper=\footnotesize\sffamily, breakable,
  title={Think $|$ Round 7}, before skip=3.5pt, after skip=3.5pt,
]
I found some images. Let me examine them more carefully. Image 1 shows a large light-blue (teal/\allowbreak{}cyan) teardrop-shaped object on the left side of what appears to be a lobby or atrium, with a red fluid-like object in the center. Image 2 shows a different room. Image 3 shows a bedroom. Image 5 shows a reception desk. Image 9 shows another lobby scene.\par Image 1 and Image 9 both show a large red fluid-like object in the center of a hall. In Image 1, to the left of the red object, there's a large light-blue teardrop-shaped object. In Image 9, there's a cyan/\allowbreak{}blue spiral-shaped decoration to the right of ...
\end{tcolorbox}
\begin{tcolorbox}[
  colback=callbg, colframe=callhead, colbacktitle=callhead, coltitle=white,
  boxrule=0.45pt, arc=1.4mm, left=5pt,right=5pt,top=4pt,bottom=4pt,
  toptitle=1.2pt,bottomtitle=1.2pt, fonttitle=\sffamily\bfseries\small,
  fontupper=\footnotesize\sffamily, breakable,
  title={Tool call $|$ web\_extract $|$ Round 7}, before skip=3.5pt, after skip=3.5pt,
]
{\ttfamily\small \{"urls": ["https://hotelinfinityvr.com/***"], "goal": "Find official screenshots of Hotel Infinity, especially any showing a large red fluid-like object in a hall with a light-blue decoration"\}}
\end{tcolorbox}
\begin{tcolorbox}[
  colback=obsbg, colframe=obshead, colbacktitle=obshead, coltitle=white,
  boxrule=0.45pt, arc=1.4mm, left=5pt,right=5pt,top=4pt,bottom=4pt,
  toptitle=1.2pt,bottomtitle=1.2pt, fonttitle=\sffamily\bfseries\small,
  fontupper=\footnotesize\sffamily, breakable,
  title={Tool response $|$ Round 7}, before skip=3.5pt, after skip=3.5pt,
]
{\sffamily\bfseries Summary preview}
\par{\sffamily\color{thinkhead}\bfseries Hotel Infinity}
\par{\scriptsize\ttfamily\color{black!55} https://hotelinfinityvr.com/***}
\par\vspace{3pt}\noindent \begin{minipage}[t]{0.16\linewidth}\centering\includegraphics[width=\linewidth,height=0.11\textheight,keepaspectratio]{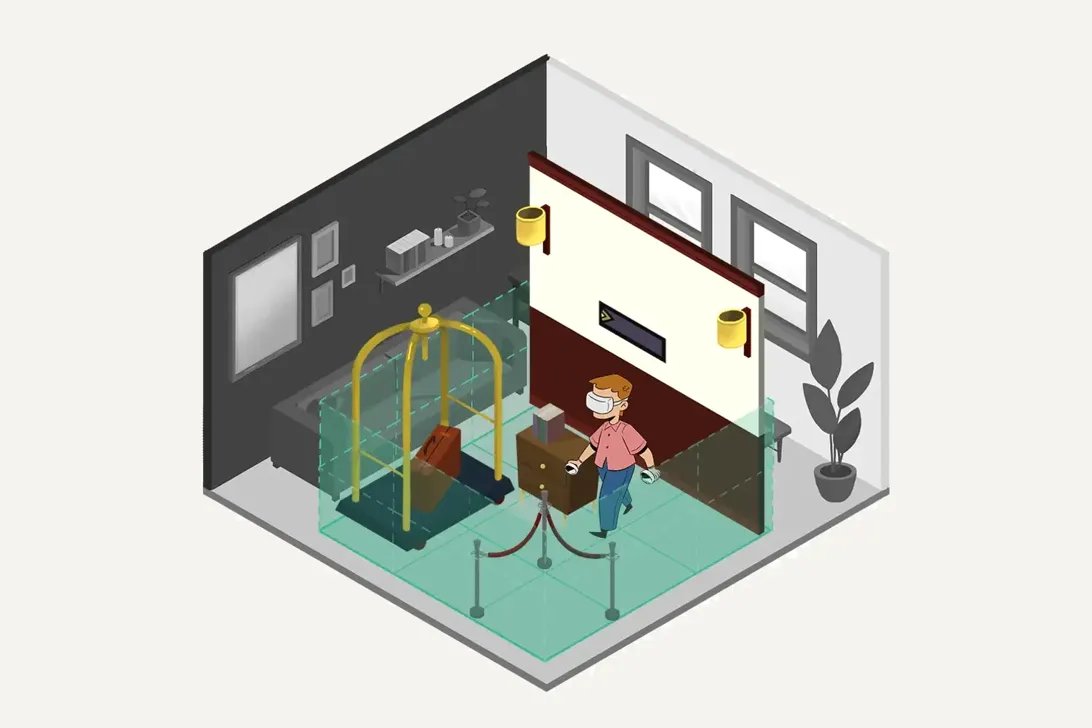}\\{\scriptsize\ttfamily extract\_458eb29c.jpg}\end{minipage}\hfill \begin{minipage}[t]{0.16\linewidth}\centering\includegraphics[width=\linewidth,height=0.11\textheight,keepaspectratio]{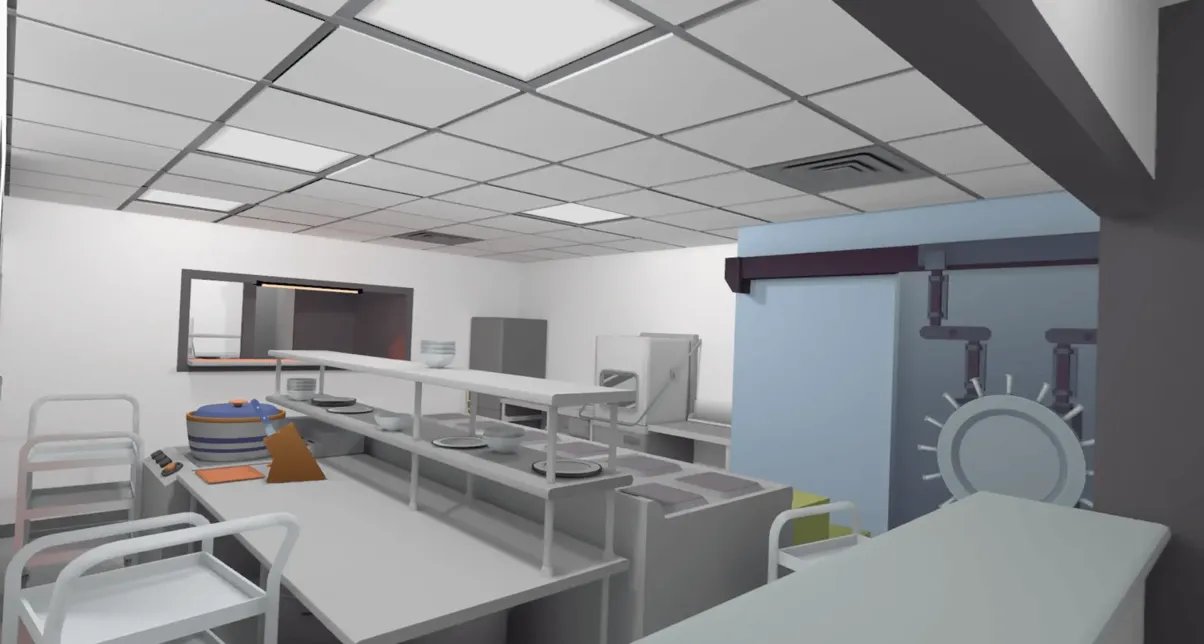}\\{\scriptsize\ttfamily extract\_6d71ed0b.jpg}\end{minipage}\hfill \begin{minipage}[t]{0.16\linewidth}\centering\includegraphics[width=\linewidth,height=0.11\textheight,keepaspectratio]{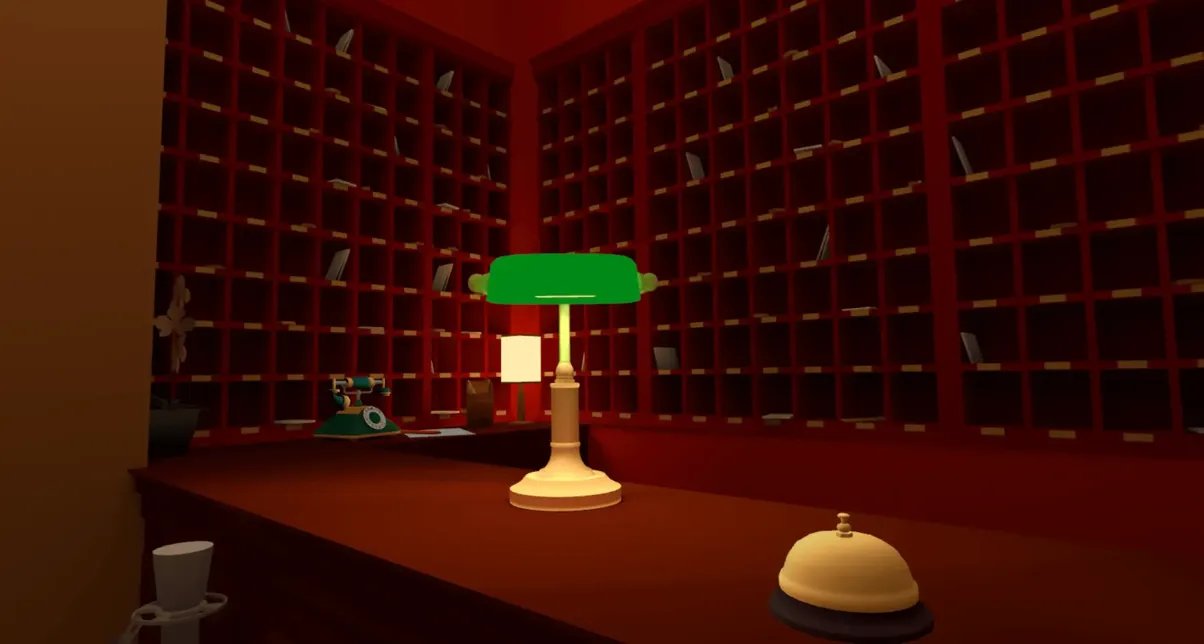}\\{\scriptsize\ttfamily extract\_5eba61bd.jpg}\end{minipage}\hfill \begin{minipage}[t]{0.16\linewidth}\centering\includegraphics[width=\linewidth,height=0.11\textheight,keepaspectratio]{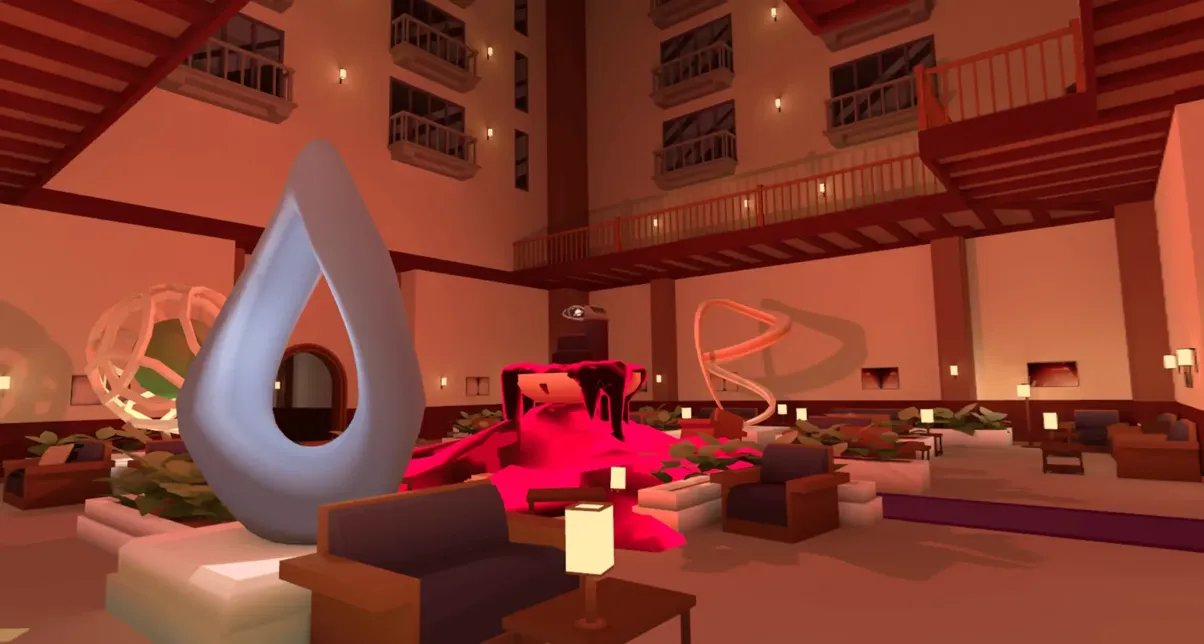}\\{\scriptsize\ttfamily extract\_a807c31b.jpg}\end{minipage}\hfill \begin{minipage}[t]{0.16\linewidth}\centering\includegraphics[width=\linewidth,height=0.11\textheight,keepaspectratio]{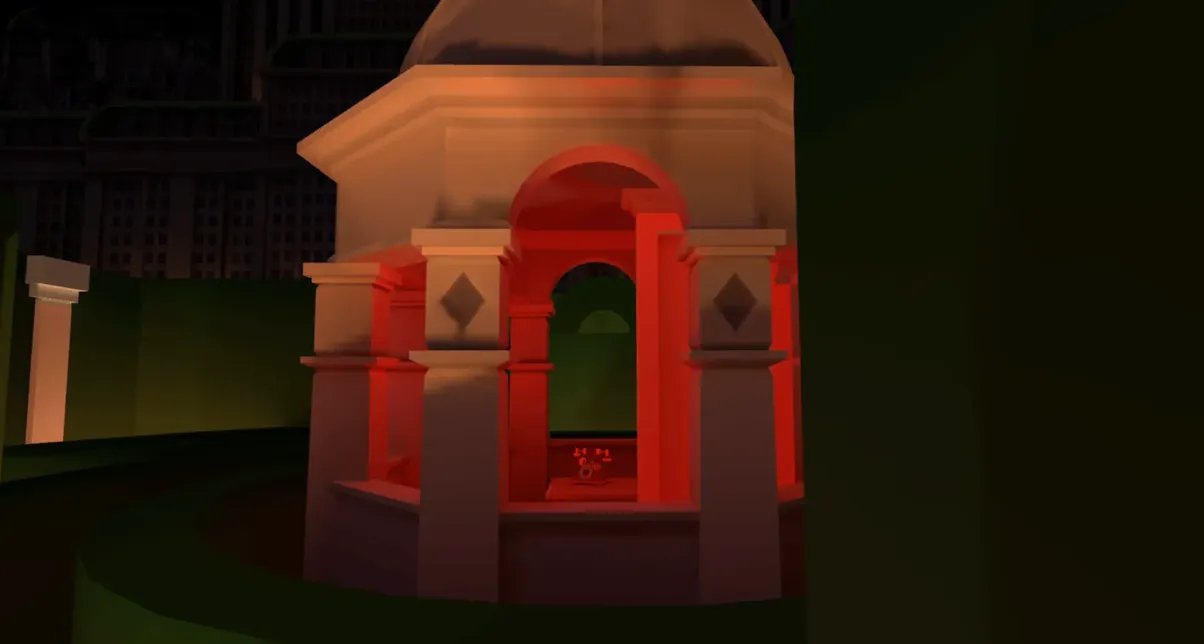}\\{\scriptsize\ttfamily extract\_6475e49c.jpg}\end{minipage}\par\vspace{2pt}{\scriptsize\sffamily\color{black!50} $\triangleright$ Full tool return (1,544 characters)}
\end{tcolorbox}
\begin{tcolorbox}[
  colback=thinkbg, colframe=thinkhead, colbacktitle=thinkhead, coltitle=white,
  boxrule=0.45pt, arc=1.4mm, left=5pt,right=5pt,top=4pt,bottom=4pt,
  toptitle=1.2pt,bottomtitle=1.2pt, fonttitle=\sffamily\bfseries\small,
  fontupper=\footnotesize\sffamily, breakable,
  title={Think $|$ Round 8}, before skip=3.5pt, after skip=3.5pt,
]
Excellent! I found the exact screenshot from the official website. Image extract\_807c31b.jpg shows the atrium scene perfectly. Let me examine it carefully:\par - The large, fluid-like bright-red object is in the center of the hall - To its left is a large light-blue decoration - The shape of this light-blue decoration is clearly a teardrop (or water drop) shape - it's a large, smooth, teardrop-shaped sculpture\par Let me also verify the award information. I found that Hotel Infinity won the "Immersive Reality Technical Achievement" at the 29th D.I.C.E. Awards on February 12, 2026. Let me extract that in ...
\end{tcolorbox}
\begin{tcolorbox}[
  colback=callbg, colframe=callhead, colbacktitle=callhead, coltitle=white,
  boxrule=0.45pt, arc=1.4mm, left=5pt,right=5pt,top=4pt,bottom=4pt,
  toptitle=1.2pt,bottomtitle=1.2pt, fonttitle=\sffamily\bfseries\small,
  fontupper=\footnotesize\sffamily, breakable,
  title={Tool call $|$ web\_extract $|$ Round 8}, before skip=3.5pt, after skip=3.5pt,
]
{\ttfamily\small \{"urls": ["http://www.shacknews.com/***"], "goal": "Confirm that Hotel Infinity won an award at the 29th D.I.C.E. Awards in February 2026 and identify the specific award"\}}
\end{tcolorbox}
\begin{tcolorbox}[
  colback=obsbg, colframe=obshead, colbacktitle=obshead, coltitle=white,
  boxrule=0.45pt, arc=1.4mm, left=5pt,right=5pt,top=4pt,bottom=4pt,
  toptitle=1.2pt,bottomtitle=1.2pt, fonttitle=\sffamily\bfseries\small,
  fontupper=\footnotesize\sffamily, breakable,
  title={Tool response $|$ Round 8}, before skip=3.5pt, after skip=3.5pt,
]
{\sffamily\bfseries Summary preview}
\par{\sffamily\color{thinkhead}\bfseries The D.I.C.E. Awards 2026 winners \& finalists}
\par{\scriptsize\ttfamily\color{black!55} http://www.shacknews.com/***}
\par\vspace{3pt}\noindent \begin{minipage}[t]{0.16\linewidth}\centering\includegraphics[width=\linewidth,height=0.11\textheight,keepaspectratio]{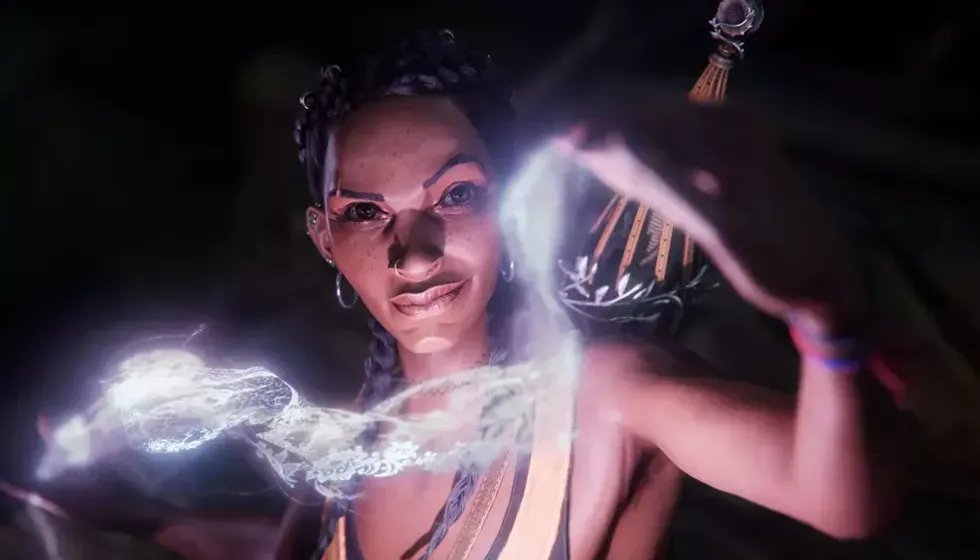}\\{\scriptsize\ttfamily extract\_cf26c5f2.jpg}\end{minipage}\hfill \begin{minipage}[t]{0.16\linewidth}\centering\includegraphics[width=\linewidth,height=0.11\textheight,keepaspectratio]{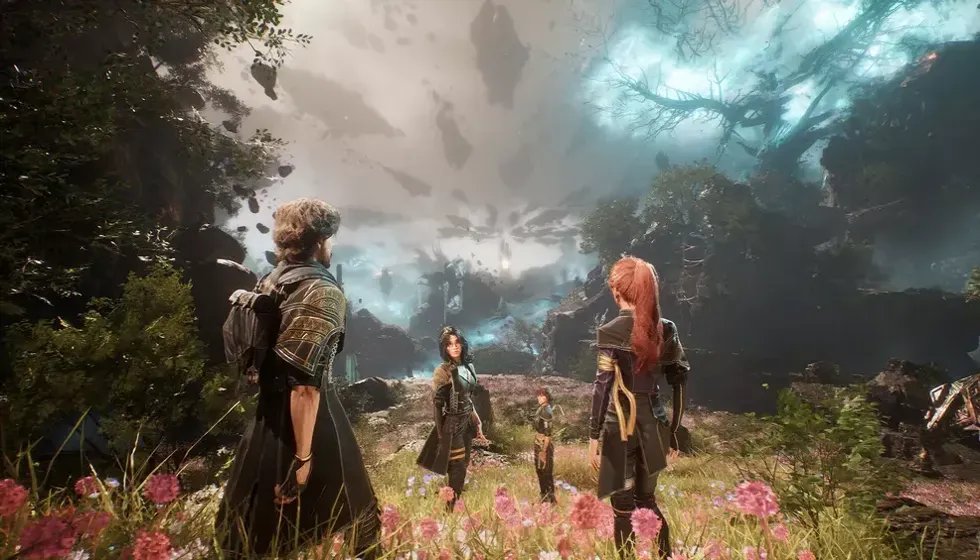}\\{\scriptsize\ttfamily extract\_21a27db5.jpg}\end{minipage}\hfill \begin{minipage}[t]{0.16\linewidth}\centering\includegraphics[width=\linewidth,height=0.11\textheight,keepaspectratio]{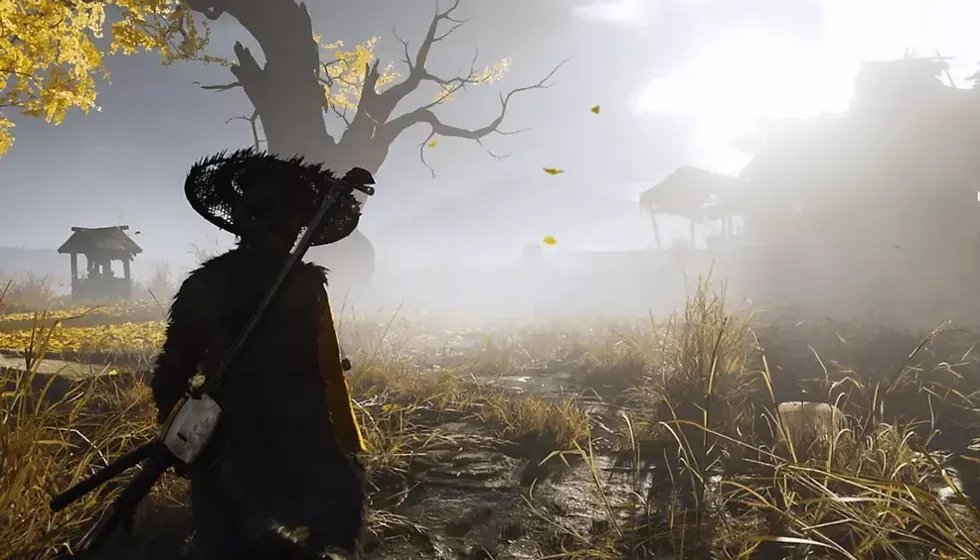}\\{\scriptsize\ttfamily extract\_9fcd2c5c.jpg}\end{minipage}\hfill \begin{minipage}[t]{0.16\linewidth}\centering\includegraphics[width=\linewidth,height=0.11\textheight,keepaspectratio]{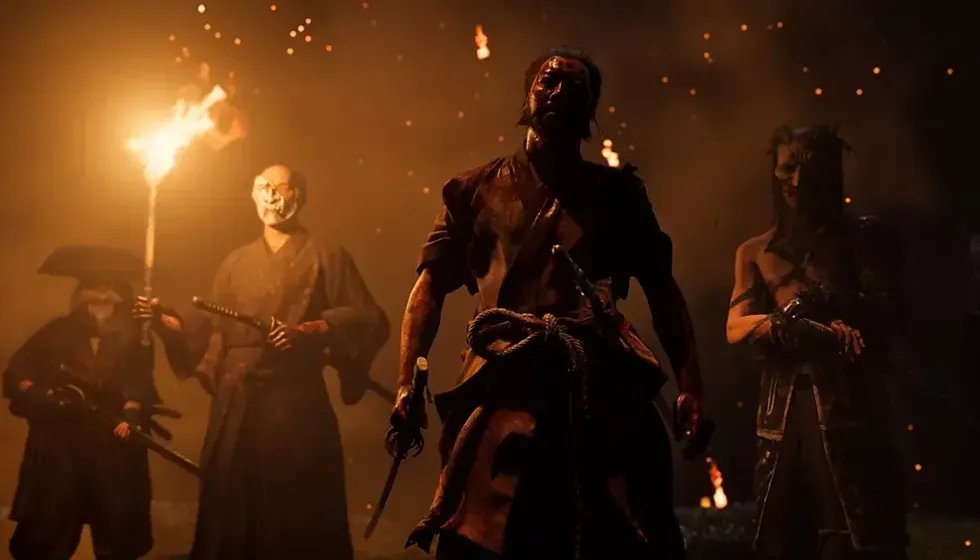}\\{\scriptsize\ttfamily extract\_6166642b.jpg}\end{minipage}\par\vspace{2pt}{\scriptsize\sffamily\color{black!50} $\triangleright$ Full tool return (1,693 characters)}
\end{tcolorbox}
\begin{tcolorbox}[
  colback=thinkbg, colframe=thinkhead, colbacktitle=thinkhead, coltitle=white,
  boxrule=0.45pt, arc=1.4mm, left=5pt,right=5pt,top=4pt,bottom=4pt,
  toptitle=1.2pt,bottomtitle=1.2pt, fonttitle=\sffamily\bfseries\small,
  fontupper=\footnotesize\sffamily, breakable,
  title={Think $|$ Final}, before skip=3.5pt, after skip=3.5pt,
]
Use the verifiable evidence gathered in the trajectory and answer the question as requested.
\end{tcolorbox}
\begin{tcolorbox}[
  colback=webg, colframe=wehead, colbacktitle=wehead, coltitle=white,
  boxrule=0.45pt, arc=1.4mm, left=5pt,right=5pt,top=4pt,bottom=4pt,
  toptitle=1.2pt,bottomtitle=1.2pt, fonttitle=\sffamily\bfseries\small,
  fontupper=\footnotesize\sffamily, breakable,
  title={Final response}, before skip=3.5pt, after skip=3.5pt,
]
Based on my research:\par 1. {\bfseries The game}: The puzzle game released in 2019 with impossible architecture is {\bfseries Manifold Garden}, developed by William Chyr. The development team later released {\bfseries Hotel Infinity} (2025), a VR title designed for room-scale experience.\par 2. {\bfseries The award}: In February 2026, Hotel Infinity won the {\bfseries Immersive Reality Technical Achievement} award at the {\bfseries 29th D.I.C.E. Awards}, which took place on February 12, 2026 at the Aria Resort \& Casino in Las Vegas.\par 3. {\bfseries The screenshot} (from the official Hotel Infinity website): In the atrium/\allowbreak{}lounge hall screenshot, there is a large, fluid-like bright-red object (glowing red ooze-like liquid) in the center. To its left is a large light-blue decoration shaped like a {\bfseries teardrop} (or water drop).
\par\vspace{4pt}\noindent \begin{minipage}[t]{0.55\linewidth}\centering\includegraphics[width=\linewidth,height=0.18\textheight,keepaspectratio]{images/rl_response_cases/case_0143_extract_a807c31b.jpg}\\{\scriptsize\ttfamily extract\_a807c31b.jpg}\end{minipage}
\end{tcolorbox}


\begin{thebibliography}{36}
\providecommand{\natexlab}[1]{#1}
\providecommand{\url}[1]{\texttt{#1}}
\expandafter\ifx\csname urlstyle\endcsname\relax
  \providecommand{\doi}[1]{doi: #1}\else
  \providecommand{\doi}{doi: \begingroup \urlstyle{rm}\Url}\fi

\bibitem[{Anthropic}(2026)]{anthropic2026opus5}
{Anthropic}.
\newblock Claude opus 5 system card.
\newblock \url{https://www.anthropic.com/research/claude-opus-5}, 2026.

\bibitem[Bai et~al.(2025)Bai, Cai, Chen, Chen, Chen, Cheng, Deng, Ding, Gao, Ge, et~al.]{bai2025qwen3vl}
Shuai Bai, Yuxuan Cai, Ruizhe Chen, Keqin Chen, Xionghui Chen, Zesen Cheng, Lianghao Deng, Wei Ding, Chang Gao, Chunjiang Ge, et~al.
\newblock Qwen3-vl technical report.
\newblock \emph{arXiv preprint arXiv:2511.21631}, 2025.

\bibitem[Chen et~al.(2026)Chen, Feng, Chen, Huang, Dai, Shou, Lin, Yue, Gao, and Pang]{chen2026opensearchvl}
Shuang Chen, Kaituo Feng, Hangting Chen, Wenxuan Huang, Dasen Dai, Quanxin Shou, Yunlong Lin, Xiangyu Yue, Shenghua Gao, and Tianyu Pang.
\newblock Opensearch-vl: An open recipe for frontier multimodal search agents.
\newblock \emph{arXiv preprint arXiv:2605.05185}, 2026.

\bibitem[Cheng et~al.(2025)Cheng, Zhang, Zhang, Yang, Guan, Wu, Li, Zhang, Liu, Mai, et~al.]{cheng2025simplevqa}
Xianfu Cheng, Wei Zhang, Shiwei Zhang, Jian Yang, Xiangyuan Guan, Xianjie Wu, Xiang Li, Ge~Zhang, Jiaheng Liu, Yuying Mai, et~al.
\newblock Simplevqa: Multimodal factuality evaluation for multimodal large language models.
\newblock In \emph{Proceedings of the IEEE/CVF International Conference on Computer Vision}, pages 4637--4646, 2025.

\bibitem[Chu et~al.(2026)Chu, Wang, Hong, Fan, Huang, Yang, Xu, Zhao, Xiang, Hu, Kuang, Liu, Qin, and Yu]{chu2026redsearcher}
Zheng Chu, Xiao Wang, Jack Hong, Huiming Fan, Yuqi Huang, Yue Yang, Guohai Xu, Chenxiao Zhao, Cheng Xiang, Shengchao Hu, Dongdong Kuang, Ming Liu, Bing Qin, and Xing Yu.
\newblock Redsearcher: A scalable and cost-efficient framework for long-horizon search agents.
\newblock \emph{arXiv preprint arXiv:2602.14234}, 2026.

\bibitem[Fu et~al.(2025)Fu, Peng, Liu, Wan, and Chen]{fu2025livevqa}
Mingyang Fu, Yuyang Peng, Benlin Liu, Yao Wan, and Dongping Chen.
\newblock Livevqa: Live visual knowledge seeking.
\newblock \emph{arXiv preprint arXiv:2504.05288}, 2025.

\bibitem[{Gemini Team} et~al.(2023){Gemini Team}, Anil, Borgeaud, Alayrac, Yu, Soricut, Schalkwyk, Dai, Hauth, Millican, et~al.]{team2023gemini}
{Gemini Team}, Rohan Anil, Sebastian Borgeaud, Jean-Baptiste Alayrac, Jiahui Yu, Radu Soricut, Johan Schalkwyk, Andrew~M. Dai, Anja Hauth, Katie Millican, et~al.
\newblock Gemini: A family of highly capable multimodal models.
\newblock \emph{arXiv preprint arXiv:2312.11805}, 2023.

\bibitem[Geng et~al.(2025)Geng, Xia, Zhang, Wang, Wang, Ding, Wang, Wu, Zhao, Li, et~al.]{webwatcher}
Xinyu Geng, Peng Xia, Zhen Zhang, Xinyu Wang, Qiuchen Wang, Ruixue Ding, Chenxi Wang, Jialong Wu, Yida Zhao, Kuan Li, et~al.
\newblock Webwatcher: Breaking new frontiers of vision-language deep research agent.
\newblock \emph{arXiv preprint arXiv:2508.05748}, 2025.

\bibitem[{Google DeepMind}(2026)]{google2026gemini36flash}
{Google DeepMind}.
\newblock Gemini 3.6 flash model card.
\newblock \url{https://deepmind.google/models/model-cards/gemini-3-6-flash/}, 2026.

\bibitem[Hu et~al.(2023)Hu, Iscen, Sun, Chang, Sun, Ross, Schmid, and Fathi]{hu2023avis}
Ziniu Hu, Ahmet Iscen, Chen Sun, Kai-Wei Chang, Yizhou Sun, David~A. Ross, Cordelia Schmid, and Alireza Fathi.
\newblock Avis: Autonomous visual information seeking with large language model agent.
\newblock In \emph{Advances in Neural Information Processing Systems}, 2023.

\bibitem[Huang et~al.(2026)Huang, Zeng, Wang, Fang, Cao, Chu, Yin, Chen, Yin, Chen, et~al.]{huang2026vision}
Wenxuan Huang, Yu~Zeng, Qiuchen Wang, Zhen Fang, Shaosheng Cao, Zheng Chu, Qingyu Yin, Shuang Chen, Zhenfei Yin, Lin Chen, et~al.
\newblock Vision-deepresearch: Incentivizing deepresearch capability in multimodal large language models.
\newblock \emph{arXiv preprint arXiv:2601.22060}, 2026.

\bibitem[Jiang et~al.(2025)Jiang, Zhang, Guo, Wu, Lei, Qiu, Lu, Chen, Song, Gao, et~al.]{jiang2024mmsearch}
Dongzhi Jiang, Renrui Zhang, Ziyu Guo, Yanmin Wu, Jiayi Lei, Pengshuo Qiu, Pan Lu, Zehui Chen, Guanglu Song, Peng Gao, et~al.
\newblock Mmsearch: Unveiling the potential of large models as multi-modal search engines.
\newblock In \emph{International Conference on Learning Representations}, 2025.

\bibitem[Jin et~al.(2025)Jin, Zeng, Yue, Yoon, Arik, Wang, Zamani, and Han]{jin2025searchr1}
Bowen Jin, Hansi Zeng, Zhenrui Yue, Jinsung Yoon, Sercan~O. Arik, Dong Wang, Hamed Zamani, and Jiawei Han.
\newblock Search-r1: Training llms to reason and leverage search engines with reinforcement learning.
\newblock \emph{arXiv preprint arXiv:2503.09516}, 2025.

\bibitem[Li et~al.(2026)Li, Chen, Xu, Zhang, and Lu]{li2026hypereyes}
Guankai Li, Jiabin Chen, Yi~Xu, Xichen Zhang, and Yuan Lu.
\newblock Hypereyes: Dual-grained efficiency-aware reinforcement learning for parallel multimodal search agents.
\newblock \emph{arXiv preprint arXiv:2605.07177}, 2026.

\bibitem[Li et~al.(2025{\natexlab{a}})Li, Zhang, Yin, Zhang, Ou, Wu, Yin, Li, Tao, Wang, Shen, Zhang, Zhang, Wu, Jiang, Yan, Xie, Huang, and Zhou]{li2025websailor}
Kuan Li, Zhongwang Zhang, Huifeng Yin, Liwen Zhang, Litu Ou, Jialong Wu, Wenbiao Yin, Baixuan Li, Zhengwei Tao, Xinyu Wang, Weizhou Shen, Junkai Zhang, Dingchu Zhang, Xixi Wu, Yong Jiang, Ming Yan, Pengjun Xie, Fei Huang, and Jingren Zhou.
\newblock Websailor: Navigating super-human reasoning for web agent.
\newblock \emph{arXiv preprint arXiv:2507.02592}, 2025{\natexlab{a}}.

\bibitem[Li et~al.(2025{\natexlab{b}})Li, Bu, Wang, Liu, Dong, et~al.]{li2025mmbrowsecomp}
Shilong Li, Xingyuan Bu, Wenjie Wang, Jiaheng Liu, Jun Dong, et~al.
\newblock Mm-browsecomp: A comprehensive benchmark for multimodal browsing agents.
\newblock \emph{arXiv preprint arXiv:2508.13186}, 2025{\natexlab{b}}.

\bibitem[{Moonshot AI}(2026)]{moonshotai2026k26}
{Moonshot AI}.
\newblock Kimi k2.6 model card.
\newblock \url{https://huggingface.co/moonshotai/Kimi-K2.6}, 2026.

\bibitem[Narayan et~al.(2025)Narayan, Xu, Cao, Nerella, Patel, Shiee, Grasch, Jia, Yang, and Gan]{deepmmsearch_r1}
Kartik Narayan, Yang Xu, Tian Cao, Kavya Nerella, Vishal~M. Patel, Navid Shiee, Peter Grasch, Chao Jia, Yinfei Yang, and Zhe Gan.
\newblock Deepmmsearch-r1: Empowering multimodal llms in multimodal web search.
\newblock \emph{arXiv preprint arXiv:2510.12801}, 2025.

\bibitem[{Nous Research}(2026)]{nousresearch2026hermes}
{Nous Research}.
\newblock Hermes agent.
\newblock \url{https://github.com/NousResearch/hermes-agent}, 2026.

\bibitem[{OpenAI}(2024)]{openai2024gpt4ocard}
{OpenAI}.
\newblock Gpt-4o system card, 2024.

\bibitem[{OpenAI}(2026)]{openai2026gpt56sol}
{OpenAI}.
\newblock Gpt-5.6 sol model card.
\newblock \url{https://developers.openai.com/api/docs/models/gpt-5.6-sol}, 2026.

\bibitem[{Qwen Team}(2026{\natexlab{a}})]{qwen2026qwen35}
{Qwen Team}.
\newblock Qwen3.5-plus model card.
\newblock \url{https://huggingface.co/Qwen/Qwen3.5-397B-A17B}, 2026{\natexlab{a}}.

\bibitem[{Qwen Team}(2026{\natexlab{b}})]{qwen2026qwen36}
{Qwen Team}.
\newblock Qwen3.6-35b-a3b model card.
\newblock \url{https://huggingface.co/Qwen/Qwen3.6-35B-A3B}, 2026{\natexlab{b}}.

\bibitem[{Qwen Team}(2026{\natexlab{c}})]{qwen2026qwen38}
{Qwen Team}.
\newblock Qwen3.8-27b model card.
\newblock \url{https://huggingface.co/Qwen/Qwen3.8-27B}, 2026{\natexlab{c}}.

\bibitem[Shoeybi et~al.(2019)Shoeybi, Patwary, Puri, LeGresley, Casper, and Catanzaro]{shoeybi2019megatron}
Mohammad Shoeybi, Mostofa Patwary, Raul Puri, Patrick LeGresley, Jared Casper, and Bryan Catanzaro.
\newblock Megatron-lm: Training multi-billion parameter language models using model parallelism.
\newblock \emph{arXiv preprint arXiv:1909.08053}, 2019.

\bibitem[Tao et~al.(2026)Tao, Teng, Su, Fu, Wu, Tao, Liu, Bai, Liu, and Kong]{tao2025mmsearchplus}
Xijia Tao, Yihua Teng, Xinxing Su, Xinyu Fu, Jihao Wu, Chaofan Tao, Ziru Liu, Haoli Bai, Rui Liu, and Lingpeng Kong.
\newblock Mmsearch-plus: Benchmarking provenance-aware search for multimodal browsing agents.
\newblock In \emph{International Conference on Learning Representations}, 2026.

\bibitem[Wang et~al.(2018)Wang, Wu, Shen, Dick, and van~den Hengel]{wang2018fvqa}
Peng Wang, Qi~Wu, Chunhua Shen, Anthony Dick, and Anton van~den Hengel.
\newblock Fvqa: Fact-based visual question answering.
\newblock \emph{IEEE Transactions on Pattern Analysis and Machine Intelligence}, 40\penalty0 (10):\penalty0 2413--2427, 2018.

\bibitem[{Wikimedia Foundation}(2026)]{wikipedia}
{Wikimedia Foundation}.
\newblock {Wikipedia}.
\newblock \url{https://www.wikipedia.org}, 2026.

\bibitem[Wu et~al.(2025)Wu, Deng, Li, Liu, You, Li, Ma, and Liu]{wu2025mmsearchr1}
Jinming Wu, Zihao Deng, Wei Li, Yiding Liu, Bo~You, Bo~Li, Zejun Ma, and Ziwei Liu.
\newblock Mmsearch-r1: Incentivizing lmms to search.
\newblock \emph{arXiv preprint arXiv:2506.20670}, 2025.

\bibitem[Yan et~al.(2026)Yan, Tong, Xue, Tang, Wang, Shi, Zhang, Li, and Zou]{yan2026metis}
Shilin Yan, Jintao Tong, Hongwei Xue, Xiaojun Tang, Yangyang Wang, Kunyu Shi, Guannan Zhang, Ruixuan Li, and Yixiong Zou.
\newblock Act wisely: Cultivating meta-cognitive tool use in agentic multimodal models.
\newblock \emph{arXiv preprint arXiv:2604.08545}, 2026.

\bibitem[Yao et~al.(2022)Yao, Zhao, Yu, Du, Shafran, Narasimhan, and Cao]{yao2022react}
Shunyu Yao, Jeffrey Zhao, Dian Yu, Nan Du, Izhak Shafran, Karthik~R. Narasimhan, and Yuan Cao.
\newblock React: Synergizing reasoning and acting in language models.
\newblock In \emph{International Conference on Learning Representations}, 2022.

\bibitem[Zeng et~al.(2026)Zeng, Huang, Fang, Chen, Shen, Cai, Wang, Yin, Chen, Chen, et~al.]{zeng2026vdrbench}
Yu~Zeng, Wenxuan Huang, Zhen Fang, Shuang Chen, Yufan Shen, Yishuo Cai, Xiaoman Wang, Zhenfei Yin, Lin Chen, Zehui Chen, et~al.
\newblock Vision-deepresearch benchmark: Rethinking visual and textual search for multimodal large language models.
\newblock \emph{arXiv preprint arXiv:2602.02185}, 2026.

\bibitem[Zhang et~al.(2025)Zhang, Hu, Sun, Wang, Wei, Yin, Pei, Shen, Xia, Peng, Xie, Li, Liu, Song, and Zhou]{zhang2025skyworkr1v4}
Yifan Zhang, Liang Hu, Haofeng Sun, Peiyu Wang, Yichen Wei, Shukang Yin, Jiangbo Pei, Wei Shen, Peng Xia, Yi~Peng, Tianyidan Xie, Eric Li, Yang Liu, Xuchen Song, and Yahui Zhou.
\newblock Skywork-r1v4: Toward agentic multimodal intelligence through interleaved thinking with images and deepresearch.
\newblock \emph{arXiv preprint arXiv:2512.02395}, 2025.

\bibitem[Zhang et~al.(2024)Zhang, Zhang, Ding, and Yue]{zhang2024vsa}
Zhixin Zhang, Yiyuan Zhang, Xiaohan Ding, and Xiangyu Yue.
\newblock Vision search assistant: Empower vision-language models as multimodal search engines.
\newblock \emph{arXiv preprint arXiv:2410.21220}, 2024.

\bibitem[Zheng et~al.(2025)Zheng, Liu, Li, Chen, Yu, Gao, Dang, Liu, Men, Yang, Zhou, and Lin]{zheng2025gspo}
Chujie Zheng, Shixuan Liu, Mingze Li, Xiong-Hui Chen, Bowen Yu, Chang Gao, Kai Dang, Yuqiong Liu, Rui Men, An~Yang, Jingren Zhou, and Junyang Lin.
\newblock Group sequence policy optimization.
\newblock \emph{arXiv preprint arXiv:2507.18071}, 2025.

\bibitem[Zheng et~al.(2024)Zheng, Yin, Xie, Sun, Huang, Yu, Cao, Kozyrakis, Stoica, Gonzalez, Barrett, and Sheng]{zheng2024sglang}
Lianmin Zheng, Liangsheng Yin, Zhiqiang Xie, Chuyue Sun, Jeff Huang, Cody~Hao Yu, Shiyi Cao, Christos Kozyrakis, Ion Stoica, Joseph~E. Gonzalez, Clark Barrett, and Ying Sheng.
\newblock Sglang: Efficient execution of structured language model programs.
\newblock \emph{arXiv preprint arXiv:2312.07104}, 2024.

\end{thebibliography}
\end{document}